\documentclass{article}

\usepackage{PRIMEarxiv}

\usepackage[utf8]{inputenc} 
\usepackage[T1]{fontenc}    
\usepackage{hyperref}       
\usepackage{url}            
\usepackage{booktabs, multirow}       
\usepackage{amsfonts}       
\usepackage{amsmath}
\usepackage{nicefrac}       
\usepackage{microtype}      
\usepackage{fancyhdr}       
\usepackage{graphicx} 
\usepackage{array}
\usepackage{makecell}
\graphicspath{{media/}}     
\usepackage{lipsum}
\usepackage{lineno}
\usepackage{float}
\usepackage{hyperref}

\usepackage{soul}
\usepackage[table]{xcolor} 
\usepackage{changepage,threeparttable} 
\usepackage{tabularray}

\title{A benchmark for computational analysis of animal behavior, using animal-borne tags
}

\author{
  Benjamin Hoffman$^*$ \\ Earth Species Project \\ \texttt{benjamin@earthspecies.org} \And 
  Maddie Cusimano$^*$ \\ Earth Species Project \\ \texttt{maddie@earthspecies.org} \AND 
  Vittorio Baglione \\ Univ. de Le\'on \\ \texttt{vbag@unileon.es} \And 
  Daniela Canestrari \\ Univ. de Le\'on \\ \texttt{dcan@unileon.es}  \And
  Damien Chevallier \\ CNRS Borea \\ \texttt{damien.chevallier@cnrs.fr } \AND
  Dominic L. DeSantis \\ Georgia College \& State Univ. \\ \texttt{dominic.desantis@gcsu.edu} \And
  Lor\`ene Jeantet\\ African Institute for \\ Mathematical Sciences, \\ Univ. of Stellenbosch \\\texttt{lorene@aims.ac.za} \And
  Monique A. Ladds \\ Department of Conservation, \\ New Zealand \\ \texttt{monique.ladds@gmail.com} \AND
  Takuya Maekawa \\ Osaka Univ. \\ \texttt{takuya.maekawa@acm.org} \And
  Vicente Mata-Silva \\ Univ. Texas El Paso \\ \texttt{vmata@utep.edu} \And
  V\'ictor Moreno-González \\ Univ. de Le\'on \\ \texttt{vmorg@unileon.es} \AND
  Anthony M. Pagano \\ U.S. Geological Survey \\ \texttt{apagano@usgs.gov} \And
  Eva Trapote \\ Univ. de Le\'on \\ \texttt{etrav@unileon.es} \And
  Outi Vainio \\ Univ. of Helsinki \\ \texttt{outi.vainio@helsinki.fi} \AND
  Antti Vehkaoja \\ Tampere Univ. \\ \texttt{antti.vehkaoja@tuni.fi} \And
  Ken Yoda \\ Nagoya Univ. \\ \texttt{yoda.ken@nagoya-u.jp} \And
  Katherine Zacarian \\ Earth Species Project \\ \texttt{katie@earthspecies.org} \AND
  Ari Friedlaender \\ Univ. California Santa Cruz \\ \texttt{ari.friedlaender@ucsc.edu}
}

\usepackage{setspace}

\begin{document}

\maketitle

$^*$ {\footnotesize Equal contribution}

\doublespacing

\begin{abstract} 

\textbf{Background:} Animal-borne sensors (‘bio-loggers’) can record a suite of kinematic and environmental data, which are used to elucidate animal ecophysiology and improve conservation efforts. Machine learning techniques are used for interpreting the large amounts of data recorded by bio-loggers, but there exists no common framework for comparing the different machine learning techniques in this domain. This makes it difficult to, for example, identify patterns in what works well for machine learning-based analysis of bio-logger data. It also makes it difficult to evaluate the effectiveness of novel methods developed by the machine learning community.

\textbf{Methods:} To address this, we present the Bio-logger Ethogram Benchmark (BEBE), a collection of datasets with behavioral annotations, as well as a modeling task and evaluation metrics. BEBE is to date the largest, most taxonomically diverse, publicly available benchmark of this type, and includes 1654 hours of data collected from 149 individuals across nine taxa. Using BEBE, we compare the performance of deep and classical machine learning methods for identifying animal behaviors based on bio-logger data. As an example usage of BEBE, we test an approach based on self-supervised learning. To apply this approach to animal behavior classification, we adapt a deep neural network pre-trained with 700,000 hours of data collected from human wrist-worn accelerometers.

\textbf{Results:} We find that deep neural networks out-perform the classical machine learning methods we tested across all nine datasets in BEBE. We additionally find that the approach based on self-supervised learning out-performs the alternatives we tested, especially in settings when there is a low amount of training data available. 

\textbf{Conclusions:} In light of these results, we are able to make concrete suggestions for designing studies that rely on machine learning to infer behavior from bio-logger data. Therefore, we expect that BEBE will be useful for making similar suggestions in the future, as additional hypotheses about machine learning techniques are tested. Datasets, models, and evaluation code are made publicly available at \url{https://github.com/earthspecies/BEBE}, to enable community use of BEBE.

\end{abstract}


\section{Background}

Animal behavior is of central interest in ecology and evolution because an individual's behavior affects its reproductive opportunities and probability of survival~\cite{daviesIntroductionBehaviouralEcology2012}. Additionally, understanding animal behavior can be key to identifying conservation problems and planning successful management interventions~\cite{berger-talIntegratingAnimalBehavior2011}, for example in rearing captive animals prior to reintroduction~\cite{walters2010status}, designing protected areas~\cite{THAXTER201253}, and reducing dispersal of introduced species~\cite{tingley2013cane}. 

One increasingly utilized approach for monitoring animal behavior is remote recording by animal-borne tags, or \emph{bio-loggers}~\cite{rutzNewFrontiersBiologging2009, wilsonPryingIntimateDetails2008, yoda2001new}. These tags can be composed of multiple sensors such as an accelerometer, gyroscope, altimeter, pressure, GPS receiver, microphone, and/or camera, which record time-series data on an individual’s behavior and their \textit{in situ} environment. Additionally, bio-logger datasets can include data from many-hour tag deployments on multiple individuals. 

\begin{figure}[tbp]
\includegraphics[width=0.88\textwidth]{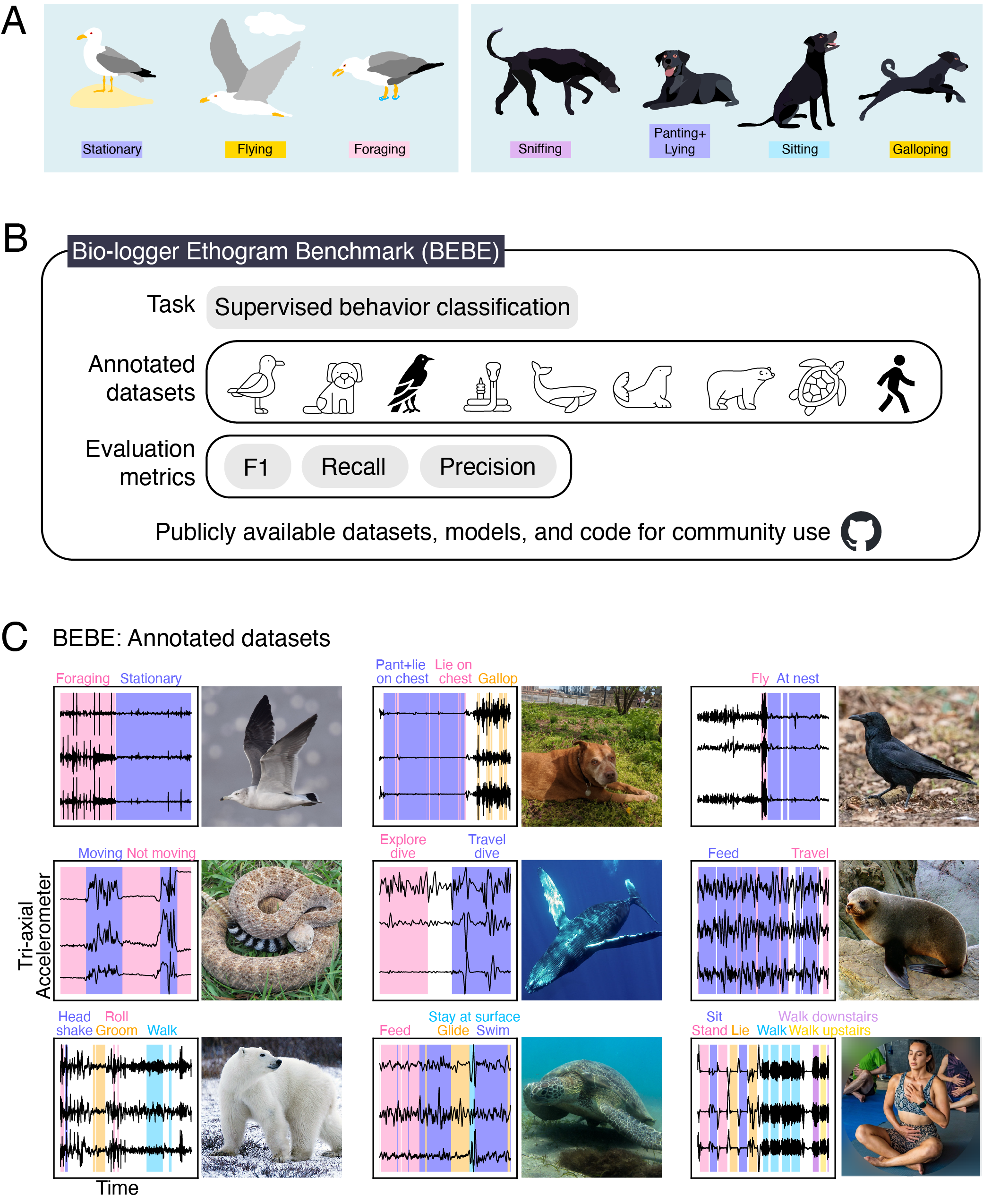}
\caption{A) Examples of ethograms in BEBE. Left: gull ethogram with three behaviors. Right: a subset of the dog ethogram, with four behaviors. B) BEBE consists of a supervised behavior classification task on nine annotated datasets, along with a set of metrics that compare model predictions with the annotations. Datasets and code are publicly available at \url{https://github.com/earthspecies/BEBE}. C) Datasets in BEBE, with a photo of a representative individual and a 5-minute clip of annotated tri-axial accelerometer (TIA) data for each. Each accelerometer channel is min-max scaled for visualization. Top row: black-tailed gull (\textit{Larus crassirostris})~\cite{korpelaMachineLearningEnables2020}, domestic dog (\textit{Canis familiaris})~\cite{kumpulainenDogBehaviourClassification2021, vehkaojaDescriptionMovementSensor2022}, carrion crow (\textit{Corvus corone})~\cite{stidsholt2019tag} (see Methods). Middle row: western diamondback rattlesnake (\textit{Crotalus atrox})~\cite{desantisIntegrativeFrameworkLongTerm2020}, humpback whale (\textit{Megaptera novaeangliae})~\cite{friedlaenderExtremeDielVariation2013}, New Zealand fur seal (\textit{Arctocephalus forsteri})~\cite{laddsSeeingItAll2016}. Bottom row: polar bear (\textit{Ursus maritimus})~\cite{paganoMetabolicRateBody2018, paganoUsingTriaxialAccelerometers2017}, sea turtle (\textit{Chelonia mydas})~\cite{jeantetBehaviouralInferenceSignal2020}, human (\textit{Homo sapiens})~\cite{anguitaPublicDomainDataset2013}. Gaps indicate that the behavior annotation is \textit{Unknown}. For image attributions, see acknowledgments.}\label{fig:intro}
\end{figure}

To give a behavioral interpretation to recorded bio-logger data, it is useful to construct an inventory of what types of actions an individual may perform~\cite{batesonMeasuringBehaviourIntroductory2021}. This inventory, or \textit{ethogram}, is then used to classify observed actions (Figure~\ref{fig:intro}A). Using an ethogram, one can quantify, for example, the proportion of time an animal spends in different behavioral states, and how these differ between groups (e.g., sex, age, populations), or change over time (e.g., seasonally), with physiological condition (e.g., healthy vs. sick) or across different environmental contexts (e.g.~\cite{laddsSurrogates}).

For classifying the behaviors underlying bio-logger data, researchers are increasingly using supervised machine learning (ML) techniques~\cite{VALLETTA2017203}. In a typical workflow, a human annotates some of the recorded bio-logger data with the tagged individuals' behavioral states using a pre-determined ethogram, based on observations made simultaneously with data recording. These annotated data are used to train a ML model, which is then used to predict behavioral labels for the remaining un-annotated portion of the dataset. A test dataset, which is held out from the training stage, can be used to evaluate how well the trained model is able to perform this behavior classification task. Using the predicted behavioral labels allows large datasets to be leveraged to address scientific questions, for example, through estimating activity budgets that vary by time or individual~\cite{minasandra2023accelerometer} or by environmental conditions~\cite{studd2022behavioural}.  In this manner, ML can help scientists to minimize manual effort required to ascribe behavioral labels to bio-logger data, or extend behavioral labels to data where manual ground-truthing is not possible.

Much research has applied ML to bio-logger data to establish its use with particular species, as well as to investigate the impact of different decisions made when using ML models for this purpose. Research questions include characterizing which methods are the most accurate, precise, sensitive, interpretable, or rapid (e.g., \cite{nathanUsingTriaxialAcceleration2012,studdBehavioralClassificationLowfrequency2019,brewsterDevelopmentApplicationMachine2018,desantisIntegrativeFrameworkLongTerm2020,jeantetBehaviouralInferenceSignal2020,patterson2019comparison,wilson2018give,laddsSuperMachineLearning2017}), how to reduce the extent of ground-truthing necessary in species that are difficult to observe (e.g., \cite{paganoUsingTriaxialAccelerometers2017, campbell2013creating,https://doi.org/10.1002/ece3.10035,sakamotoCanEthogramsBe2009,leos-barajasAnalysisAnimalAccelerometer2017,hanscom2023study,ferdinandy2020challenges}), and  
what kinds of behaviors are detectable with particular sensors and models (e.g., \cite{kumpulainenDogBehaviourClassification2021,studdBehavioralClassificationLowfrequency2019,surImprovedSupervisedClassification2017,fehlmann2017,shamoun-baranesSensorDataAnimal2012,clarkeUsingTriaxialAccelerometer2021,mccluneTriaxialAccelerometersQuantify2014}). However, the majority of studies focus on data from a single or a few closely-related species, making it difficult to identify patterns in how behavior classification methods are applied across multiple datasets. 

A commonly used tool in ML for improving our understanding of analysis techniques is the \textit{benchmark} (e.g.~\cite{russakovskyImageNetLargeScale2015}). A benchmark consists of a publicly available dataset, a problem statement specifying a model's inputs and the desired outputs (a \textit{task}), and a procedure for quantitatively evaluating a model's success on the task (using one or several \textit{evaluation metrics}). In a common use-case for a benchmark, researchers report the performance of a proposed technique on the benchmark, helping the field to draw comparisons between different techniques and consolidate knowledge about promising directions. 
Developing benchmarks has been identified as an area of focus for ML applications in wildlife conservation~\cite{tuiaPerspectivesMachineLearning2022} and animal behavior~\cite{ng2022animal, chen2023mammalnet}.

For behavior classification from bio-loggers, a benchmark could assess model performance across a breadth of study systems in order to identify relevant patterns, such as how modeling decisions can influence classification performance. Indeed, researchers have analyzed multi-species datasets to identify best practices for other key challenges in bio-logging, such as sensor calibration~\cite{garde2022ecological} and signal processing~\cite{MartnLpez2020OverallDB, qasem2012tri}. Previous studies have applied one or more behavior classification techniques on multiple bio-logger datasets, with varying degrees of variability in the species and individuals included~\cite{otsuka2024exploring,chimientiUseUnsupervisedLearning2016,campbell2013creating,bidderLoveThyNeighbour2014,yuEvaluationMachineLearning2021,shepardIdentificationAnimalMovement2008,wilson2018give,hammond2016using,resheff2024treat}, but none have attempted to compile a managed, publicly available and diverse database that others could compare against as a benchmark.

In order to fill this gap, we present the Bio-logger Ethogram Benchmark (BEBE), designed to capture challenges in behavior classification from diverse bio-logger datasets. BEBE combines nine datasets collected by various research groups, each with behavioral annotations, as well as a supervised behavior classification task with corresponding evaluation metrics (Figure~\ref{fig:intro}B). These datasets are diverse, spanning multiple species, individuals, behavioral states, sampling rates, and sensor types (Figure~\ref{fig:intro}C), as well as large in size, ranging from six to over a thousand hours in duration. We focus on data collected from tri-axial accelerometers (TIA), in addition to gyroscopes, and environmental sensors. TIA are widely incorporated into bio-loggers because they are inexpensive and lightweight~\cite{rutzNewFrontiersBiologging2009}. Additionally, the data they collect has been used to infer behavioral states on the order of seconds, in a wide variety of species~\cite{wilsonPryingIntimateDetails2008}. 

As a first application of BEBE, we make and test several hypotheses about ML usage in bio-logger data (Table~\ref{hypothesestable}). We base these hypotheses on recent trends in ML, as well as based on their potential to influence the workflow of researchers using ML with bio-logger data. First, in many applications of ML, deep neural networks that make predictions based on raw data, out-perform classical ML methods such as random forests, which make predictions based on hand-crafted summary statistics or \textit{features}~\cite{lecunDeepLearning2015}. Deep neural networks that operate on raw data, such as convolutional and recurrent neural networks, have previously been applied to behavior classification in wild non-human animals \cite{otsuka2024exploring,AulsebrookQuantifyingMating2024,jeantet2021fully,schoombie2024identifying}, captive non-human animals~\cite{eerdekensAutomaticEquineActivity2020} and in human activity recognition~\cite{yuan2022selfsupervised,zhangHumanActivityRecognition,ordonezDeepConvolutionalLSTM2016,chenDeepLearningSensorbased2021,hammerlaDeepConvolutionalRecurrent2016,saeedMultitaskSelfSupervisedLearning2019,yangDeepConvolutionalNeural}. We distinguish these from the multilayer perceptron, a type of neural network that utilizes hand-crafted features and has been used in several bio-logging studies (e.g., ~\cite{resheffAcceleRaterWebApplication2014,nathanUsingTriaxialAcceleration2012,yuEvaluationMachineLearning2021}). Studies differ on whether deep neural networks show performance benefits compared to classical methods (e.g., \cite{AulsebrookQuantifyingMating2024,otsuka2024exploring}), and random forests remain the most commonly used ML methods for bio-logger data~\cite[Table 2]{thiebaultAnimalborneAcousticData2021}, for which feature engineering is a key challenging step. In line with ML trends and results from \cite{otsuka2024exploring}, we predict that deep neural networks will outperform techniques using hand-chosen features (H1).

Second, we examine how data recorded for one species can be used to inform behavior predictions for a different species. Some previous works (e.g.~\cite{otsuka2024exploring, paganoUsingTriaxialAccelerometers2017,campbell2013creating}) have adopted a cross-species transfer learning strategy, by training a supervised model on one species and then applying this trained model to another related species. Our approach is different in that we use self-supervised learning. In self-supervised learning, a ML model (typically a deep neural network) is \textit{pre-trained} to perform an auxiliary task on an unlabeled dataset. Importantly, training the model to perform this auxiliary task does not require any human-generated annotations of the data. In this way, self-supervised pre-training can make use of a large amount of un-annotated data that is easy to obtain (e.g., a species where a large dataset exists). 
Later, the pre-trained model can be trained (or \textit{fine-tuned}) to perform the task of interest (such as behavior classification in a different species), using a small amount of annotated data. By learning to perform the auxiliary task the model learns a set of features, which often provide a good set of initial model parameters for performing the task introduced in the fine-tuning step. Inspired by the recent success of self-supervised learning in other domains of ML (e.g. language~\cite{NEURIPS2020_1457c0d6} and computer vision~\cite{9709990}), we predict that a deep neural network pre-trained on a large amount of human accelerometer data~\cite{yuan2022selfsupervised} will outperform alternative methods, after fine-tuning (H2). Moreover, the self-supervised pre-training step can reduce the amount of annotated data required to meet a given level of performance~\cite{NEURIPS2020_fcbc95cc}. Therefore, we predict that this trend will hold when we reduce the amount of training data by a factor of four (reduced data setting; H3). 

Finally, we investigate how the performance of our best models varies by behavior class and how this per-behavior performance scales with the amount of training data. In recent years, performance on some human behavior classification benchmarks has shown little improvement in spite of methodological advancements~\cite{TongDeadEnd2020}. This suggests that the sensor data may not contain sufficient information to discriminate activities of interest. As found for human activity recognition and behavior classification in other animals \cite{clarkeUsingTriaxialAccelerometer2021,fehlmann2017,kumpulainenDogBehaviourClassification2021,mccluneTriaxialAccelerometersQuantify2014,shamoun-baranesSensorDataAnimal2012,studdBehavioralClassificationLowfrequency2019}, we expect that there will be a large degree of variation in per-behavior performance. Here, we test one possibility for improving classification performance: increasing the amount of training data. If some behaviours are not well discriminated by sensor data, we predict that increasing the amount of training data will show only minimal improvement (H4).

While we focus on a specific classification task in this study, all datasets, models, and evaluation code presented in BEBE are available at \url{https://github.com/earthspecies/BEBE} for general community use. Researchers may use the standardized task to test classification methods, or adapt BEBE datasets for their own research questions (see Discussion for examples). Given that one aim of BEBE is to improve our understanding of classification methods in bio-logger data, we are also seeking contributions to create an expanded benchmark with improved taxonomic coverage, a broader range of sensor types, additional standardization, and a wider variety of modeling tasks. Details about how to contribute in this way can also be found at our GitHub.

In summary, the main contributions of this study include:
\begin{enumerate}
\item Publicly available multi-species bio-logger benchmark dataset, centered on tri-axial accelerometers
\item Standardized evaluation framework for supervised behavior classification, with accompanying code and model examples
\item Demonstration of benchmark usage to investigate patterns in ML behavior classification performance (Table~\ref{hypothesestable}), including:
\begin{itemize}
\item A comparison of deep learning and classical techniques on non-human bio-logger data
\item Successful cross-species application of a self-supervised neural network trained on human bio-logger data (based on \cite{yuan2022selfsupervised})
\end{itemize}
\end{enumerate}

\begin{table}
    \centering
    \begin{tabular}{|p{0.8\linewidth}|c|}
    \hline\hline
        \textbf{Hypothesis} & \textbf{Hypothesis confirmed?} \\
         \hline
         (H1) Deep neural network-based approaches will outperform classical approaches based on hand-chosen summary statistics. & Yes, for the approaches we tested \\
         \hline
         (H2) Self-supervised pre-training using human accelerometer data will improve classification performance. & Partly \\
         \hline
         (H3) Self-supervised pre-training using human accelerometer data will improve classification performance when the amount of training data is reduced by a factor of four by removing individuals. & Yes \\
         \hline
         (H4) In terms of a single model's predictive performance, there is minimal improvement in some behavior classes when increasing the amount of training data by four times by adding individuals. & Yes \\
          \hline\hline
    \end{tabular}
    \caption{Hypotheses tested in this work. BEBE provides a means to identify patterns in behavior classification methods applied across multiple species and sensor types.}
    \label{hypothesestable}
\end{table}

\section{Methods}

\subsection{Benchmark Datasets}
\label{benchmark datasets}

We brought together nine animal motion datasets into a benchmark collection called the Bio-logger Ethogram Benchmark (BEBE) (Table~\ref{summarytable}). BEBE introduces a previously unpublished dataset (Crow); otherwise, these data were all collected in previous studies. Of the datasets included in BEBE, four are publicly available for the first time (Whale, Crow, Rattlesnake, Gull) and five were already publicly available (HAR, Polar bear, Sea turtle, Seals, Dog). We summarize datasets' hardware, data collection, ethogram definition, and ground-truthing in Section~\ref{dataset summary section}. For full details, including details on synchronization, calibration, and annotation validation, we refer readers to the original papers.

\begin{figure}[htbp]
\includegraphics[width=0.95\textwidth]{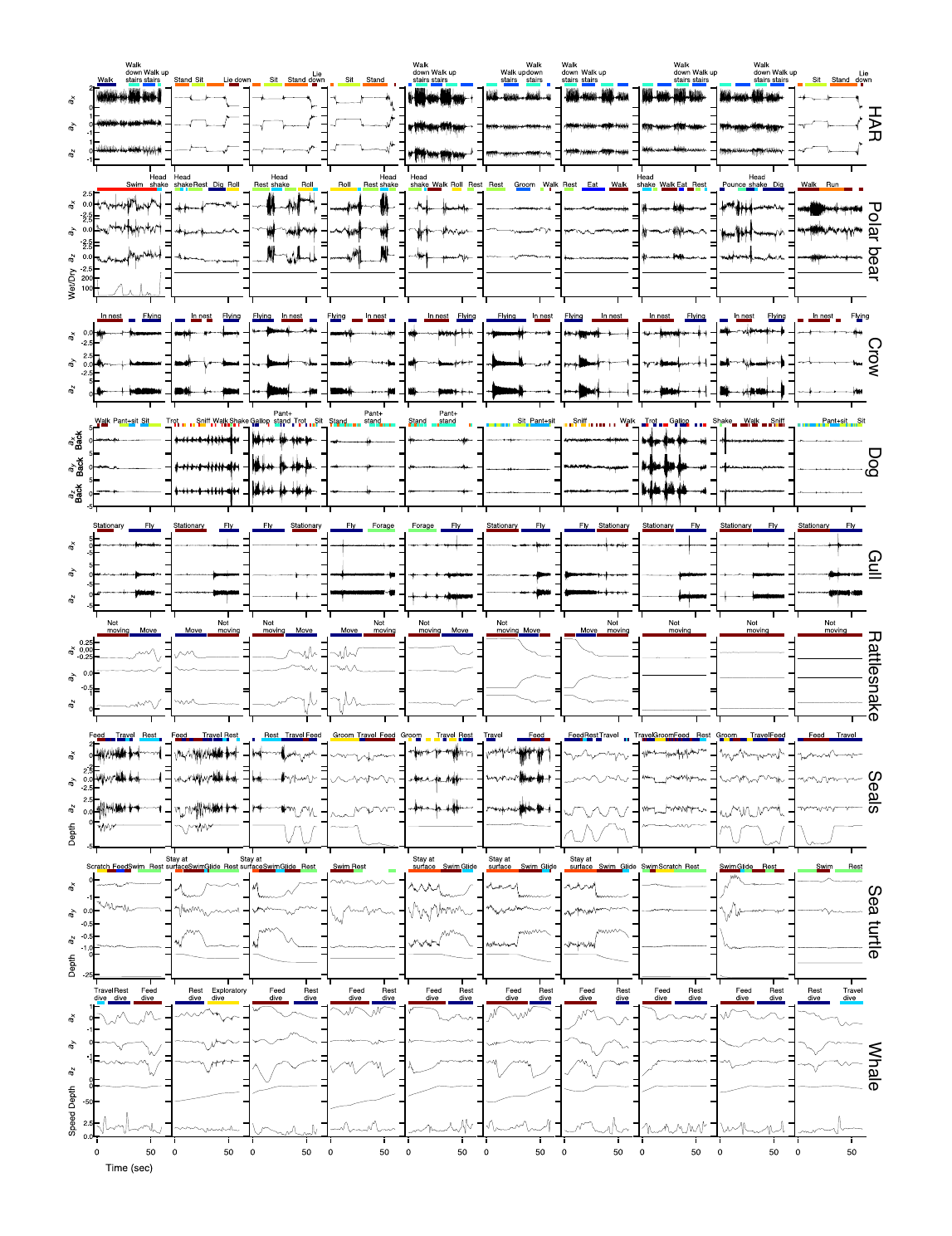}
\caption{Example data from BEBE. Each row displays ten 1-minute clips from one dataset, showing behavior labels, three tri-axial accelerometer channels ($g$), as well as speed ($m/s$), saltwater conductivity (wet/dry), and/or depth ($m$) if available. Examples were chosen to focus on transitions between behaviors. Acceleration traces for behavior classes range from highly stereotyped (e.g., \textit{Sit} in HAR) to highly variable (e.g., \textit{Feed} in Seals). For examples of each behavior in each dataset, with the full set of dataset channels, see Supplemental Figures~\ref{HAR_examples_supplement}-\ref{dog_examples1_supplement}. }
\label{behavior_examples}
\end{figure}

In each dataset, data were recorded by bio-loggers attached to several different individuals of the given species. Each dataset contains one species, except for the Seals dataset which contains four \textit{Otariid} species. These bio-loggers collected kinematic and environmental time series data, such as acceleration, angular velocity, pressure, and conductivity (Figure~\ref{behavior_examples}). While each dataset in BEBE includes acceleration data, different hardware configurations were used across studies. As a result, each dataset comes with its own particular set of data channels, and with its own sampling rate. We used calibrated data as provided by the original dataset authors.

\definecolor{Silver}{rgb}{0.95,0.95,0.95}
\begin{table}
\scriptsize
\centering
\resizebox{\linewidth}{!}{%
\begin{tblr}{
  width = \linewidth,
  colspec = {Q[108]Q[81]Q[135]Q[87]Q[83]Q[100]Q[75]Q[108]Q[83]Q[67]},
  row{even} = {Silver},
  hline{1-2,18} = {-}{},
}
                           & {Human Activity Recognition (HAR) \\ \cite{anguitaPublicDomainDataset2013,reyes2016transition}}                                                          & {Rattlesnake$^*$ \\ \cite{desantisIntegrativeFrameworkLongTerm2020}}                     & {Polar Bear \\ \cite{paganoMetabolicRateBody2018, paganoUsingTriaxialAccelerometers2017}}                                                            & {Dog \\ \cite{kumpulainenDogBehaviourClassification2021, vehkaojaDescriptionMovementSensor2022}}                                                                                                   & {Whale$^*$ \\ \cite{friedlaenderExtremeDielVariation2013} }                                                 & {Sea Turtle  \\ \cite{jeantetBehaviouralInferenceSignal2020}}                                                  & {Seals \\ \cite{laddsSeeingItAll2016,laddsSuperMachineLearning2017}}                                                & {Gull$^*$ \\ \cite{korpelaMachineLearningEnables2020}}                     & {Crow$^*$ \\ (see Methods)}             \\
Species                    & Human                                                        & {Western\\diamondback\\rattlesnake} & Polar bear                                                            & Domestic dog                                                                                          & Humpback whale                                        & Green turtle                                                 & Otariid spp.                                         & Black-tailed gull         & Carrion crow     \\
{Typical\\Body\\mass\\(kg)}             & 60-70                                                        & 1-3                             & {150-300 (F),\\300-800 (M)}                                           & 5-70                                                                                                  & 40000                                                 & 68-190                                                       & {Fur seals: \\40 (F),~140 (M)~\\Sea lion: \\100-300} & .4-.6                     & .4-.6            \\
Location                    & {Genoa,\\ Italy}                                                       & {Texas, USA}                              & {Arctic Ocean}                                                       & {Helsinki,\\Finland}                                                                                                    & {Wilhelmina Bay, \\Western\\Antarctica\\Peninsula}                                                    & {Grande Anse\\d'Arlet,\\Martinique,\\ France}                                                & {Marine \\ facilities \\ on west coast\\ of Australia}                                                   & {Kabushima Island,\\ Japan}                        & {León,\\Spain} \\  
{Tag~\\position}           & Waist                                                        & Body                            & Neck                                                                  & Back and neck                                                                                         & Dorsal surface or flank                               & Carapace                                                     & Back                                                 & Back or abdomen           & Tail             \\
\# indiv.                  & 30                                                           & 13                              & 5                                                                     & 45                                                                                                    & 8                                                     & 14                                                           & 12                                                   & 11                        & 11               \\
{Behavior\\classes}        & {Lie down\\Stand\\Sit\\Walk downstairs\\Walk upstairs\\Walk} & {Moving\\Not moving}            & {Walk\\Swim\\Run\\Roll\\Rest\\Pounce\\Head shake\\Groom\\Eat\\Dig\\~} & {Walk\\Trot\\Stand\\Sniff\\Sit\\Shake\\Pant+stand\\Pant+sit\\Pant+lie on chest\\Lie on\\chest\\Gallop} & {Feed dive\\Exploratory dive\\Travel dive\\Rest dive} & {Swim\\Stay at\\surface\\Scratch\\Rest\\Glide\\Feed\\Breathe} & {Feed\\Groom\\Rest\\Travel}                          & {Fly\\Forage\\Stationary} & {In nest\\Fly}   \\
{Sample~\\rate (Hz)}       & 50                                                           & 1                               & 16                                                                    & 100                                                                                                   & 5                                                     & 20                                                           & 25                                                   & 25                        & 50               \\
{Data\\channels}           & {TIA\\Gyroscope}                                             & TIA                             & {TIA\\Conductivity}                                                   & {2x TIA\\2x gyroscope}                                                                                & {TIA\\Depth\\Speed}                                   & {TIA\\Gyroscope\\Depth}                                      & {TIA\\Depth}                                         & TIA                       & TIA              \\
{Hardware}           & {Galaxy S II\\Samsung}                                             & {AXY-3,\\AXY-4\\(Technosmart)}                             & {Video collar: Exeye \\Logger: TDR10-X-340D\\(Wildlife Computers)}                                                   & {ActiGraph\\GT9X Link\\(ActiGraph LLC)}                                                                                & {
DTAG \cite{johnson2003digital}}                                   & {CATS \\  (Customized Animal\\Tracking Solutions)}                                      & {CEFAS G6a+\\(CEFAS Technology\\Ltd)}                                         & {TDK MPU-9250 (InvenSense)}
& {miniDTAG \cite{stidsholt2019tag}, \\ TIA: KX022-1020 (Kionix)}              \\
{Attach.\\Method}           & {Belt}                                             & {Implant}                             & {Collar}                                                   & {Back: Harness\\Neck: tape}                                                                                & {
Suction}                                   & {Suction}                                      & {Fur seal: tape\\Sea lion:\\harness}                                         & {Tape, harness}                       & {Elastic}              \\
Duration (hrs)             & 6.2                                                          & 30.9                            & 1108.4                                                                & 29.5                                                                                                  & 184.6                                                 & 77.1                                                         & 14.0                                                 & 85.7                      & 114.6            \\
Duration annotated (hrs)   & 4.2                                                          & 30.9                            & 196.1                                                                 & 16.9                                                                                                  & 114.1                                                 & 67.8                                                         & 11.6                                                 & 85.0                      & 3.4              \\
Unknown behaviors (\%)   & 33.4 & 0.0  & 82.3 & 42.7 & 38.2 & 12.1 & 17.0 & 0.8 & 95.7  \\
{Mean\\annot.\\dur. (sec)} & 17.5                                                         & 721.8                           & 127.2                                                                 & 15.5                                                                                                  & 119.8                                                 & 47.2                                                         & 24.8                                                 & 2823.7                    & 14.1             \\
{Annot.\\method}           & {Direct obs.,\\off tag\\video}                                                  & {Direct obs.,\\off tag\\video}                     & {On tag \\video}                                                      & {Direct obs., \\ Off tag \\video}                                                                                     & {Motion,\\on tag audio}                                                & {On tag \\video}                                             & {Direct obs., \\ Off tag \\video}                                    & {On tag \\video}          & {On tag \\audio} \\
License                    & Custom                                                       & {Creative\\Commons}                              & {Public\\ Domain}                                                         & {Creative\\Commons}                                                                                                    & {Creative\\Commons}                                                    & {Public\\Domain}                                                & {Creative\\Commons}                                                   & {Creative\\Commons}                        & {Creative\\Commons}               
\end{tblr}
}

\vspace{1em} \caption{Summary of datasets in BEBE. Out of nine datasets, one comes from humans, three come from other terrestrial species, three come from aquatic species, and two come from flying species. We provide a summary of hardware, data collection, ethogram definition,  and ground-truthing in Section~\ref{dataset summary section}; see the original papers for more details, including details on synchronization, calibration, and annotation validation.
Datasets marked with an asterisk are publicly available for the first time in BEBE. \textit{Duration}: the total duration of the dataset. \textit{Duration annotated}: the duration of annotated data. \textit{Mean annot. dur.}: the average duration an individual spends in a behavioral state (described in Supplemental Information). \textit{Annot. method}: the type of data that was used to make the behavioral annotations. \textit{Attach. method}: manner of attaching the bio-logger to the organism.}\label{summarytable}
\end{table}

In addition to the time series bio-logger data, each dataset in BEBE comes with human-generated behavioral annotations (Figure~\ref{behavior_examples}, colored bars). Seven datasets were ground-truthed using either on or off tag video; the other two were ground-truthed using audio. In each dataset, each sampled time step is annotated with the current behavioral state of the tagged individual, which can be one of several discrete behavioral classes. At some time steps, it was not possible to observe the individual, or it was not possible to classify the individual's behavior using the predefined behavioral classes. In these cases, this time step is annotated as \textit{Unknown}. These \textit{Unknown} behavioral annotations are disregarded during model training and evaluation. 

There are multiple time scales of behavior represented across the nine ethograms in BEBE, with some datasets including brief activities (e.g. shaking), and some including longer duration activities (e.g. foraging). In Table~\ref{summarytable} we report the mean duration (in seconds) of an annotation in each dataset, as a rough estimate of the mean duration an individual spends in a given behavioral state.

For previously published datasets, the intentions were to validate an ethogram for use in free-ranging individuals (Polar Bear, Seals), use the ethogram to understand activity patterns (Whale, Sea Turtle, Rattlesnake), develop on-device algorithms to detect a specific rare behavior of interest (Gull), or provide a publicly available dataset (HAR, Dog). The data used for annotation also varied from on-sensor (Polar Bear, Whale, Sea Turtle, Gull, Crow) to off-sensor (HAR, Seals, Rattlesnake, Dog). Given the range of purposes and data collection methods, the ethograms vary in how much of the animal’s time is accounted for and how fine-grained the behavior categories are.

In addition to Figure~\ref{behavior_examples}, we provide additional data visualizations in the Supplemental Information. Examples of each behavior class with the full set of channels reveal varying degrees of stereotypy in the behavioral classes (Supplemental Figures~\ref{HAR_examples_supplement}-\ref{dog_examples1_supplement}). For example, in the Sea Turtle dataset, \textit{Stay at surface} appears more stereotyped than \textit{Feed}. Summary statistics across different individuals for each behavior class suggest the presence of discriminative features for some datasets, as well as differences between individuals (Supplemental Figures~\ref{crow_histogram_supplement}-\ref{dog_histogram_supplement}). For example, in the Rattlesnake dataset, \textit{Move} shows higher variance in raw accelerometer values compared to \textit{Not Moving}, although to different degrees in different individuals. 

\subsubsection{Dataset collection}
\label{si-dataset-collection}

Datasets had to meet the following criteria to be included in BEBE:
\begin{enumerate}
\item Include animal motion data recorded by tri-axial accelerometer at $\ge 1$ Hz;
\item Include annotations of animal behavioral states;
\item Comprise data recorded from tags attached to at least five individuals in order to reflect variation in sensor placement and individual motion patterns;
\item Contain over 100000 sampled time steps with behavioral annotations;
\item Contribute to a diversity of taxa, as well as a balance among the categories of terrestrial, aquatic, and aerial species;
\item Have previously appeared in a peer-reviewed publication (with the exception of the Crow dataset, which is previously unpublished and described in more detail below);
\item Be licensed for modification and redistribution; or come with permission from dataset authors for modification and public distribution.
\end{enumerate}

Four datasets were not previously publicly available and were collected by coauthors (Whale: A. Friedlaender; Crow: D. Canestrari, V. Baglione, V. Moreno-González, E. Trapote; Gull: T. Maekawa, K. Yoda; Rattlesnake: D. DeSantis, V. Mata-Silva). For these datasets, coauthors provided permission to publicly distribute the data. Through an informal literature search, we found five publicly available datasets (HAR, Polar Bear, Dog, Sea Turtle, Seals). Of these, four were collected by coauthors (Polar Bear: A. Pagano; Dog: O. Vainio, A. Vehkaoja; Sea Turtle: L. Jeantet, D. Chevallier; Seals: M. Ladds). Finally, we assessed datasets from papers covered by a recent systematic literature review of automatic behavioral classification from bio-loggers \cite[Page 12]{thiebaultAnimalborneAcousticData2021}. The supplemental material of \cite{thiebaultAnimalborneAcousticData2021} provides a table with the results of their systematic review, containing metadata on whether a paper used supervised learning, species, number of individuals, and number of timepoints. We looked exclusively at the supervised learning papers because these would require annotated datasets (criterion 2). Assessing criteria 1, 3, and 4 above resulted in twelve potential datasets out of 214. Of the twelve, two were already included in BEBE (Rattlesnake, Sea Turtle), nine studied terrestrial animals, a category which was already well-represented in BEBE, and one did not provide annotations. Therefore, no new datasets were added based on the results of the systematic literature review by \cite{thiebaultAnimalborneAcousticData2021}.

\subsubsection{Dataset summaries}

\label{dataset summary section}

In the following, we summarize the study design and data collection protocols for each of the datasets in BEBE. See also Table~\ref{summarytable}.

\paragraph{Human Activity Recognition (HAR)}

The study~\cite{anguitaPublicDomainDataset2013} was designed to provide a publicly available dataset of human (\textit{Homo sapiens}) activities recorded by smartphone tri-axial accelerometers and gyroscopes. Thirty human subjects were instructed to perform a sequence of activities (Walking, Standing, Sitting, Lying Down, Walking Upstairs, and Walking Downstairs) while wearing a waist-mounted Samsung Galaxy S II smartphone. Behaviors were annotated based on video footage, and no information is provided about synchronization between video and motion sensor data. The ethogram used covers all behaviors performed by individuals, except transition periods between activities (e.g. moving from sitting to standing) which are treated as Unknown.

\paragraph{Rattlesnake}

The study~\cite{desantisIntegrativeFrameworkLongTerm2020} sought to to quantify and evaluate variation in long-term activity patterns in free-ranging western diamondback rattlesnakes (\textit{Crotalus atrox}) in the Indio Mountains Research Station, located in Texas. Individuals were implanted with Technosmart AXY-3 or Technosmart AXY-4 tri-axial accelerometers. Their behavior was directly observed, and recorded with a hand-held video camera. All recorded time steps were assigned to one of two behavior categories, Moving and Not Moving, in order to accommodate low-frequency recording and maximizing recording duration. Annotations were made based on field notes and recorded video.

\paragraph{Polar Bear}

The study~\cite{paganoUsingTriaxialAccelerometers2017, paganoMetabolicRateBody2018} was designed to validate the usage of tri-axial accelerometers and conductivity sensors, for the purpose of constructing daily activity budgets of polar bears (\textit{Ursus maritimus}) on sea ice in the Beaufort Sea (Arctic Ocean). Bears were captured and equipped with Exeye video collars that also contained Wildlife Computers TDR10-X-340D motion loggers. Collars were retrieved after they fell off, or after the bear was re-captured. Motion data was annotated based on synchronized video footage. The ethogram was designed to cover all common behaviors observed in this footage, but excluded behaviors that were rare (such as fighting, breeding, drinking), extremely brief, or nondescript. Excluded behaviors were marked as Unknown. Additionally, the video camera was set to a 90-second duty cycle. Behaviors were marked as Unknown during time periods where the video camera was off.

\paragraph{Dog} The study~\cite{kumpulainenDogBehaviourClassification2021, vehkaojaDescriptionMovementSensor2022} was intended to provide a dataset for developing methods that could be used to classify domestic dog (\textit{Canis familiaris}) behaviors. Dogs were equipped with two ActiGraph GT9X Link loggers. One was placed on the neck using a collar, and the other was placed on the back using a harness. In an indoor arena, dogs were guided by their owners through a series of activities: sitting, standing, lying down, trotting, walking, playing, and treat-searching. Behaviors were annotated based on synchronized video footage. The ethogram was designed to reflect all the behaviors commonly performed during these activities. Ambiguous behaviors were recorded as Unknown. 

\paragraph{Whale} The study~\cite{friedlaenderExtremeDielVariation2013} characterized daily activity budgets of humpback whales (\textit{Megaptera novaeangliae}) in Wilhelmina Bay, Antarctica, late in the feeding season. DTAG devices~\cite{johnson2003digital} were attached via suction cups to whales' dorsal surface or flank. They were programmed to release suction after 24 hours, and were retrieved after release. The ethogram was designed to include common diving behavior, as well as resting. To identify different behaviors, whales' feeding lunges were first detected using an algorithm, based on recorded acoustic flow noise, as well as changes in the accelerometer signal. Then, dives were classified based on maximum dive depth, duration, and the presence and number of feeding events. These annotations were reviewed by two of the original study authors.

\paragraph{Sea Turtle} The study~\cite{jeantetBehaviouralInferenceSignal2020} aimed to develop a machine learning method to compute activity budgets for green turtles from accelerometer and gyroscope data. CATS devices (Customized Animal Tracking Solutions, Germany) were attached using suction cups to the carapaces of free-ranging immature green turtles (\textit{Chelonia mydas}) in Martinique, France. Behaviors were annotated based on synchronized on-device video footage. To design the ethogram, forty-six behaviors were initially identified in video footage. From these, seven frequent behavior categories were identified, and the remaining behaviors (such as regurgitation, pursuit of other turtle) grouped into an `other' category. The `other' category was considered Unknown in the present study. 

\paragraph{Seals} The studies aimed to validate machine learning methods for behavior classification in otariids on captive seals, intended for eventual application in wild seals. Tags including CEFAS G6a+ accelerometers were attached to the backs of captive fur seals (\textit{Arctocephalus forsteri} and \textit{Arcocephalus tropicalis}) and sea lions (\textit{Neophoca cinerea}). Behaviors were filmed in a swimming pool, by two or three underwater cameras (GoPro Hero 3 – Black edition) and one handheld camera above water (Sony HDRSR11E). In observation sessions, individuals either received a food item or were requested to perform behaviors learned through operant conditioning. The requested behaviors were chosen to reflect behaviors performed by wild seals. Twenty-six behaviors were identified in the video footage. Based on prior knowledge of wild seals, these initial behaviors were grouped into four categories or `other' (such as direct feeding by trainer, seal out of sight of camera). The `other' category was considered Unknown in the present study

\paragraph{Gull} The study aimed to develop on-device machine learning algorithms to detect rare, ecologically important behaviors in accelerometer data (e.g., foraging) in order to control resource-intensive sensors like video cameras. The bio-loggers included a tri-axial accelerometer (TDK MPU-9250; InvenSense), integrated video camera, as well as other low-cost sensors which were not used in this study. Using waterproof tape and teflon harnesses, these were attached to either the back or abdomen of free-ranging black-tailed gulls (\textit{Larus crassirostris}) in a colony on Kabushima Island near Hachinohe City, Japan. The main behavior of interest in this study was foraging, which included a variety of behaviors such as surface-dipping and plunging. Two other common non-foraging behaviors (flying and stationary)  were also labeled. Behaviors were labeled based on video footage collected by the on-device video camera. 

\paragraph{Crow}
The Crow dataset is presented for the first time here, and we provide complete details in the following section. 

\subsection{Crow dataset details} The data logger, called miniDTAG, was adapted from a 2.6 g bat tag integrating microphone, tri-axial accelerometer and tri-axial magnetometer \cite{stidsholt2019tag} with changes that enable long duration recordings on medium-sized birds. The triaxial accelerometer (Kionix KX022-1020 configured for ± 8 g full scale, 16-bit resolution) was sampled at 1000 Hz and decimated to a sampling rate of 200 Hz before saving to a 32 GB flash memory. The 1.2 Ah lithium primary battery (Saft LS14250) allowed continuous recording for about 6 days both in lab and field settings. Each miniDTAG was packaged with a micro radio transmitter (Biotrack Picopip Ag376) and attached to the two central tail feathers of carrion crows (\textit{Corvus corone}) with a piece of the stem of a colored balloon following the procedure described in \cite{rutz2013programmable} (axes: $x$ (backward-forward), $y$ (lateral), $z$ (down-up)). The thin rubber balloon material progressively deteriorated and finally broke, letting the miniDTAG fall to the ground, where it was radio-tracked using a Sika Biotrack receiver. 

Accelerometer data were calibrated using Matlab tools from www.animaltags.org following standard procedures \cite{johnson2003digital,martin2016tracking}. The sensor channel was decimated by a factor of 4 before calibration, resulting in a sampling rate of 50 Hz. We normalized the tri-axial acceleration channels so the average magnitude of the acceleration vector was equal to $1$.

For the present study, we tagged 11 individuals  from 7 different territories near Le\'on, Spain, in a population that breeds cooperatively. Here crows live in stable kin group, in which a dominant breeding pair is assisted by subordinate helpers in raising the young~\cite{baglione2002cooperatively}. Of these individuals, three were breeding males, two were helper males, five were breeding females, and one was a helper female, and all were attending an active nest. Data were collected in spring 2019, when all the birds were raising their nestlings. The miniDTAG plus battery (12.5 g) accounted on average (± SE) for the 2.66 ± 0.09\% of the crow body mass (range 2.29 – 3.15\%). None of the crows abandoned the territory or deserted the nest after being tagged. From the recordings of these individuals, we selected 20 contiguous segments for annotation (average segment duration: 5.73 hours), favoring segments where begging vocalizations and wing beats could be identified at multiple times during the recording (see Annotations below).

We divided the recorded data into five-second long non-overlapping segments (\textit{clips}). Each accelerometer clip came with synchronized audio, which we used to assign behavioral annotations. If there were sounds of wingbeats for the entire duration of a clip, we annotated all sampled time steps in that clip as \textit{Flying}. Additionally, if there were wingbeats followed by wind noise (interpreted as soaring), we also annotated all sampled time steps in that clip as \textit{Flying}. Similarly, if a clip included sounds of chick begging calls, and no sounds of wingbeats, we annotated all sampled time steps in that clip as \textit{In Nest}. Therefore, the label \textit{In Nest} likely encompasses several behavioral states, such as resting, brooding, incubating, feeding chicks, and preening, which may occur at or near the nest. Crucially, these states do not include flying, and so there is no ambiguity between the two labels. Clips that did not fit either of the criteria for being labeled \textit{Flying} or \textit{In Nest} were labeled as \textit{Unknown}.

The two behaviours we chose for our ethogram (\textit{Flying} and \textit{In Nest}) are highly relevant for ethology research: Individual chick provisioning effort, measured as frequency of nest visits, is one of the key variables in the study of parental care behaviour in this species. It may be possible to infer nest visits based on alternating periods of \textit{Flying} and \textit{In Nest}. Counting nest visits typically requires either many hours of direct observations, which is time-consuming and difficult to carry out without interfering with the animals, or using video cameras at the nest, which requires costly equipment, high effort to install, and daily visits to change the battery. 

\subsubsection{Data pre-processing}

For full implementation details, we refer the reader to the dataset preprocessing source code\footnote{https://github.com/earthspecies/BEBE-datasets/}. For all datasets, we used calibrated data and annotations provided by the original dataset authors; with the exception of the Crow dataset (described above and below), we refer the reader to the original publications for details. For two datasets (Sea Turtle, Gull), the average magnitude of the acceleration vector varied by more than 10\% between tag deployments. To control for these differences, we normalized the tri-axial acceleration channels so that the average magnitude of the acceleration vector was equal to $1$. For the Gull dataset, there were two possible tag placement positions (back or abdomen). To reduce data heterogeneity due to differences in tag placement, we rotated the calibrated data from some deployments by 180 degrees, around the axis parallel to the tagged individual's body. After performing this step, all deployments had, on average, positive acceleration in the vertical axis. While this step reduced heterogeneity between deployments, is unlikely to remove all differences between the data recorded by these different tag placements. We do not perform any additional special pre-processing steps on the datasets in BEBE, and we left each dataset in its original measurement units. 

\subsubsection{Annotations} 

In all datasets in BEBE, annotations indicated time intervals when the behavior class occurred (even when this time interval is only a few seconds long), rather than the occurrence of discrete behavioral events. For example, multiple discrete feeding events could occur within a time interval labeled as ‘foraging’. The modeling task (per-time step classification) and evaluation procedure (per-time step classification precision and recall) are designed with this in mind. These annotations would be those required to describe the amount of time an animal spends performing different activities during a day.

With the exception of one dataset (Crow), the annotations in BEBE are derived from annotations made in the original studies. As a result, datasets in BEBE are annotated in a variety of ways (Table ~\ref{summarytable}, row \textit{Annotation method}). Datasets also vary in the specificity of their behavior classes: a behavior class may include several related behaviors and datasets vary in how much behaviors are grouped or split. For example, in the Dog dataset, there are fine distinctions between different behavioral classes (e.g. \textit{Sit} vs. \textit{Pant+sit}), relative to the Rattlesnake dataset, in which a behavior is only summarized as \textit{Moving} or \textit{Not Moving}.

For the remaining eight datasets, we used annotations as provided by the original dataset authors. For behaviors with few annotated timepoints in the original dataset, we treat these behaviors as \textit{Unknown} (see dataset pre-processing source code for details). 
For all datasets, we used the time alignment between annotations and tag data that were produced by the original dataset authors.

\subsection{Task and model evaluation}
\label{model evaluation}

\begin{figure}[htbp]
\includegraphics[width=\textwidth]{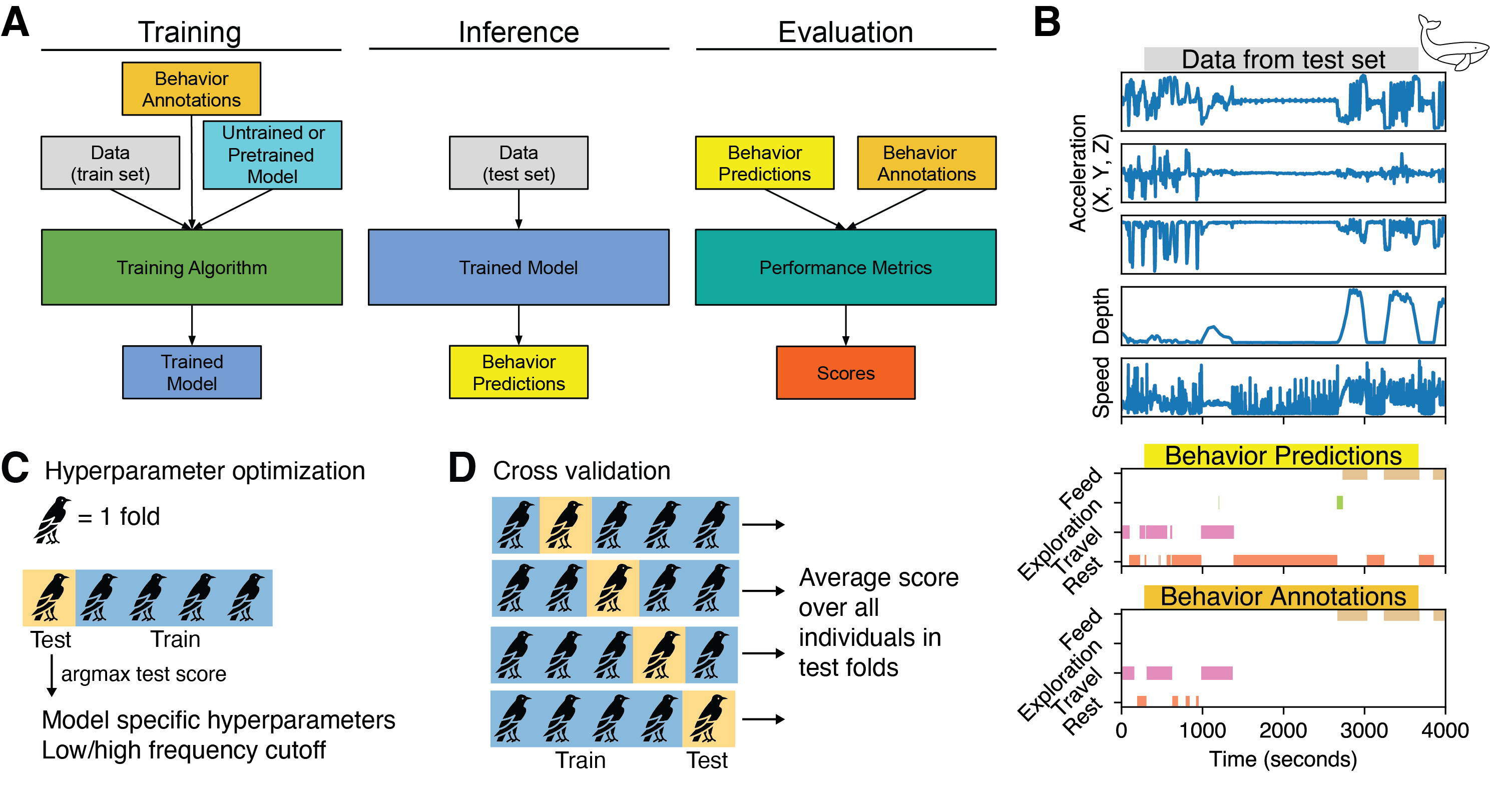}
\caption{A) Summary of training and evaluation. Our process of data analysis follows the standard three steps of creating and evaluating machine learning models.  In the first step (Training), the model learns from the train set of one dataset, including behavioral annotations. In the second step (Inference), the model makes predictions about the behavioral annotation for the test set data, which comprises data from a set of individuals distinct to those in the train set. In the third step (Evaluation), the model's predictions are evaluated based on their agreement with known behavioral annotations. B) Example data from the Whale dataset~\cite{friedlaenderExtremeDielVariation2013}, and predictions made by a \texttt{CRNN} model. The trained model is fed raw time series data, which it uses to make behavior predictions. These predictions are compared with annotations to arrive at performance scores. In this case, the model predicts the annotations well. Gaps in the behavior annotations indicate the behavior is \textit{Unknown} at those samples; those samples are ignored in the evaluation metrics. C) During hyperparameter optimization, we train a set of models with various hyperparameters and low/high frequency cutoffs. We obtain the model hyperparameters and low/high frequency cutoff from the model that maximizes the F1 score on the first test fold. D) During cross-validation, we compute the test scores for the other four folds. The final score is averaged across all individuals in the test folds. The first test fold, used for hyperparameter optimization, is not used for testing.}
\label{evaluation_summary}
\end{figure}

We provide a standard method for measuring different models' ability to classify behavior from bio-logger data (Figure~\ref{evaluation_summary}) that consists of a formal task, as well as a set of evaluation metrics. It reflects the following workflow. First, the researcher has defined the ethogram categories of interest and annotated the dataset. Then, the annotated dataset is split between the train set, used to train ML model, and the test dataset, which is used to evaluate the performance of the trained model. 

Trained models are evaluated on their ability to predict behavior annotations. For each individual, we measure classification precision, recall, and F1 scores averaged across all sampled time steps from that individual and averaged across all behavioral classes. We disregard the time steps for which the annotation is \textit{Unknown}. The entire pipeline, including training, inference, and evaluation, is repeated for each dataset in BEBE.

We split each dataset into five groups, or \textit{folds}, so that no individual appears in more than one fold~\cite{ferdinandy2020challenges}. For evaluation, we use a cross validation procedure. During cross validation, we train a model on the individuals from four folds, and test it on the individuals from the remaining fold. For all datasets, Figure~\ref{supplement_class_representation} shows the proportion of behavior classes for each fold. This data partition reflects a common use case where researchers train a model from one set of individuals and then apply it to a separate unseen set of individuals (e.g., when behavior labels cannot be manually assigned for the latter). 

\subsubsection{Behavior classification task}
\label{supervised task}

Each dataset consists of a collection of multivariate discrete time series, where each time series $\{\mathbf{x}_t\}_{t \in \{1,2,\dots, T\}}$ consists of samples $\mathbf{x}_t \in \mathbb{R}^D$. Here $D$ is the number of data channels and $T$ is the number of sampled time steps. Note that the number $T$ may vary between different time series contained in a single dataset. Each time series is sampled from one bio-logger deployment attached to one individual and is sampled continuously at a fixed dataset-specific sampling rate (Table~\ref{summarytable}).

Each time series in a dataset also comes with a sequence of annotations $\{l_t\}_{t \in \{1,2,\dots, T\}}$, where each $l_t \in \{\textit{Unknown}, c_1, c_2, \dots, c_C\}$ encodes either the behavioral class $c_j$ of the animal at time $t$, or the fact that the behavioral class is \textit{Unknown}. Here $C$ denotes the number of known behavioral classes in the dataset. The behavioral classes $c_j$ vary between datasets in BEBE, and could be e.g. $c_j = \textit{Foraging}, c_j = \textit{Sniffing},$ or $c_j = \textit{Flying}.$

The behavior classification task is to predict the behavioral annotation $l_t$ of each sampled time step $\mathbf{x}_t$ (Figure~\ref{evaluation_summary}B). 
During training, models are given access to the behavioral annotations in the train set. We refer the reader to~\cite{wangMachineLearningInferring2019} for a review of studies with a similar task description. While behavior classification can also be formulated as a continuous time problem~\cite{McClintockDiscrete2014}, we focus on a discrete time problem formulation in order to match the majority of prior studies.

\subsubsection{Dataset Splits} 

A key part of a benchmark dataset is how it partitions the data used for model training (the \emph{train set}) from the data used for model evaluation (the \textit{test set}). This evaluation provides an estimate of how well a model performance generalizes outside of its train set. Therefore, the specific partition chosen determines what domains the ML model should generalize over.

In BEBE, we split each dataset into five groups (\emph{folds}), which are used in a cross validation procedure. During cross validation, each time the model is trained, the train set consists of the data from four of these five folds, and the test set consists of the data from the remaining fold. For each dataset in BEBE, we divided the data so that no individual appears in more than one fold, and so that each fold has the same number of individuals represented ($\pm 1$ individual). Therefore, during testing, a model's performance reflects its ability to generalize to new individuals, where effects such as tag placement~\cite{garde2022ecological} may influence model predictions.

Figure~\ref{supplement_class_representation} displays the distribution of annotations across folds for all datasets in BEBE. Most datasets in BEBE have some behaviors with high representation (up to 92.4 percent of known behaviors, Rattlesnakes \textit{Not Moving}), and some behaviors with very low representation (as little as 0.1 percent of known behaviors, Polar Bears \textit{Pounce}).

\subsubsection{Evaluation Metrics} Trained models are evaluated on their ability to predict the behavioral annotations of the test set. For each individual in the test set, we measure macro-averaged precision, recall and F1 scores of model predictions. By macro-averaging, performance on each behavioral class is weighted equally in the final metrics, regardless of their relative proportions in the test set. Finally, we average these scores across all individuals in the test set. 
In measuring these scores, we disregard the model's predictions for those time steps $\mathbf{x}_t$ for which $l_t = \textit{Unknown}$. More precisely, for each individual in the test set we measure:
\begin{equation} \label{metrics}
\operatorname{Prec} = \frac{1}{C} \sum_{j = 1}^{C} \operatorname{Prec}_j, \quad \operatorname{Rec} = \frac{1}{C} \sum_{j = 1}^{C} \operatorname{Rec}_j, \quad \operatorname{F1}= \frac{1}{C} \sum_{j = 1}^{C} \operatorname{F1}_j,
\end{equation}
where for each behavioral class index $j\in \{1,\dots, C\},$
\[
\operatorname{Prec}_j = \frac{\operatorname{TP}_j}{\operatorname{TP}_j + \operatorname{FP}_j}, \quad
\operatorname{Rec}_j = \frac{\operatorname{TP}_j}{\operatorname{TP}_j + \operatorname{FN}_j}, \quad
\operatorname{F1}_j = 2\cdot \frac{\operatorname{Prec}_j \cdot \operatorname{Rec}_j}{\operatorname{Prec}_j + \operatorname{Rec}_j}.
\]
Here, $\operatorname{TP}_j, \operatorname{FP}_j,$ and $\operatorname{FN}_j$ denote, respectively, the number of sampled time steps correctly predicted to be of class $c_j$ (true positives), the number incorrectly predicted to be of class $c_j$ (false positives), and the number incorrectly predicted to be not of class $c_j$ (false negatives). Precision, recall, and F1 range between 0 and 1, with 1 reflecting optimal performance. After computing these scores for each individual, we calculate the average taken across all individuals in the test set. In addition to precision, recall, and F1 score, we compute confusion matrices for model predictions (see examples in Figures~\ref{supplement_confusion_matrices_crnn} and~\ref{supplement_confusion_matrices_rf}, with full set available at \url{https://zenodo.org/records/7947104}).

\subsubsection{Hyperparameter Tuning and Cross Validation}
\label{xval section}
All models we tested require the user to choose some parameters (known as \textit{hyperparameters}) before training. To select hyperparameters for a given type of ML model and dataset, we performed an initial grid search across a range of possible values, using the first fold of the dataset as the test set and the remaining four folds of the dataset as the train set. We saved the hyperparameters which led to the highest $\operatorname{F1}$ score, averaged across individuals in the test set. The hyperparameter values included in the grid search are specified below, and the hyperparameter values that were saved for subsequent analyses are available at \url{https://github.com/earthspecies/BEBE}.


While it is common in the field of ML to use a single fixed train/test split of a dataset, we chose to use cross validation in order to capture the variation in motion and behavior between as many individuals as possible. After the initial hyperparameter grid search, we used the saved hyperparameters to train and test a model using each of the remaining four train/test splits of the dataset (which were not used for hyperparameter tuning). The final scores (precision, recall, and F1) we report are averaged across individuals taken from these four train/test splits.

All of the models we trained involve some randomness in the training process, which can introduce variance into model performance~\cite{bouthillierAccountingVarianceMachine}. In addition, model performance varies between different individuals. Understanding the magnitude of this variation may be important when applying these techniques in new contexts.

To quantify variation in model test performance, for each model type we compute the standard deviation of each performance metric, taken across all individuals represented in the four test folds of the dataset that were not used for hyperparameter tuning. These values, given in parentheses in Figures~\ref{basic_results} and~\ref{representation_learning_results}, therefore reflect variation in these scores due to differences in individual motion, as well as due to sources of variation in model training. 

We do not perform significance tests using the variance in performance metrics computed through cross validation. In cross validation, data are reused in different train sets. The resulting metrics therefore violate the independence assumptions of many statistical tests, leading to underestimates in the likelihood of type I error~\cite{Dietterich1998ApproximateST}. Bootstrapping can produce better estimates of variance in model performance, but this involves high computational investment which may discourage future community use of a benchmark~\cite{bouthillierAccountingVarianceMachine}. Therefore, as is typical for ML, we report variance in model performance in order to give a sense for its magnitude.

\subsubsection{Low and high frequency components of acceleration}

A common technique in analysis of acceleration data is to isolate acceleration due to gravity using a high- or low-pass filter~\cite{shepardIdentificationAnimalMovement2008}, resulting in separate static and dynamic acceleration channels. It has been shown that the choice of cutoff frequency can have a strong effect on subsequent analyses~\cite{MartnLpez2020OverallDB}. Often, this frequency is chosen based on expert knowledge of an individual's physiology and typical movement patterns. As an alternative data-driven approach, we treated the cutoff frequency as a hyperparameter to be selected during model training. We use the terms "high frequency component" and "low frequency component" instead of "dynamic component" and "static component", to reflect that in this data-driven approach, the cutoff frequency that leads to the best classification performance might not match the cutoff frequency that would isolate the acceleration due to gravity.

In more detail, for each raw acceleration channel, we apply a high-pass delay-free filter (using a linear-phase (symmetric) FIR filter with a Hamming window, followed by group delay correction) to obtain the high frequency component of the acceleration vector \cite{animaltagtoolbox}. The high frequency component is then subtracted from the raw acceleration to obtain the low frequency component of the acceleration vector. The separated channels are then passed on as input for the rest of the model. For each dataset, the specific cutoff frequencies we selected from were 0 Hz (no filtering), 0.1 Hz, 0.4 Hz, 1.6 Hz, and 6.4 Hz. We omitted this step in the Rattlesnake dataset, where the high frequency components of the data had already been isolated. We also omitted this step for \texttt{harnet} models, since they were pre-trained using raw acceleration data, and for models using wavelet features, since the wavelet transform already decomposes a signal into different frequency components. For the experiments in Section~\ref{selfsupervised_lowdata_section} (Figure~\ref{representation_learning_results}), we used the cutoff frequency selected during the full data experiments in Section~\ref{deep_networks_win} (Figure~\ref{basic_results}).

\subsection{Model Implementation and Training Details}

The methods we compared included the classical ML models Random Forests (\texttt{RF}), Decision Tree (\texttt{DT}), and Support Vector Machine (\texttt{SVM}), which are widely used to classify behavior recorded by bio-loggers~(reviewed in \cite[Table 2]{thiebaultAnimalborneAcousticData2021, wangMachineLearningInferring2019}). These methods make predictions based on a set of pre-computed summary statistics, also known as \textit{features}. For each classical method, we compared two different feature sets. The first feature set (denoted \texttt{Nathan}) was introduced in~\cite{nathanUsingTriaxialAcceleration2012}. The second feature set (denoted \texttt{Wavelet}) consisted of spectral features, computed using a wavelet transform, inspired by~\cite{sakamotoCanEthogramsBe2009} (see Methods).

In addition to classical methods, we compared methods based on deep neural networks that make predictions based on raw bio-logger data. We compared two types of models that are commonly used in other ML applications: a one-dimensional convolutional neural network (\texttt{CNN}), and a convolutional-recurrent neural network (\texttt{CRNN}). These types of methods have been employed by some recent studies focused on classifying animal and human behavior~\cite{bohnslavDeepEthogramMachineLearning2021, eerdekensAutomaticEquineActivity2020, ordonezDeepConvolutionalLSTM2016}. 

Additionally, we compared a convolutional-recurrent neural network which had been pre-trained with self-supervision, using human wrist-worn accelerometer data (\texttt{harnet}~\cite{yuan2022selfsupervised}). This network was pre-trained using over 700,000 days of un-annotated human wrist-worn accelerometer data recorded at 30 Hz~\cite{yuan2022selfsupervised}. The model was trained to predict whether these data had been modified, for instance by changing the direction of time, or by permuting the channels (Figure~\ref{representation_learning_results}A). We adapted this pre-trained model to our behavior classification task, which required small modifications of the network architecture (Figure~\ref{representation_learning_results}B; see details below). We refer to this modified model as \texttt{harnet frozen} (or \texttt{harnet} for short). 

Models were implemented in Python 3.8, using PyTorch 1.12~\cite{NEURIPS2019_bdbca288} and scikit-learn 1.1.1~\cite{scikit-learn}. We used a variety of computing hardware depending on their availability through our computing platform (Google Cloud Platform). Deep neural networks (\texttt{CNN}, \texttt{CRNN}, \texttt{harnet}) used GPUs, and the rest of the models used CPUs. Our pool of GPUs included NVIDIA A100 and NVIDIA V100 GPUs. A single GPU was used to train each model. Our pool of CPUs included machines with 16, 32, 64, 112 and 176 virtual CPUs.

For full implementation details, we refer readers to the source code\footnote{https://github.com/earthspecies/BEBE/}, which also contains the specific configurations that were evaluated during hyperparameter optimization. For all models we trained, we weighted the loss associated with each behavior class in inverse proportion to the frequency that that behavior occurred in the data; in initial experiments we found that this method for accounting for differences in behavior representation improved classification performance.
For all the experiments presented, we trained over 2500 models for the purpose of hyperparameter tuning. For the hyperparameters that were then selected and used to obtain the reported results, we refer readers to our dataset repository \footnote{https://zenodo.org/record/7947104}.

\subsubsection{Supervised Neural Networks}

All neural networks were implemented in PyTorch, with model-specific details given below. For each dataset in BEBE, we trained each type of model for 100 epochs using the Adam optimizer~\cite{kingmaAdamMethodStochastic2015}. In each epoch, we randomly chose a subset of contiguous segments (\textit{clips}) to use for training. The number of clips chosen per epoch was equal to twice the number of sampled time steps, divided by the clip length in samples. The clip length varied between datasets, and is specified below.

We used categorical cross-entropy loss, weighted in proportion to annotation imbalance. We applied cosine learning rate decay~\cite{loshchilovSGDRStochasticGradient2017}, and a batch size of 32. We masked all loss coming from sampled time steps annotated as \textit{Unknown}.

\paragraph{\texttt{CNN} and \texttt{CRNN} models}

\texttt{CNN} consists of two dilated convolutional layers, a linear (i.e. width-1 convolution) prediction head, and a softmax layer. \texttt{CRNN} consists of two dilated convolutional layers, a bidirectional gated recurrent unit (GRU), a linear prediction head, and a softmax layer. In both \texttt{CNN} and \texttt{CRNN}, all convolutional layers are followed by ReLU activations and batch normalization. Each convolutional layer has 64 filters of size 7, and the GRU layer has 64 hidden dimensions. The outputs of these models are interpreted as class probabilities. \texttt{CRNN} models for behavior classification have previously been used by \cite{otsuka2024exploring,AulsebrookQuantifyingMating2024}.

 We used a default clip length of 2048 samples (the time this represents will vary with the sampling rate of the dataset). However, two datasets include some deployments with fewer than 2048 recorded samples. For these, we used a shorter clip length (Rattlesnake, 64 samples; Seals, 128 samples).

For our initial hyperparameter grid search, learning rate was selected from $\{1 \times 10^{-2}, 3\times 10^{-3}, 1 \times 10^{-3}\},$ and convolutional filter dilation was selected from $\{1, 3, 5\}$.

\paragraph{\texttt{Harnet} and its variations}

For \texttt{harnet}, we use the pre-trained model described in~\cite{yuan2022selfsupervised}, and obtained model weights from \href{https://github.com/OxWearables/ssl-wearables}{https://github.com/OxWearables/ssl-wearables}. For a description of the pre-training setup using 700,000 hours of un-annotated data from human wrist-worn accelerometers, see Figure~\ref{representation_learning_results} and~\cite{yuan2022selfsupervised}. As a default, we used the weights from the pre-trained model \texttt{harnet30} available at this repository. For the Rattlesnake and Seals datasets, we used the pre-trained model \texttt{harnet5} instead, since it was pre-trained on shorter duration clips. Both \texttt{harnet30} and \texttt{harnet5} consist of a sequence of convolution blocks, following~\cite{heDeepResidualLearning2015}. Each pair of convolution blocks is separated by a pooling operation, which includes downsampling in the time domain. For our \texttt{harnet} model, we used the outputs of the second convolution block (of either \texttt{harnet30} or \texttt{harnet5}), which had been downsampled by a factor of 4. These outputs were then upsampled to their original temporal duration, before being passed through a bidirectional GRU with 64 hidden dimensions, and finally a linear (i.e. width-1 convolution) prediction head. The weights of the GRU and linear layers were randomly initialized before training. 

Because the model trained in~\cite{yuan2022selfsupervised} operated on tri-axial accelerometer data, we only passed tri-axial accelerometer data through the convolutional blocks. To match the settings of the original pre-training setup, we normalized the acceleration channels so that the average magnitude of the acceleration vector was equal to $1$. Additional data channels (depth, conductivity, and speed) were appended before being passed into the GRU. We omitted gyroscope channels; the data in these channels were relatively complex and outside the scope of the work of~\cite{yuan2022selfsupervised}. For our \texttt{harnet} model, we froze all weights during training except those in the GRU and prediction head. Our model \texttt{harnet unfrozen} was identical to \texttt{harnet}, except we did not freeze any weights during training.

We evaluated several ablations of \texttt{harnet} (for results, see Figure~\ref{representation_learning_results}). The first, \texttt{harnet random}, has the same architecture as \texttt{harnet} and \texttt{harnet unfrozen}, except we used randomly initialized weights instead of the weights obtained by the pre-training procedure of~\cite{yuan2022selfsupervised}. This ablation was intended to disentangle the effect of pre-training from the effect of using this particular model architecture. The weights of \texttt{harnet random} were all unfrozen during training.

The second ablation, \texttt{RNN}, omits the convolutional layers and passes the raw data directly into a GRU and linear prediction head. This ablation was intended to confirm that the improved performance of \texttt{harnet} was not due to the RNN architecture we used for the non-frozen part of the \texttt{harnet} model.

The third ablation, \texttt{RNN wavelet}, replaces the convolutional layers of \texttt{harnet} with a wavelet transform. Each data channel is transformed using a Morlet wavelet transform, with 15 wavelets, using the {\tt scipy.signal.cwt} module~\cite{scikit-learn} (see details below). 
These wavelet transformed features are then passed into GRU and linear layers, as in our \texttt{harnet} model. This ablation was intended to confirm that the improved performance of \texttt{harnet} could not be matched by computing spectral features, which would not require a complicated pre-training step.
We tuned two hyperparameters for the wavelet transform, $\omega$ and $C_{max}$ which are described in the following section. We selected $\omega$ and $C_{max}$ from the same values as with the classical models.

For all the models described above, for our initial hyperparameter grid search, learning rate was selected from $\{1 \times 10^{-2}, 3\times 10^{-3}, 1 \times 10^{-3}\}$. To match the pre-training setup of~\cite{yuan2022selfsupervised}, we used a default clip length of 900 samples, and a clip length of 150 samples for Rattlesnake and Seals datasets.

\subsubsection{Features for classical models}

We tested two sets of features for the classical models (\texttt{RF}, \texttt{DT}, and \texttt{SVM}). The first set (\texttt{Nathan}) consists of summary statistics derived from \cite{nathanUsingTriaxialAcceleration2012}, which have been used or adapted in a variety of behavior classification problems (e.g.~\cite{laddsSeeingItAll2016,resheffAcceleRaterWebApplication2014}). The second set are wavelet features (\texttt{Wavelet})~\cite{sakamotoCanEthogramsBe2009}, which are commonly used to identify periodic motions like steps or tail beats (e.g.~\cite{brewsterDevelopmentApplicationMachine2018,  clarkeUsingTriaxialAccelerometer2021}).

For the \texttt{Nathan} feature set, we first defined the feature set for tri-axial accelerometer channels in the same way as \cite{nathanUsingTriaxialAcceleration2012}. For each time step $t$, we first computed the root-mean-square amplitude $q$ from the  $x,y,z$-axes, as well as the high frequency component acceleration (as described above). Then, for the $x, y, z$-axes and $q$, we computed a \textit{basic} set of features over a contiguous segment of data (\textit{clip}) centered at $t$: mean, standard deviation, skew, kurtosis, maximum, minimum, 1-sample autocorrelation, and best fit slope. We also computed three pairwise correlations between the $x,y,z$-axes, the circular variance of inclination for $q$-axis, the circular variance of azimuth for $q$-axis, and a quantity analogous to the mean overall dynamic body acceleration (ODBA) using the high frequency component acceleration. If there were any tri-axial gyroscope channels, we computed the \textit{basic} set of features as well as the pairwise correlations on the $x,y,z$-channels. For other channels (conductivity, depth and speed), we only computed the \textit{basic} features. Because the datasets included different channels, they had a different number of input features. For our initial grid search, the duration (in seconds) of the clip used for feature computation was selected from $\{0.5, 1, 2, 4, 8, 16\}$ seconds. For the Dog dataset, this duration was selected from $\{0.5, 1, 2, 4, 8\}$ seconds due to memory limitations.

For the \texttt{Wavelet} feature set, we normalized each channel independently ($z$-score), and then for each channel we followed the procedure in~\cite{sakamotoCanEthogramsBe2009}. We computed the continuous wavelet transform using \texttt{scipy.signal.cwt} with the complex Morlet wavelet function (\texttt{scipy.signal.morlet2}), using 15 wavelets per data channel. We tuned two hyperparameters for the wavelet transform. The first hyperparameter, the dimensionless $\omega$, controls the tradeoff between resolution in the time and frequency domains. The second, $C_{max}$ controls the largest wavelength (in seconds) out of the $15$ wavelets used. The parameter $\omega$ was selected from $\{5,10,20\}$, and $C_{max}$ was selected from $\{1, 10, 100, 1000\}$. Once $C_{max}$ was fixed, the wavelength of the $k^{th}$ wavelet ($k=0,1,\dots,14$) was equal to $C_{min} * \left(\frac{C_{max}}{C_{min}}\right)^{k / 14}$. The minimum wavelength, $C_{min}$, was set to $2/\operatorname{sr}$, where $\operatorname{sr}$ is the dataset-specific sampling rate.

\subsubsection{Classical models}

All classical models were implemented with Scikit-learn version 1.1.1, with model-specific details given below. We used loss functions weighted to account for annotation imbalance (corresponding to \texttt{class\_weight} = balanced). During training, we did not include any sampled time steps which were annotated as \textit{Unknown}.

\paragraph{Random Forest}

\texttt{RF} was implemented using {\tt RandomForestClassifier} from the \texttt{sklearn.ensemble}  package. For each tree we used 1/10 of the available training data (\texttt{max\_samples} = 0.1). Other than \texttt{max\_samples} and \texttt{class\_weight}, we used the default settings. The model consists of 100 decision trees.

\paragraph{Decision Tree}

\texttt{DT} was implemented using {\tt DecisionTreeClassifier} from the \texttt{sklearn.tree} package, using default settings except for \texttt{class\_weight}.

\paragraph{Support Vector Machine}

\texttt{SVM} was implemented using {\tt LinearSVC} from the \texttt{sklearn.svm} package, chosen for its scaling properties to large numbers of samples. We selected the algorithm to solve the primal optimization problem (i.e., \texttt{dual} = \texttt{False}) because \texttt{n\_samples} > \texttt{n\_features} . Other than \texttt{dual} and \texttt{class\_weight}, we used the default settings.

\section{Results}



\subsection{Deep neural networks improve classification performance}
\label{deep_networks_win}

We used the datasets and model evaluation framework in BEBE to compare different methods for predicting behavior from bio-logger data. For each dataset, we compared the performance of three classical ML models with three deep neural networks. We predicted that neural network approaches would outperform the classical approaches we tested (Table \ref{hypothesestable}, H1).
The F1 scores for these models are given in Figure~\ref{basic_results}, with precision and recall are presented in Supplemental Figure~\ref{supplement_precision_recall_basic_results}.

In terms of classification F1 score, the methods we tested that were based on deep neural networks performed the best on all nine datasets in BEBE, confirming hypothesis (H1). The top performing model was always either \texttt{CRNN} or \texttt{harnet}. The top performing deep neural net on a dataset achieved a F1 score that was .072 greater, on average, than the F1 score of the top performing classical model on that dataset. Deep neural networks achieved the best recall on all datasets and the best precision in seven out of nine datasets.

\begin{figure}[htbp]
\centering
\includegraphics[width=0.86\textwidth]{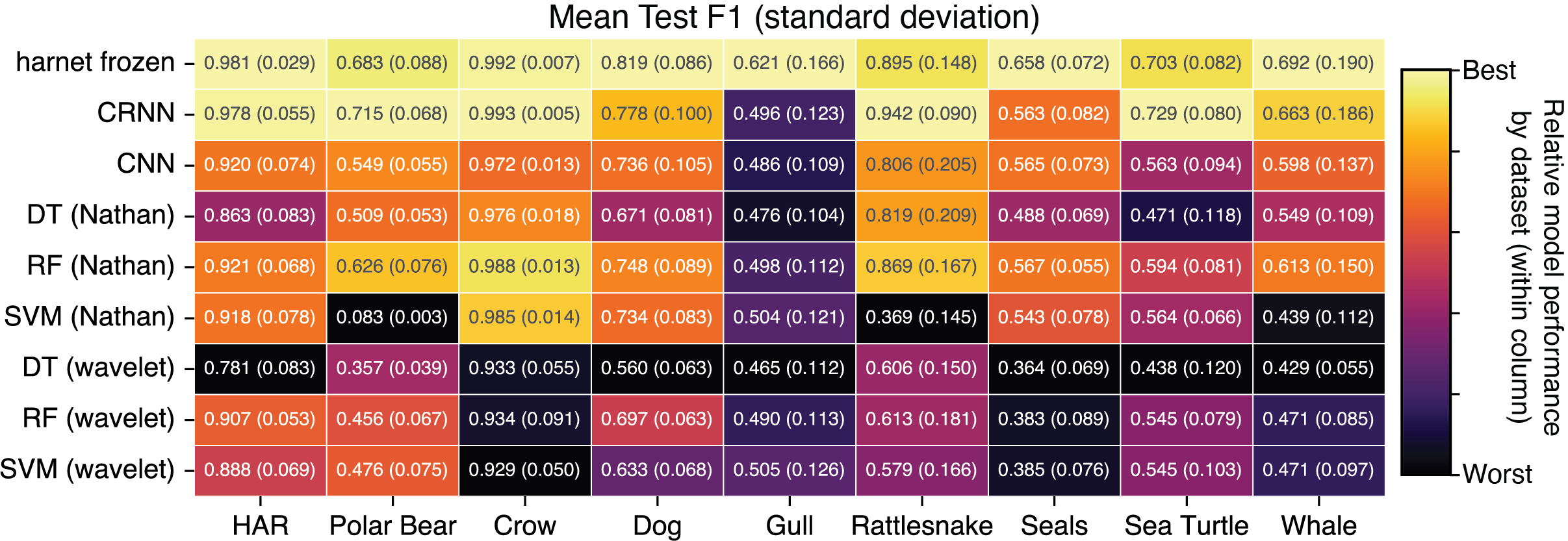}
\caption{
F1 scores on the test set for supervised task. Here and elsewhere, the table is color-coded such that within a dataset (column), the brightest color indicates the best performing model for that metric, and the darkest color indicates the worst performing model. Numbers indicate the average score across individuals in the test folds, with the standard deviation in parentheses. The F1 score is macro-averaged across classes. Out of nine datasets, \texttt{harnet} does best on five datasets for F1, as indicated by the bright yellow entries in its row. \texttt{CRNN} does best on the other four datasets. For precision and recall results, see Figure~\ref{supplement_precision_recall_basic_results}.}
 \label{basic_results}
\end{figure}

\subsection{Self-supervised pre-training enables low-data applications}
\label{selfsupervised_lowdata_section}

To evaluate the effectiveness of self-supervised pre-training for analysis of bio-logger data, we focus on the \texttt{harnet} neural network. This network was pre-trained using over 700,000 days of un-annotated human wrist-worn accelerometer data recorded at 30 Hz~\cite{yuan2022selfsupervised}.

\begin{figure}[htbp]
\centering
\includegraphics[width=0.86\textwidth]{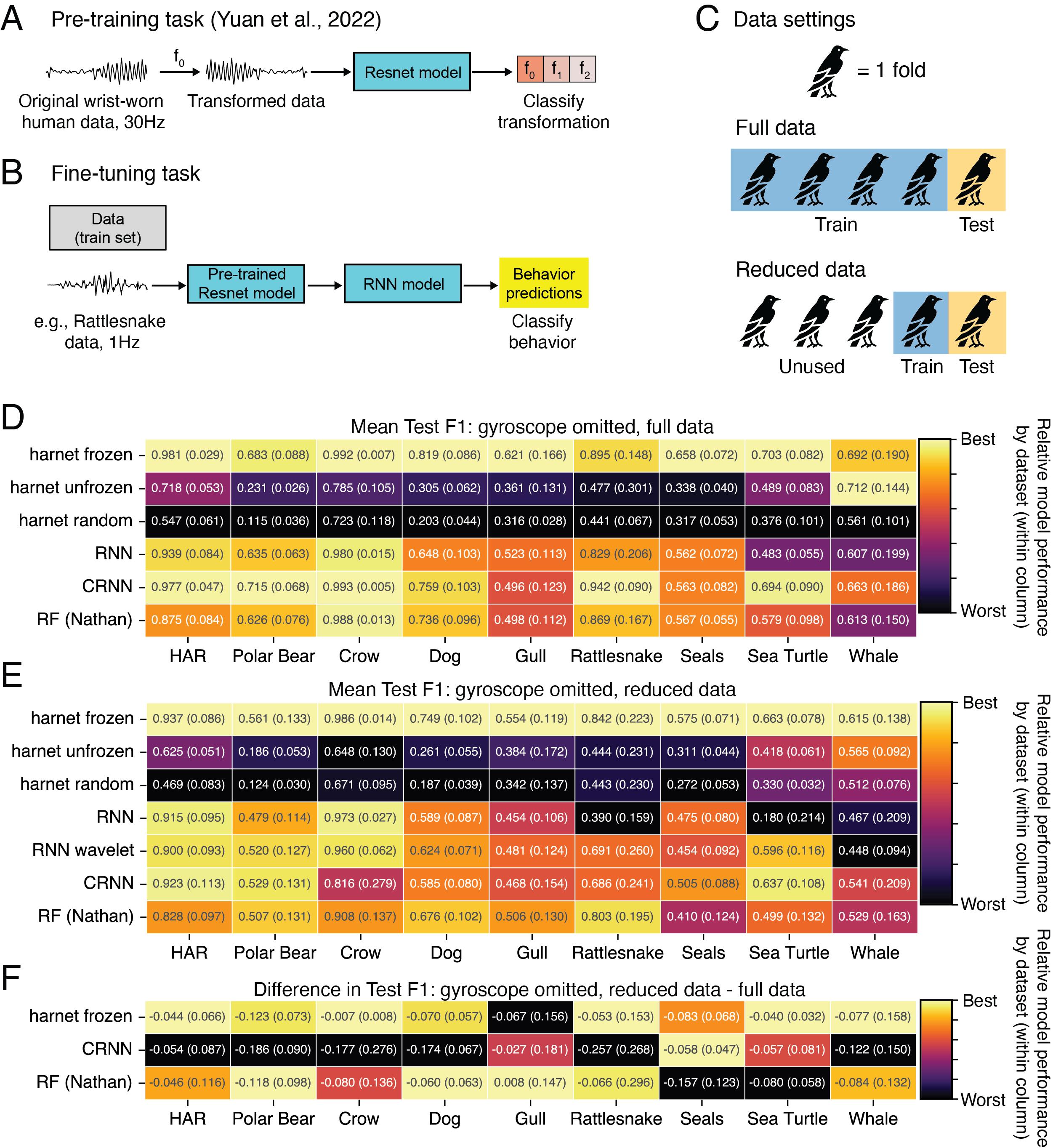}
\caption{Self-supervised pre-training and reduced data setting. A) Pre-training task (performed in~\cite{yuan2022selfsupervised}): The main component of our \texttt{harnet} model has a Resnet architecture~\cite{heDeepResidualLearning2015}. The Resnet was pre-trained with un-annotated human wrist-worn accelerometer data, which was modified with one of a set of signal transformations (e.g. $f_0$ = reversal in time). The network was trained to classify which transformation was applied to the original data. B) In our \texttt{harnet} model, the input to the pre-trained Resnet was animal bio-logger data, without any modification to sampling rate. The outputs of the Resnet were passed to a recurrent neural network (RNN), which produced the behavior predictions. This full \texttt{harnet} model was then trained as shown in Figure~\ref{evaluation_summary}. C) In the full data setting, four out of five folds are used to train the model in one-instance of cross validation. In the reduced data setting, only one fold is used for training while the test set is the same. In other words, approximately four times more individuals are included in the train set in the full data setting, than in the reduced data setting. D) F1 scores for full data task. \texttt{harnet frozen} does best on five datasets and \texttt{CRNN} does best on three datasets. We omitted the \texttt{RNN wavelet} model from the full data experiments, due to high computational resources required for training, and its poor performance in the reduced data setting. E) F1 scores for the reduced data task. \texttt{harnet frozen} does the best on all nine datasets. F) Difference in F1 between reduced and full data tasks. For five datasets, \texttt{harnet frozen} shows the smallest decrease in F1 when using reduced data. For precision and recall results, see Figure~\ref{supplement_precision_recall_representation_results}.}
\label{representation_learning_results}
\end{figure}

We predicted that \texttt{harnet} would out-perform the alternative methods we tested (Table \ref{hypothesestable}, H2). We measured the performance of \texttt{harnet} on each dataset in BEBE, after fine-tuning (Figure~\ref{representation_learning_results}B). To better understand the contribution of the pre-training step on performance, we compared these results with various ablations of \texttt{harnet}: \texttt{RNN}, \texttt{RNN wavelet}, and \texttt{harnet random} (for justification, see Methods). Additionally, we compared \texttt{harnet} with \texttt{\texttt{CRNN}} and \texttt{RF (Nathan)}, which were the best alternatives to \texttt{harnet} in Section~\ref{deep_networks_win}. Finally, we also compared \texttt{harnet} with an alternative setup, \texttt{harnet unfrozen}, which had more tunable parameters (see Methods).
The results of these comparisons are in Figure~\ref{representation_learning_results}D, with precision and recall scores in Supplemental Figure~\ref{supplement_precision_recall_representation_results}. Gyroscope data were not included in the pre-training procedure of~\cite{yuan2022selfsupervised}, and so we omitted these channels when obtaining these results in order to concentrate on the effect of the pre-training methodology. Other channels (e.g. depth) were still included, following the procedure detailed in Methods.

In terms of F1 score, \texttt{harnet} achieved the top score on five of the nine datasets, partly confirming hypothesis (H2). In three of the remaining four cases, \texttt{CRNN} achieved the top score, and in the remaining case, the \texttt{harnet unfrozen} variant achieved the top score. None of the ablations we tested approached the performance of \texttt{harnet}, indicating that the pre-training step, and not another design choice, was responsible for its high performance. The alternative setup for the pre-trained model, \texttt{harnet unfrozen}, achieved lower scores than \texttt{harnet} on eight of nine datasets (average F1 drop: .263). 

To investigate the potential of self-supervised learning in low-data applications, we performed an additional set of computational experiments. For these, during cross validation, we trained each model using only one of five folds (rather than four of five) of the dataset (Figure ~\ref{representation_learning_results}C). The folds used for testing remained the same. Because the folds partition the tagged individuals, this setting reflects a situation where the researcher can only annotate training data from a quarter of the tagged individuals. We predicted that in this setting, \texttt{harnet} would outperform the alternative models we tested (Table \ref{hypothesestable}, H3). The F1 scores of models in the reduced data setting are in Figure~\ref{representation_learning_results}E.

After reducing the amount of training data, the pre-trained \texttt{harnet} model dropped in F1 performance by .056 on average, as compared with a drop of .112 by \texttt{CRNN} and .069 by \texttt{RF (Nathan)} (Figure~\ref{representation_learning_results}F). Additionally, \texttt{harnet} achieved the top F1 score on all nine datasets in the reduced data setting (Figure~\ref{representation_learning_results}E) and across most  behavior classes (Supplemental Figure~\ref{harnet_vs_others_reduced_by_class}), and  \texttt{harnet} achieved the best recall across all datasets. The ablations of \texttt{harnet} consistently performed poorly, relative to other models, in this reduced data setting. Taken together, these results confirm hypothesis (H3).

\subsection{For some behavior classes, model performance improves minimally with increased training data}

\begin{figure}[htbp]
\centering
\includegraphics[width=0.86\textwidth]{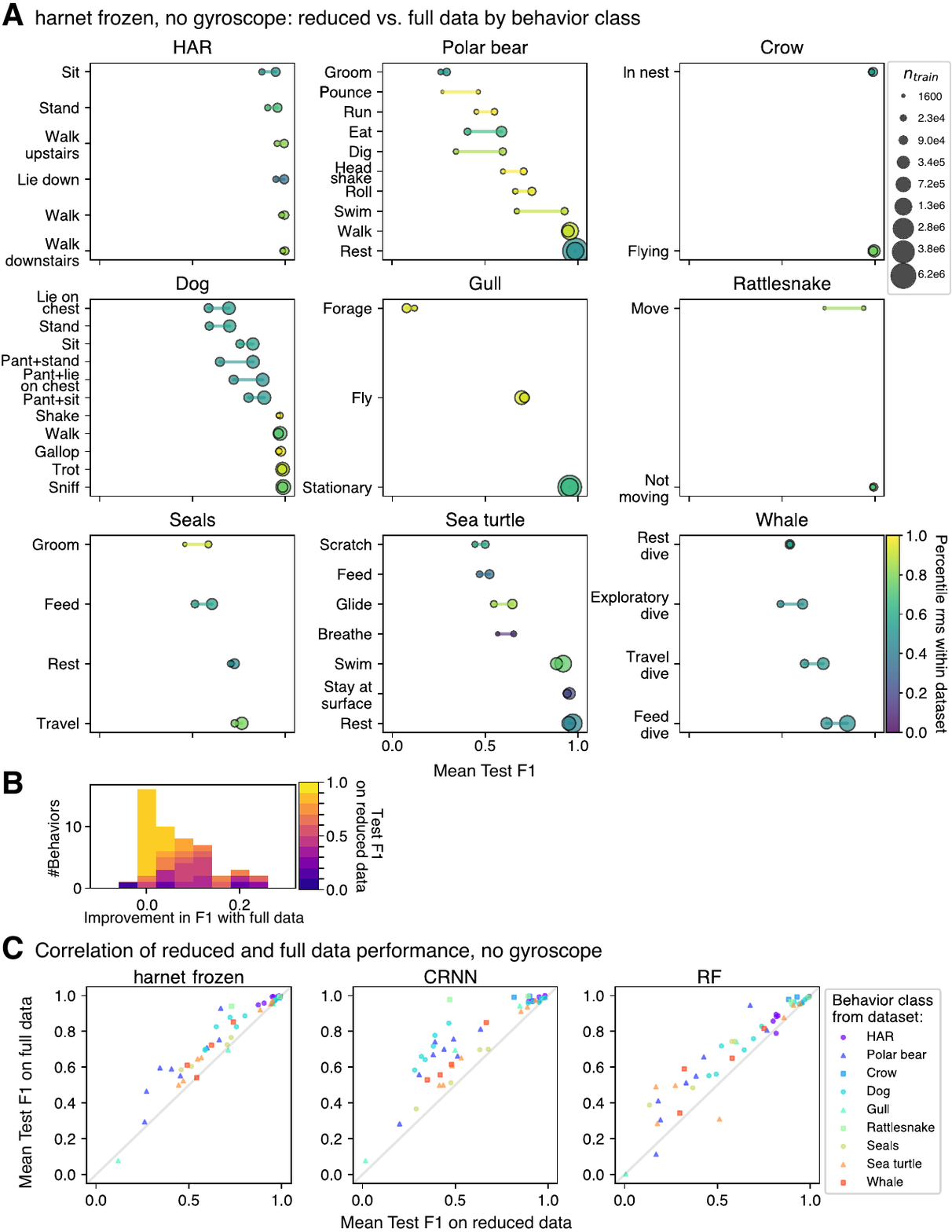}
\caption{Variation in model performance is conserved across data scales. A) Performance of \texttt{harnet} in reduced and full data settings, by behavior class. Size of the marker indicates training dataset size (mean across folds). To help visualize the type of behavior, color indicates the percentile of the average root-mean-square (rms) amplitude of datapoints in that class, as compared to the root-mean-square amplitude of all labeled datapoints. Performance typically improves from the reduced to full data setting, but the rank order of behavior classes remains similar. B) Histogram of per-behavior improvement in \texttt{harnet}'s F1 score across all datasets, moving from reduced data to full data. Colors indicate F1 score in the reduced data setting. C) Performance in reduced data setting versus performance in full data setting, for three models. Each point represents one behavioral class. If performance is the same in the reduced and full data settings, the point  lies on the line. For all three models, there is a high correlation between the reduced and full data settings. \texttt{CRNN} shows the greatest improvements in F1 score with increased data (although \texttt{CRNN} and \texttt{harnet} have comparable performance in the full data setting; Figure~\ref{representation_learning_results}D).}
\label{results_by_class}
\end{figure}

Using BEBE, we investigated the variation in F1 score across different behavioral classes within a single dataset. Using \texttt{harnet} in the full data setting, the inter-class range in F1 score ranged from small (Crow dataset, range: [.990, .995], max-min difference: .0055) to large (Gull dataset, range: [.0768, .955], max-min difference: .878) (Figure~\ref{results_by_class}A, larger dots). This variation also existed in the reduced data setting, and for the \texttt{CRNN} and \texttt{RF (Nathan)} models (Figure~\ref{results_by_class}A, smaller dots; Supplemental Figures~\ref{supplement_rf_results_by_class}-\ref{supplement_crnn_results_by_class}). We found that classes with relatively few training examples can perform as well as classes with many training examples (e.g., Sea turtle: \textit{Stay at surface} vs. \textit{Swim}).

Next, we examined the difference in performance between the reduced and full data settings. We predicted that model performance would improve minimally for some behavioral classes, when trained with four times as much training data (Table~\ref{hypothesestable}, H4; Figure~\ref{representation_learning_results}C). For \texttt{harnet}, the degree of improvement in per-class scores ranged from null (e.g., \textit{Rest dive} in Whale dataset) to moderate (\textit{Swim} in Polar bear dataset: .26 improvement in F1 score), with the median at .053 (Figure~\ref{results_by_class}B). Several classes with F1 scores lower than .9 in the reduced data setting showed small improvements (e.g., \textit{Rest} in Seals dataset). This is consistent with what we would expect if predictive performance for these behaviors had reached an invisible ceiling, but does not conclusively demonstrate it: increasing the data even further or switching models may improve performance on these classes. Nevertheless, we expect that such classes are unlikely to become highly recognizable even going beyond a fourfold increase in training data.

Finally, we quantified the extent to which one could use model performance in the reduced data setting to predict performance in the full data setting. 
For the three models we tested, there was a high degree of correlation between the per-class scores at these two data scales (\texttt{harnet}: $r(47) = .96, p < .001$, \texttt{CRNN}: $r(47) = .89, p < .001$, \texttt{RF}: $r(47) = .94, p < .001$; Figure~\ref{results_by_class}C). Last, we found that the rank ordering of classes within a dataset was conserved when the amount of training data was reduced. Within a dataset, the rank correlation between per-class scores in the reduced and full data settings was typically high (see Supplemental Table~\ref{spearmancorrelations}), indicating that similar classes performed better in the reduced and full data settings.


\section{Conclusions}

To support the development and application of methods for behavior classification in bio-logger data, we designed the Bio-logger Ethogram Benchmark (BEBE), a collection of nine annotated bio-logger datasets. BEBE is the largest, most diverse, publicly available bio-logger benchmark to date. As an example of how BEBE can be used by the community, we tested several hypotheses about ML methods applied to bio-logger data. Based on our results, we are able to make concrete suggestions for those designing studies that rely on ML to infer behavior from bio-logger data.

First, we found that methods based on deep neural networks out-performed the classical ML methods we tested (Table~\ref{hypothesestable}, H1; Figure~\ref{basic_results}). While a similar trend has been observed in other applications of ML, the majority of studies involving bio-logger data still rely on classical methods such as random forests~\cite[Table 2]{thiebaultAnimalborneAcousticData2021}. Contrary to this trend, we suggest that researchers use methods based on deep neural networks in studies where the main objective for ML is to maximize the accuracy of the predicted behavior labels, and with datasets of comparable size to those in BEBE. In particular, for the behavior classification task we describe in this work, we suggest the use of a convolutional-recurrent architecture (which was used by both \texttt{CRNN} and \texttt{harnet}). In contrast to random forests, these deep neural networks learn their features directly from data and therefore do not require an intensive feature engineering step. Additionally, the dominance of CRNN over CNN across datasets demonstrates the importance of incorporating time scale as a learnable parameter (in contrast to RF and CNN where it is fixed).

While we found evidence that convolutional-recurrent networks tend to outperform feature-based methods like RF, they also tend to be more labor-intensive to implement and train. Thus, studies employing ML-based behavior classification may want to weigh the benefits of adopting these methods against their costs. Classification errors can propagate into downstream analyses, increasing the need to correct results for systematic bias~\cite{resheff2022correcting}. Additionally, classification errors can increase uncertainty in downstream analyses. The uncertainty contributed by model classification errors can be quantified, for example, using bootstrap sampling from a confusion matrix~\cite{weinstein2023capturing}. Researchers may consider beginning with a feature-based model, such as RF, and adopting higher-performance methods if the feature-based method does not predict behaviors accurately enough to answer their ultimate scientific question.

We also found that a neural network pre-trained with self-supervision using a large amount of human wrist-worn accelerometer data achieved the best performance on just over half of the datasets (Table~\ref{hypothesestable}, H2; Figure~\ref{basic_results}). This pattern became more pronounced when the amount of training data was reduced (Table~\ref{hypothesestable}, H3; Figure~\ref{representation_learning_results}F). Deep neural networks without self-supervision were not appreciably better than random forests in this reduced data setting (\texttt{CRNN} and \texttt{harnet random} vs. \texttt{RF}; Figure~\ref{representation_learning_results}E). Therefore, we suggest that studies examine adapting pre-trained neural networks (such as~\cite{yuan2022selfsupervised}), rather than training a neural network from scratch. This approach is especially promising for improving behavior classification in situations where a relatively small amount of annotated data is available, for example, due to the difficulty of obtaining ground truth behavior. Self-supervision could complement the use of surrogate species, which has had variable levels of success \cite{campbell2013creating,dickinson2021limitations,paganoUsingTriaxialAccelerometers2017}. We believe this performance is particularly notable since the pre-training data came from only one species (humans, represented in only one dataset in BEBE), and came from only one attachment position (wrist, not represented in BEBE). Additionally, the pre-training data were recorded at 30 Hz (different from all datasets in BEBE), and we did not adjust for differences in sampling rate during fine-tuning. Last, one limitation of this approach was that the pre-training involved only data from accelerometers. A potential future direction would be to perform pre-training using data from more diverse taxa, using a wider variety of tag placements, sampling rates, and sensor types.

Finally, we found that model performance improves minimally for some behaviors, when increasing training data by adding individuals (Table~\ref{hypothesestable}, H4; Figure~\ref{results_by_class}B). Annotating a large train set may not provide sufficient benefit if some of the behaviors of interest are inherently difficult to identify from the available sensors. We also found that per-class model performance is correlated across data scales (Figure~\ref{results_by_class}C). This suggests that, as a part of designing a data analysis procedure, it may be worthwhile to attempt the analysis after annotating a small portion of the available data. This could provide a sense of the expected relative per-behavior model performance, and allow for adjustments to the ethogram if the performance is unlikely to reach the range necessary to address the study question; for example, one might decide to reduce the number of behaviors in the ethogram in order to improve classification performance~\cite{laddsSuperMachineLearning2017, studdBehavioralClassificationLowfrequency2019}.
With a diversity of bio-logging studies represented, BEBE may provide a shared resource for practitioners to identify which behaviors are easily discriminated from sensor data. While we cannot conclusively determine whether we reached ceiling performance for behaviors in BEBE with the experiments presented, we found that some behaviors showed minimal improvement with more training data (Figure~\ref{results_by_class}).
Therefore, when using BEBE as a benchmark, we suggest that researchers track per-behavior performance improvements.

An important aspect of BEBE is that the data and evaluation code is openly available. This allows others to test hypotheses beyond those in Table~\ref{hypothesestable}, and to test future, yet to be developed ML methods. For example, future work could employ BEBE to systematically test different data augmentation techniques such as those suggested by~\cite{otsuka2024exploring}, intended to improve performance with a small amount of annotated data. Additionally, it is possible that one could improve upon the conclusions in this article. For example, when testing (H1), for the classical models we tested, we used our own implementation of one of two types of generic hand-engineered features as model inputs~\cite{nathanUsingTriaxialAcceleration2012,sakamotoCanEthogramsBe2009}. It is possible that other feature sets or model types would improve on the performance of the models we evaluated. For example, Wilson et al. \cite{wilson2018give}, suggest using features tailored to the bio-mechanics of specific behaviors of interest. (For instance, to detect feeding in sea turtles, Jeantet et al. \cite{jeantetBehaviouralInferenceSignal2020} pre-segmented data based on the variance in the angular speed in the sagittal plane.) Using BEBE as a common framework for comparison, others can interrogate the results presented here and improve suggestions for the bio-logging community.

A number of prior works have tested multiple ML algorithms for behavior classification, typically on a single dataset (e.g.,~\cite{brewsterDevelopmentApplicationMachine2018,nathanUsingTriaxialAcceleration2012,https://doi.org/10.1002/ece3.10035}). We extend this trend by testing multiple ML algorithms on multiple datasets. Bio-logger datasets tend to be very heterogenous, and can differ in study system, sensor type, sampling rate, ethogram definition and train-test split design. When comparing the results of disparate studies, it can be difficult to disentangle the effect of model design from the effects of the dataset. 
While we observed that there is a large amount of variability in model performance between datasets, we found that certain techniques performed consistently well (relative to alternatives). Therefore, by systematically testing techniques in a variety of settings, we are able to observe patterns in how ML methods are applied to bio-logger data, generally.

While this study focuses on supervised behavior classification, unsupervised behavior classification (which does not rely on labeled data) is of interest where there is little knowledge of relevant behaviors or where behavior is difficult to observe \cite{chimientiUseUnsupervisedLearning2016,sakamotoCanEthogramsBe2009}. Based on an analysis of two seabird datasets, Sur et al. \cite{https://doi.org/10.1002/ece3.10035} argue that these methods perform worse than supervised methods in recovering pre-defined behaviors. They are also difficult to systematically evaluate. It is typically unclear what aspect of behavior they target and their outputs therefore require interpretation. To address this particular challenge, ethograms that are defined hierarchically, i.e., with behaviors composed of multiple more specific behaviors (e.g., \cite{jeantetBehaviouralInferenceSignal2020,laddsSeeingItAll2016, studdBehavioralClassificationLowfrequency2019,shepardIdentificationAnimalMovement2008}), may provide a promising basis for evaluation, by providing insight into which aspects of behavior are modeled by a certain method.  

We note limitations and potential improvements to our approach. First, the datasets in BEBE are primarily based on tri-axial accelerometers, which may not be able to represent motion accurately enough to distinguish all behaviors of interest~\cite{TongDeadEnd2020}. Bio-loggers incorporating other sensor types, such as gyroscopes, audio, and video, will likely give researchers more complete characterizations of individuals' behaviors. Other benchmarks exist for animal behavior detection entirely from video \cite{chen2023mammalnet,ng2022animal} but these do not focus on bio-logger data, which may present additional challenges such as intermittent video logging  or a limited field of view \cite{korpelaMachineLearningEnables2020,paganoUsingTriaxialAccelerometers2017}. A future benchmark could include data types not examined in BEBE.


Second, bio-loggers can shed new light on conservation problems and interventions, as well as on patterns of animal behavior and energy expenditure~\cite{husseyAquaticAnimalTelemetry2015, kaysTerrestrialAnimalTracking2015, tuiaPerspectivesMachineLearning2022, VALLETTA2017203, wilsonPryingIntimateDetails2008}. In this study, we provide a standardized task for one extremely common analysis, behavior classification. Depending on the intended application, other analyses may be useful. These could include detecting unusual patterns in data~\cite{Anomaly2021} that may indicate changes in behavior or environmental conditions~\cite{Jetz2022BiologicalEO}, or counting the rate at which a specific type of behavioral event occurs~\cite{batesonMeasuringBehaviourIntroductory2021}. Future studies could use BEBE datasets, but formalize new tasks and evaluation metrics for use-cases that arise in these settings. In addition, studies based on BEBE could explore evaluation metrics to promote advances in on-device ML~\cite{korpelaMachineLearningEnables2020}, such as device energy consumption metrics to assess on-device feasibility. This could additionally give insight into environmental impacts due to model usage~\cite{henderson2020towards,rajiAIEverythingWhole2021,tuiaPerspectivesMachineLearning2022}. Overall, we expect BEBE can support researchers by facilitating access to a breadth of study systems, which may involve using our standard task or creating different uses of BEBE datasets.

Third, there is variation in how the datasets were collected and annotated, and BEBE has limited taxonomic spread and no taxonomic replication. While some variation is desirable in order to promote generalizable methods development, it also complicates between-dataset or between-behavior comparisons. These types of comparisons could illuminate how a model's predictive ability is related to biological factors, such as phylogeny or body size,
and to non-biological factors, such as the choice of ethogram or data modalities included. It may be possible to quantify the effects of these factors using a benchmark with more datasets available and better data standardization.

Fourth, in our standardized evaluation framework, we exclude Unknown behaviors in the training objective and evaluation metric. The presented models assume that the provided known behavior labels are the only possible categories, and will apply one of them to all datapoints. This would be disadvantageous in applying supervised learning to bio-logger data where we usually know some behaviors have not been labeled. Approaches to accounting for unknown behaviors include using an “other” category~\cite{jeantetBehaviouralInferenceSignal2020, laddsSeeingItAll2016} and thresholding classification probabilities to make a prediction~\cite{glass2020accounting}. One potential future usage of BEBE would be to test these different methodologies for accounting for unknown behaviors, to elucidate their impact on recovering the behaviors of interest. 

Finally, human-generated behavior annotations are susceptible to error, due to e.g. difficulty in observing a behavior, mistakes during data entry, or differences in human judgment. To mitigate this, behaviors can be annotated by multiple raters, and then checked against one another. However, this can be extremely time consuming, and in BEBE only the authors of the Rattlesnake, Whale, and Seal datasets performed this step. In spite of this potential annotation noise, we were able to observe consistent patterns in model performance across multiple datasets in BEBE. In the future, if more studies report between-rater agreement in annotations, it may become possible to quantify the magnitude of these errors.



\paragraph{Call for Collaboration} The code repository includes instructions on how datasets outside of BEBE may be formatted for use with the methods in BEBE. Interested researchers may make their formatted datasets discoverable from the BEBE repository. These datasets would become easily available for others, but would not become part of the task in this paper, which must remain standardized.

However, it is typical for benchmarks to be updated when key challenges are sufficiently met~\cite{dehghani2021benchmark}. In light of the preceding discussion, we seek community contributions that could lead to a more comprehensive benchmark, with three main objectives: \begin{enumerate}
\item
To provide researchers with means to understand how modeling decisions influence model performance,
\item To enable analyses which compare recorded behavior across taxa, and 
\item To formalize tasks which reflect a variety of real-world applications, including conservation applications. 
\end{enumerate} 
We expect these objectives will be best served by a benchmark with more diversity in its representation of taxa, data types, tag placement positions, sensor configurations, ethograms, and modeling tasks. Possible contributions include (1) annotated datasets to be made openly available to the research community (whether already available or not), (2) design of data and annotation standardization, and (3) design of benchmark tasks that reflect applications of ML and bio-logger technology. For any ensuing publications, contributors would have the option to co-author the manuscript. Interested researchers can follow the instructions at \url{https://github.com/earthspecies/BEBE}. 

We have proposed that benchmarks can encourage the development and rigorous evaluation of ML methods for behavioral ecology. We envision many possible future outcomes for this line of research: for example, best practices for bio-logger data analysis, an ML-based toolkit that can be adapted to different study systems, or powerful species-agnostic tools that can be applied across taxa and sensor types.  This could, in turn, inform more effective conservation interventions, as well as guide the development and testing of hypotheses about animal behavior.

\subsection*{Ethics Statement}

All animal behavior datasets except the Crow dataset were reported in previous publications. Crow behavior data were collected in accordance with ASAB/ABS guidelines and Spanish regulations for animal research, and were authorized by Junta de Castilla y León (licence: EP/LE/681-2019).

\section*{Author Contributions}

 M. Cusimano, A. Friedlaender, and B. Hoffman conceived the ideas;
 M. Cusimano and B. Hoffman designed methodology; V. Baglione, D. Canestrari, D. Chevallier, D. DeSantis, A. Friedlaender, L. Jeantet, M. Ladds, T. Maekawa, V. Mata-Silva, V. Moreno-González, A. Pagano, E. Trapote, O. Vainio, A. Vehkaoja, K. Yoda, contributed the data; M. Cusimano, B. Hoffman, and K. Zacarian coordinated data contributions; M. Cusimano and B. Hoffman analysed the data;  M. Cusimano and B. Hoffman led the writing of the manuscript. All authors contributed critically to the drafts and gave final approval for publication.

\section*{Data Availability}

The datasets generated and/or analysed during the current study are available in the Zenodo repository, (doi: 10.5281/zenodo.7947104). All model results used to create the figures are also available at the same repository. Code used to format the datasets is available at \url{https://github.com/earthspecies/BEBE-datasets/}. Code used to implement, train, and evaluate models is available at \url{https://github.com/earthspecies/BEBE/}.

\section*{Acknowledgments}

This project was supported (in part) by a grant from the National Geographic Society. Compute resources were provided in part by Google Cloud Platform. We thank Phoebe Koenig for input on benchmark design. We thank Christian Rutz and Mark Johnson for critical contributions to an earlier version of the manuscript. We thank Felix Effenberger, Masato Hagiwara, Sara Keen, Jen-Yu Liu, and Marius Miron for constructive discussions. Any use of trade, firm, or product names is for descriptive purposes only and does not imply endorsement by the United States Government. 

Whale behavior data collection was funded by National Science Foundation Office of Polar Programs grants awarded to A. Friedlaender, and were collected under National Marine Fisheries Service Permits 14809 and 23095, Antarctic Conservation Act permits and UCSC IACUC Permit Friea2004. 

Crow behavior data collection was funded by the Ministerio de Economía y Competitividad – España (Grant CGL2016 – 77636-P to V. Baglione).

Rattlesnake behavior data collection was funded by a National Science Foundation Graduate Research Fellowship (NSF-GRFP) awarded to D. L. DeSantis and grants from the UTEP Graduate School (Dodson Research Grant) awarded to D. L. DeSantis. J. D. Johnson, A. E. Wagler, J. D. Emerson, M. J. Gaupp, S. Ebert, H. Smith, R. Gamez, Z. Ramirez, and D. Sanchez contributed to dataset development, and C. Catoni and Technosmart Europe srl. developed the accelerometers. 

Dog behavior data collection was funded by Business Finland, a Finnish funding agency for innovation, grant numbers 1665/31/2016, 1894/31/2016, and 7244/31/2016 in the context of ``Buddy and the Smiths 2.0'' project. 

Gull behavior data collection was funded by JSPS KAKENHI Grant Number JP21H05299 and JP21H05294, Japan. 

Sea turtle behavior data was collected within the framework of the Plan National d’Action Tortues marines des Antilles et the Plan National d’Action Tortues marines de Guyane Française, with the support of the ANTIDOT project (Pépinière Interdisciplinaire Guyane, Mission pour l’Interdisciplinarité, CNRS) and BEPHYTES project (FEDER Martinique) led by D. Chevallier. DEAL Martinique and Guyane, the CNES, the ODE Martinique, POEMM and ACWAA associations, Plongée-Passion, Explorations de Monaco team, the OFB Martinique and the SMPE Martinique provided technical support and field assistance. Numerous volunteers and free divers participated in the field operations to collect data. EGI, France Grilles and the IPHC Computing team provided technical support, computing and storage facilities for the original development of the Sea Turtle dataset.

Polar bear behavior data collection was funded by the U.S. Geological Survey Changing Arctic Ecosystems Initiative and the Species and Land Management programs of the U.S. Geological Survey Ecosystems Mission Area. 

Seals behavior data collection was assisted by the marine mammal staff at Dolphin Marine Magic, Sealife Mooloolaba and Taronga Zoo Sydney.

Animal icons in Figure 1B and repeated throughout the paper have a Flaticon license (free for personal/commercial use with attribution), attributions are as follows: (1) Gull: user \href{https://www.flaticon.com/free-icon/seagull_1137601}{Smashicons}, (2) Rattlesnake: user \href{https://www.flaticon.com/free-icon/rattlesnake_4343800}{Freepik}, (3) Polar bear: user \href{https://www.flaticon.com/free-icon/polar-bear_1063435}{Chanut-is-Industries}; (4) Dog: user \href{https://www.flaticon.com/free-icon/sitting-dog_87987}{Freepik}, (5) Whale: user \href{https://www.flaticon.com/free-icon/whale_9375062}{The Chohans Brand}, (6) Turtle: user \href{https://www.flaticon.com/free-icon/turtle_2622036}{Freepik}, (7) Crow: user \href{https://www.flaticon.com/free-icon/raven_3504719}{iconixar}, (8) Seal: user \href{https://www.flaticon.com/free-icon/seal_1996768}{monkik}, (9) Human: user \href{https://www.flaticon.com/free-icon/walking_6591691}{Bharat Icons}. Image attributions for Figure 1C are as follows: (1) Gull: scaled and cropped from user Wildreturn ( \href{https://flic.kr/p/C7b267}{Flickr}; CC BY 2.0), (2) Rattlesnake: scaled and cropped from  user snakecollector (\href{https://flic.kr/p/4QXXP7}{Flickr}; CC BY 2.0), (3) Polar bear: scaled and cropped from user usfwshq (\href{https://www.flickr.com/photos/usfwshq/6862203253/in/photostream/}{Flickr}; CC BY 2.0), (4) Dog: Andrea Austin, (5) Whale: scaled and cropped image from  user onms ( \href{https://flic.kr/p/2isRFaH}{Flickr}; CC BY 2.0), (6) Sea turtle: scaled and cropped from user dominic-scaglioni on (\href{https://flic.kr/p/cGfgdm}{Flickr}; CC BY 2.0), (7) Crow: scaled and cropped from user alexislours (\href{https://flic.kr/p/2n2CdQg}{Flickr}; CC BY 2.0), (8) Seal: scaled and cropped from volvob12b (Bernard Spragg) (\href{https://flic.kr/p/qM5AM4}{Flickr}; Public domain), (9) Human: Katie Zacarian. 

\bibliographystyle{vancouver} %
\bibliography{bebe} %

\newpage

\section*{Supplemental Information}

\subsection*{Time Scales} Animal behavior can be described hierarchically, in which actions are nested into multiple time scales~\cite{adamJointModellingMultiscale2019, bermanPredictabilityHierarchyDrosophila2016}: for example, the human behavior \emph{Walking} may be hierarchically composed of two repeating, shorter time-scale behaviors, the left and right forward steps. For simplicity, in this study we focus on a single non-hierarchical set of annotations per dataset. However, there are multiple time scales represented across the nine ethograms in BEBE. For example, some behavior classes reflect brief, low-level activities (e.g. shaking), whereas some reflect longer, higher-level activities (e.g. foraging, exploration). In order to give a rough quantification of the time scales present in these ethograms, for each dataset we computed the average amount of time an individual spends in a known behavioral state, before it switches to a different known behavioral state or an \textit{Unknown} state. This quantity is reported in Table~\ref{summarytable} as the mean annotation duration. The mean annotation duration should only be taken as a rough estimate of the typical duration of a behavioral state, because the annotations in the original studies were not necessarily produced with the intention of measuring onsets and offsets of behavioral states.

For the Polar Bear dataset, to compute mean annotation duration, we had to account for the fact that the video footage used to make annotations was duty cycled. Because of this duty cycling, there are periodic intervals of up to 90 seconds in which annotations are \textit{Unknown}. To account for these \textit{Unknown} intervals, we assumed that if the bear is in the same behavioral state before and after an \textit{Unknown} interval of less than 91 seconds, then the bear was in that behavioral state during the \textit{Unknown} interval. This procedure was only used to compute mean annotation duration, and not to add additional annotations for model training or evaluation.

\subsection*{Color Mapping}

For Figure~\ref{basic_results}, as well as Supplemental Figures \ref{supplement_precision_recall_basic_results} and \ref{supplement_precision_recall_representation_results}, we use the perceptually uniform {\tt inferno} color mapping provided by the MatPlotLib~\cite{Hunter:2007} Python package. Before applying the color mapping, we rescale the values in each column (i.e., average scores for a set of models evaluated on the same data) in order to emphasize the relative performance of the models. To do so, for each F1, precision, and recall table, we linearly rescale the values in each column so that the maximum value in each column is 1 and the minimum value in each column is 0.

\newpage

\subsection*{Additional Figures}

\renewcommand{\thefigure}{S\arabic{figure}}
\renewcommand{\theHfigure}{S\arabic{figure}}
\setcounter{figure}{0}

\renewcommand{\thetable}{S\arabic{table}}
\setcounter{table}{0}

\begin{figure}[H]
\centering
\includegraphics[width=\textwidth]{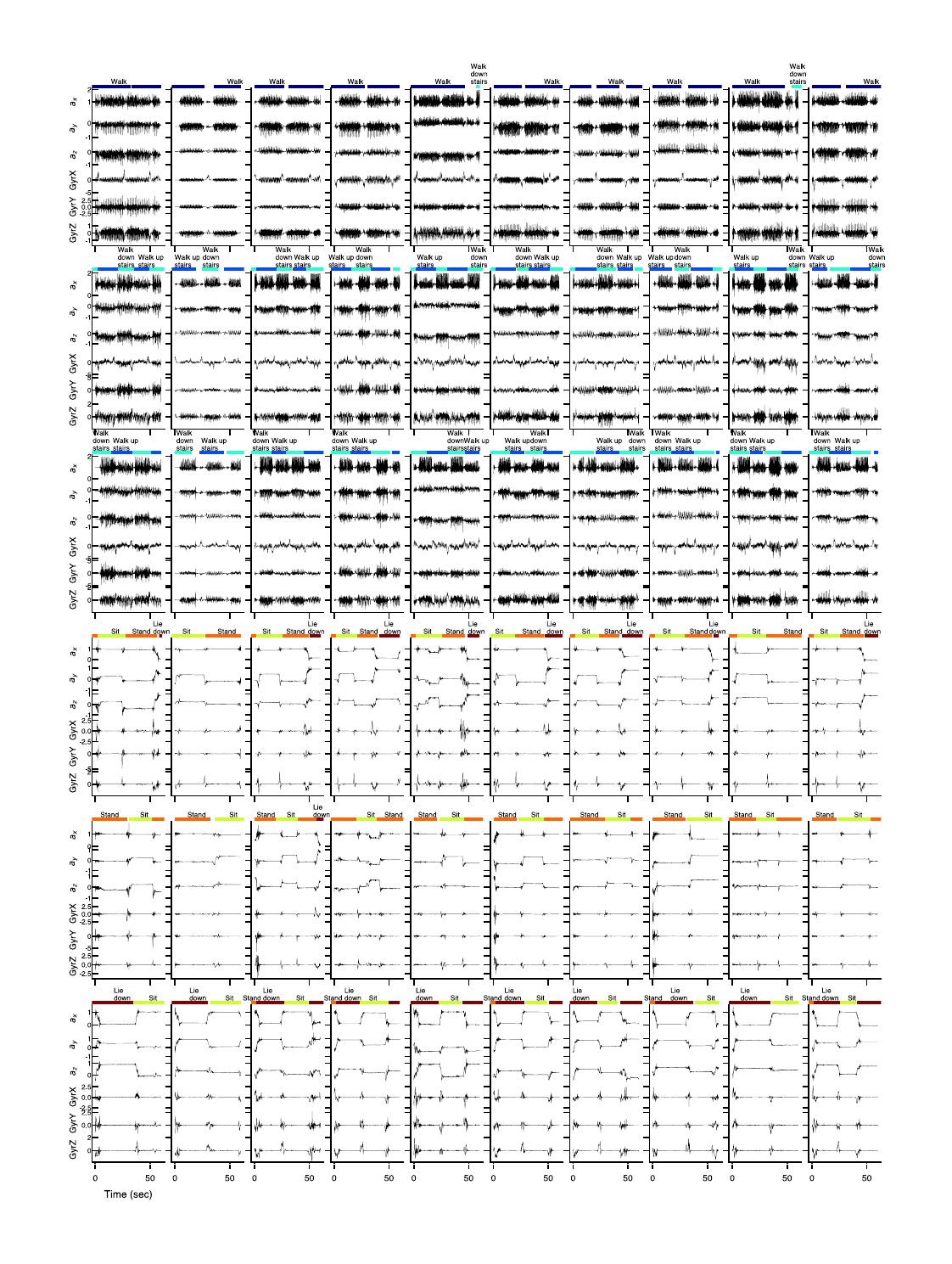}
\caption{10 examples of each class in the Human Activity Recognition (HAR) dataset, beginning five seconds before the class onset. Accelerometer units are $g$, gyroscope units are $rad/sec$. No information was provided on the axes orientation of the sensors.}
\label{HAR_examples_supplement}
\end{figure}

\begin{figure}[H]
\centering
\includegraphics[width=\textwidth]{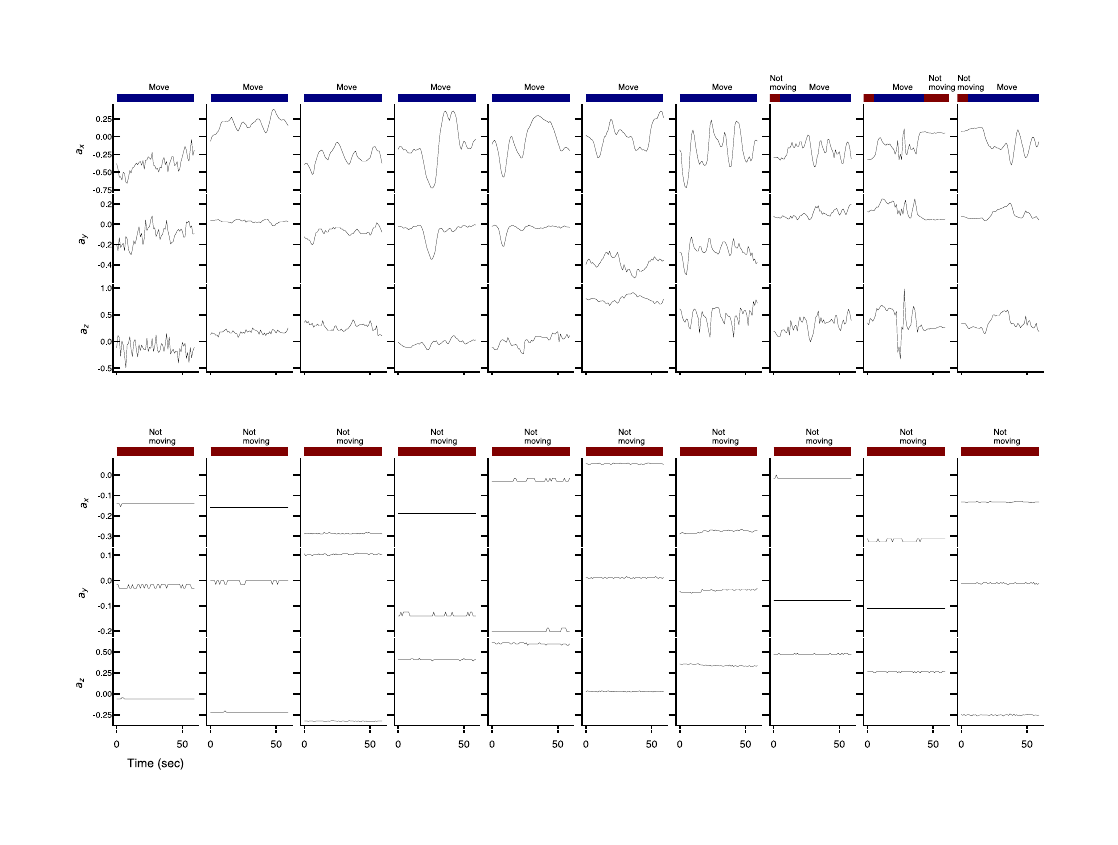}
\caption{10 examples of each class in the Rattlesnake dataset, beginning five seconds before the class onset. Accelerometer units are $g$. Bio-logger was surgically implanted at 2/3 the body length of the rattlesnake. Axes: $x$ (forward- backward), $y$ (left-right), $z$ (up-down).} 
\label{rattlesnake_examples_supplement}
\end{figure}

\begin{figure}[H]
\centering
\includegraphics[width=\textwidth]{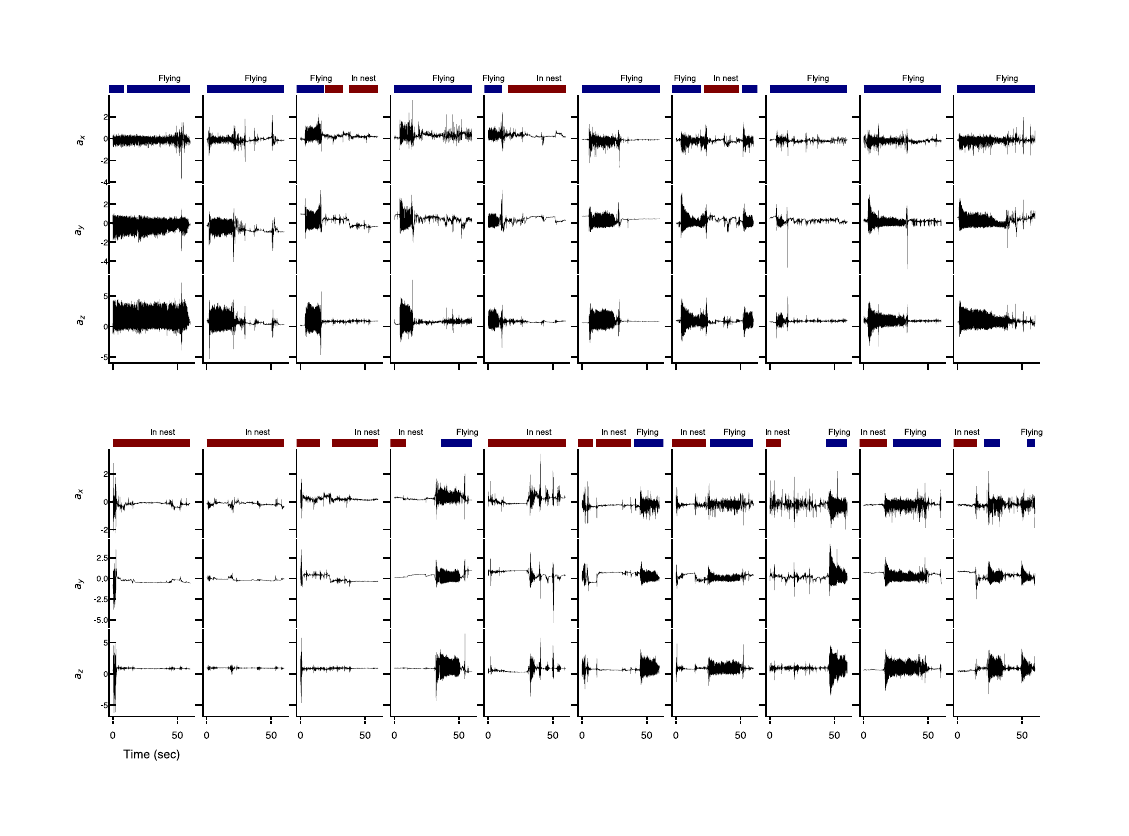}
\caption{10 examples of each class in the Crow dataset. Accelerometer units are $g$. Bio-logger was attached to the base of the crow's tail. Axes: $x$ (backward-forward), $y$ (lateral), $z$ (down-up).} 
\label{crow_examples_supplement}
\end{figure}

\begin{figure}[H]
\centering
\includegraphics[width=\textwidth]{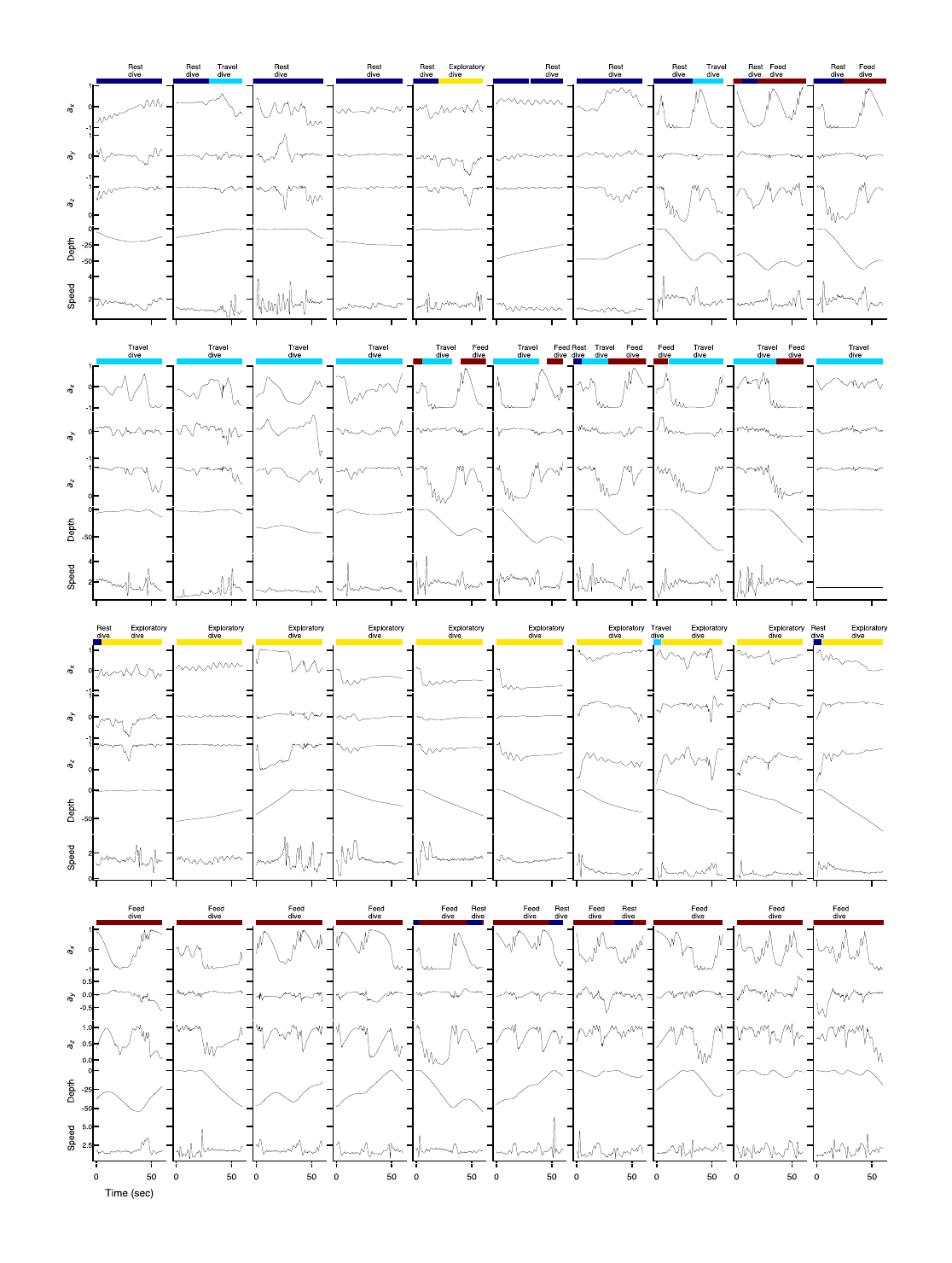}
\caption{10 examples of each class in the Whale dataset. Accelerometer units are $g$. Depth units are $m$ from the water's surface. Speed units are $m/s$. Bio-loggers were placed on the dorsal surface of the whale or high on the flank, forward of the dorsal fin. Axes: $x$ (backward-forward), $y$ (lateral), $z$ (down-up).} 
\label{whale_examples_supplement}
\end{figure}

\begin{figure}[H]
\centering
\includegraphics[width=\textwidth]{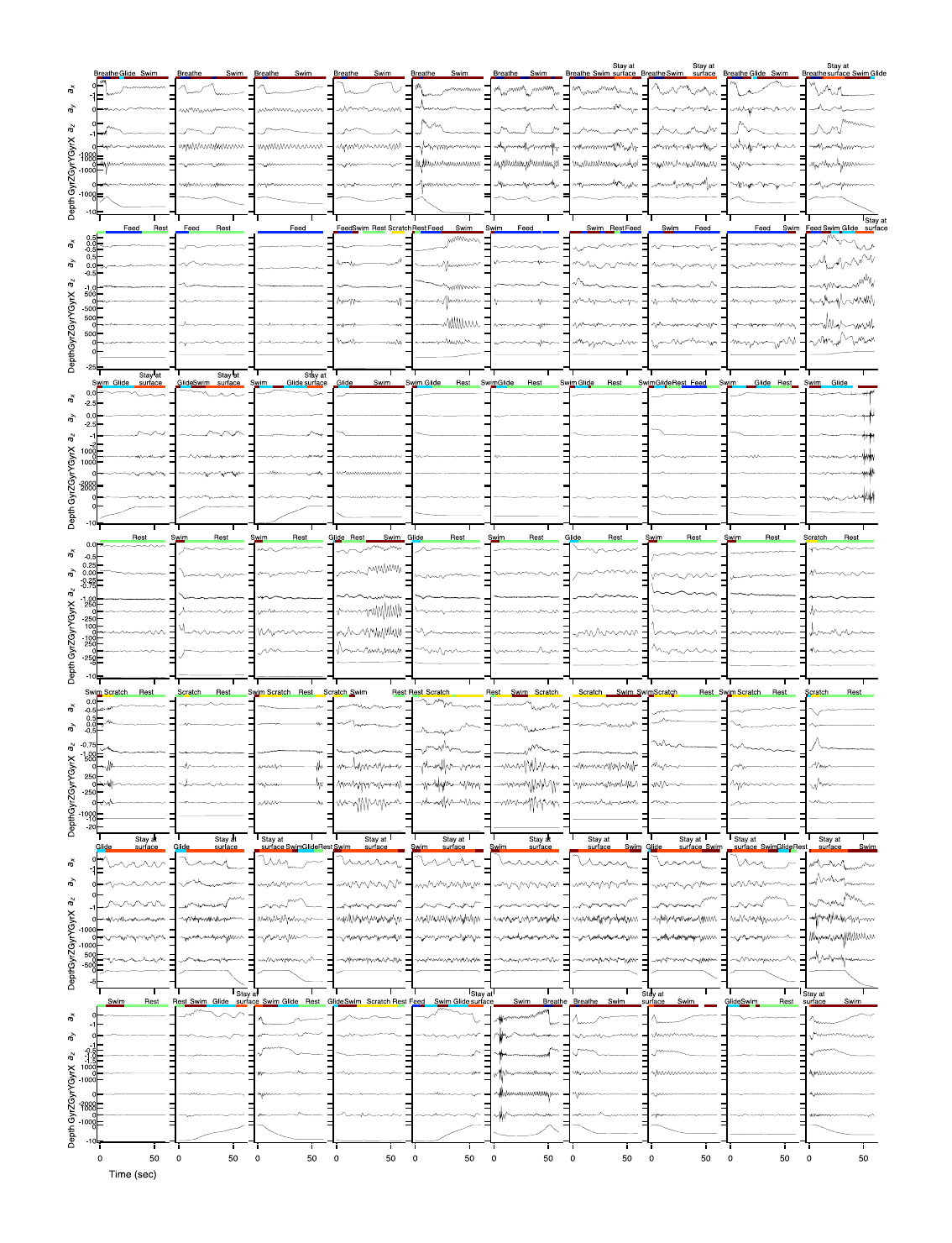}
\caption{10 examples of each class in the Turtle dataset. Accelerometer units are $g$. Gyroscope units are $mrad/sec$. Depth units are $m$ from the water's surface.  Bio-loggers were placed on the carapace in a tilted orientation. Axes: $x$ (back-to-front), $y$ (right-to-left), $z$ (bottom-to-top).} 
\label{turtle_examples_supplement}
\end{figure}

\begin{figure}[H]
\centering
\includegraphics[width=\textwidth]{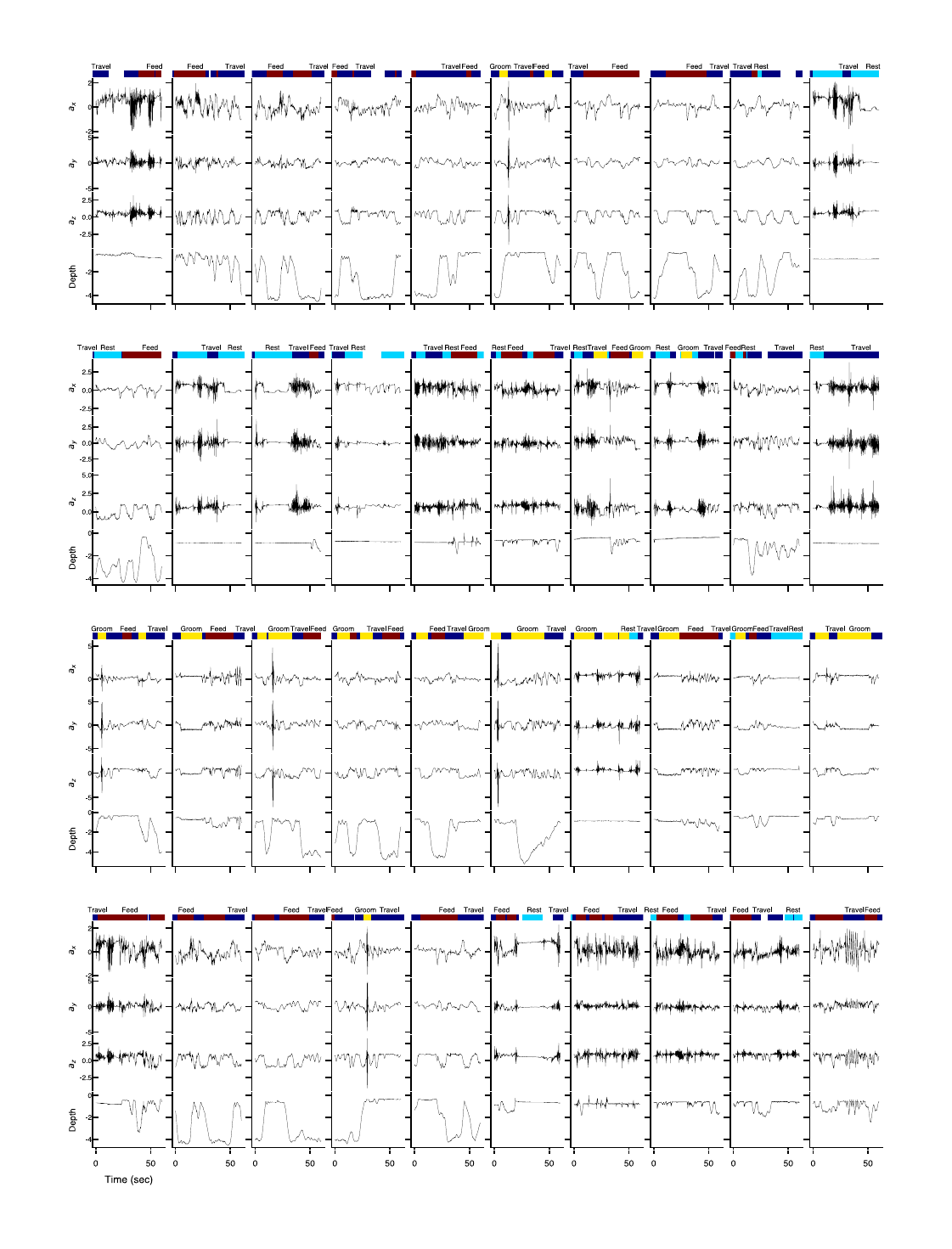}
\caption{10 examples of each class in the Seals dataset. Accelerometer units are $g$. Gyroscope units are $mrad/sec$. Depth units are $m$ from the water's surface.  Bio-loggers were placed between the shoulder blades. Axes: $x$ (anterior-posterior), $y$ (lateral), $z$ (dorsal-ventral).} 
\label{seal_examples_supplement}
\end{figure}

\begin{figure}[H]
\centering
\includegraphics[width=\textwidth]{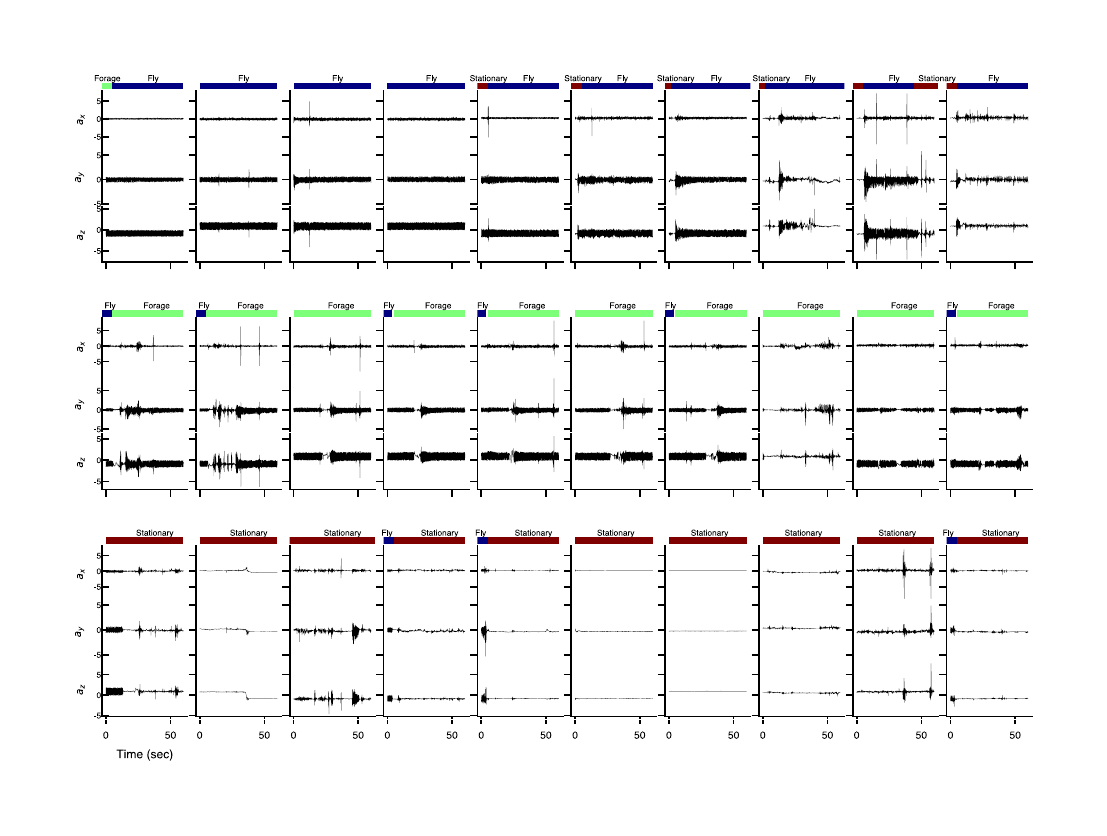}
\caption{10 examples of each class in the Gull dataset. Accelerometer units are $g$. Bio-loggers were placed on the back or abdomen. Axes: $x$ (lateral), $y$ (forward-backward), $z$ (down-up).} 
\label{gull_examples_supplement}
\end{figure}

\begin{figure}[H]
\centering
\includegraphics[width=\textwidth]{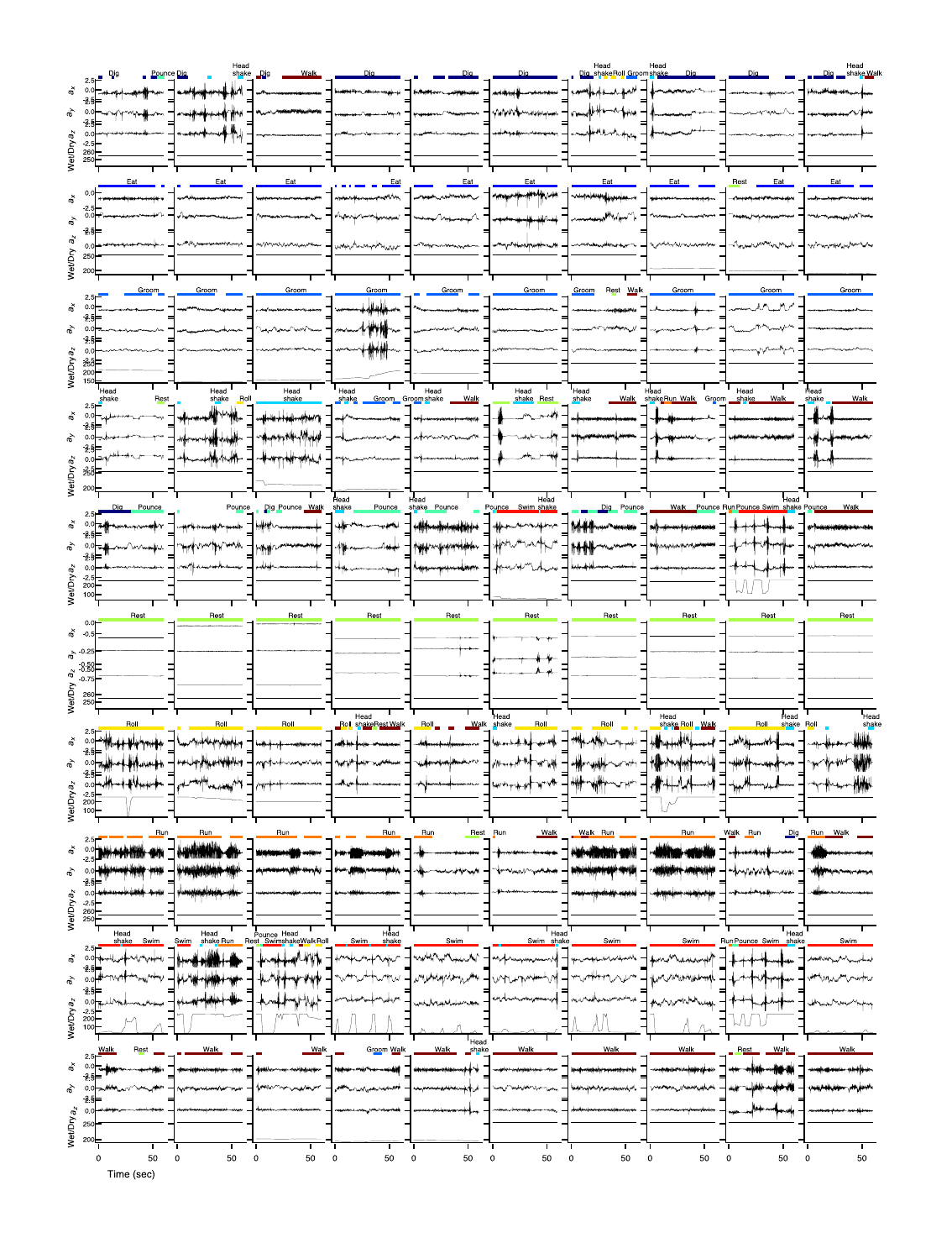}
\caption{10 examples of each class in the Polar bear dataset. Accelerometer units are $g$. The wet sensor ranges from 0(wet)-255(dry). Bio-loggers were placed as a collar around the polar bear's neck. Axes: $z$ (lateral), $x$ (backward-forward), $y$ (down-up).} 
\label{polarbear_examples_supplement}
\end{figure}

\begin{figure}[H]
\centering
\includegraphics[width=\textwidth]{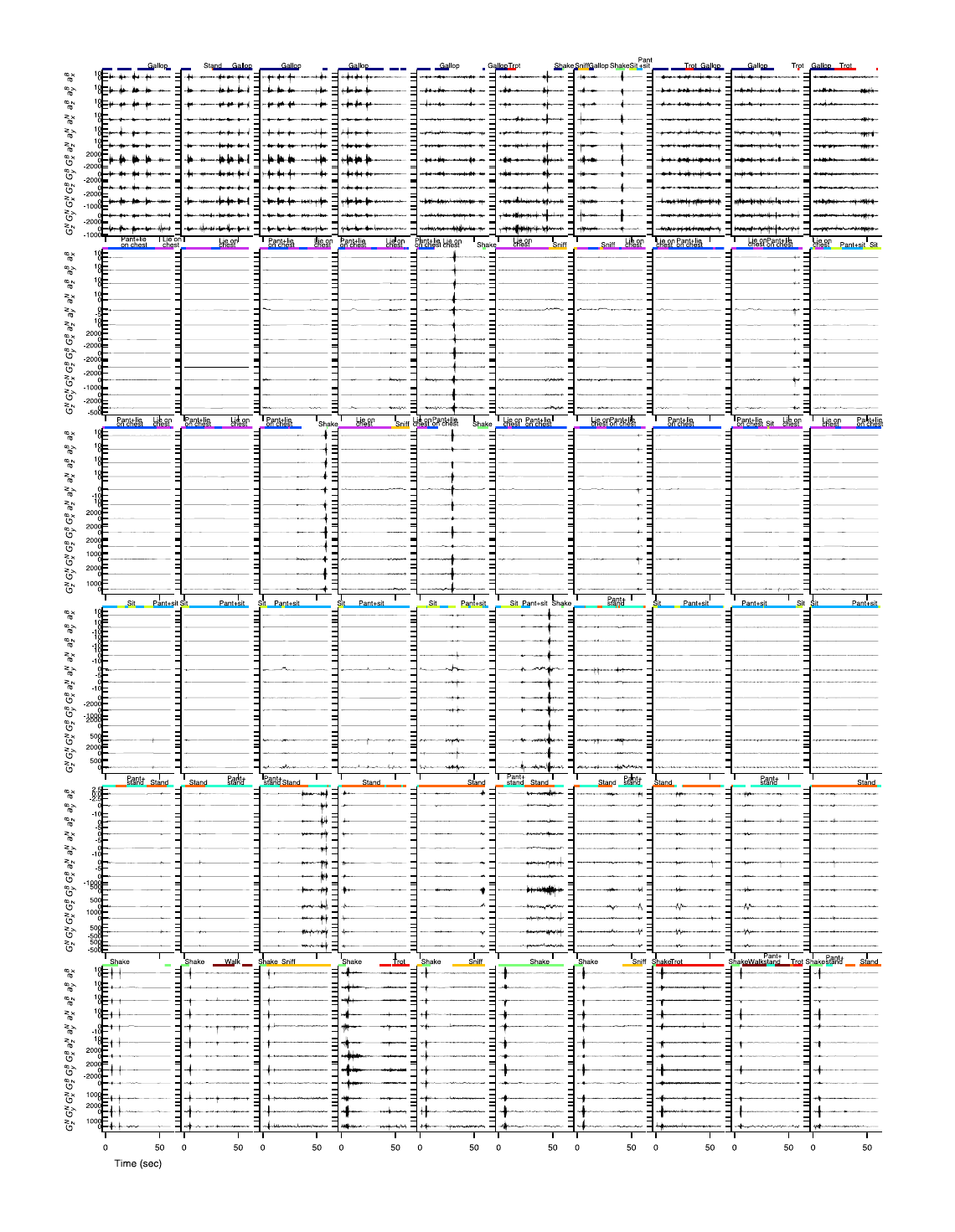}
\caption{10 examples of  classes in the Dog dataset. See next figure for remaining classes. Accelerometer units are $g$. No information on axes orientation was provided.} 
\label{dog_examples0_supplement}
\end{figure}

\begin{figure}[H]
\centering
\includegraphics[width=\textwidth]{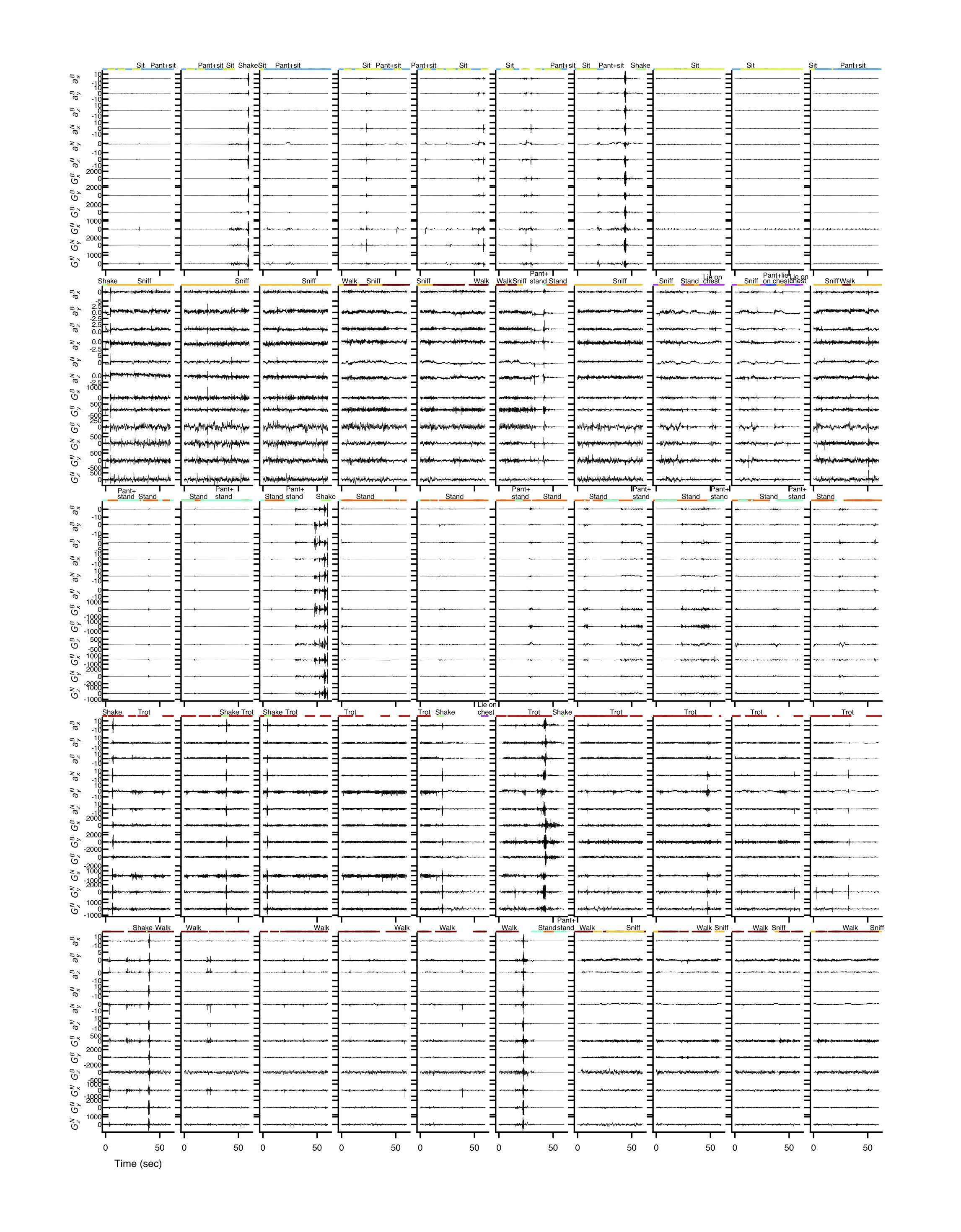}
\caption{10 examples of  classes in the Dog dataset. See preceding figure for remaining classes. Accelerometer units are $g$. No information on axes orientation was provided.} 
\label{dog_examples1_supplement}
\end{figure}

\begin{figure}[H]
\centering
\includegraphics[width=\textwidth]{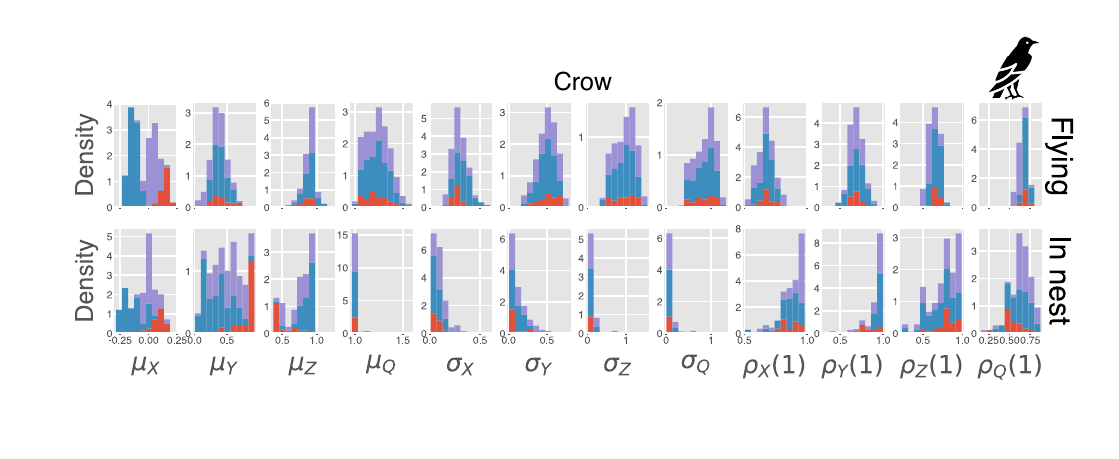}
\caption{Crow dataset summary statistics of the accelerometer $(x,y,z)$-axes and $q$-axis (root-mean-square amplitude), for each class. Each color represents a different individual (not all individuals are shown): the histograms are stacked. $\mu$: mean, $\sigma$: standard deviation, $\rho(1)$: one-sample auto-correlation. Summary statistics computed over 20 seconds. These statistics show within and between class variation, as well as between individual variation. For example, note the difference in $\mu_Q$ for the two classes.} 
\label{crow_histogram_supplement}
\end{figure}

\begin{figure}[H]
\centering
\includegraphics[width=\textwidth]{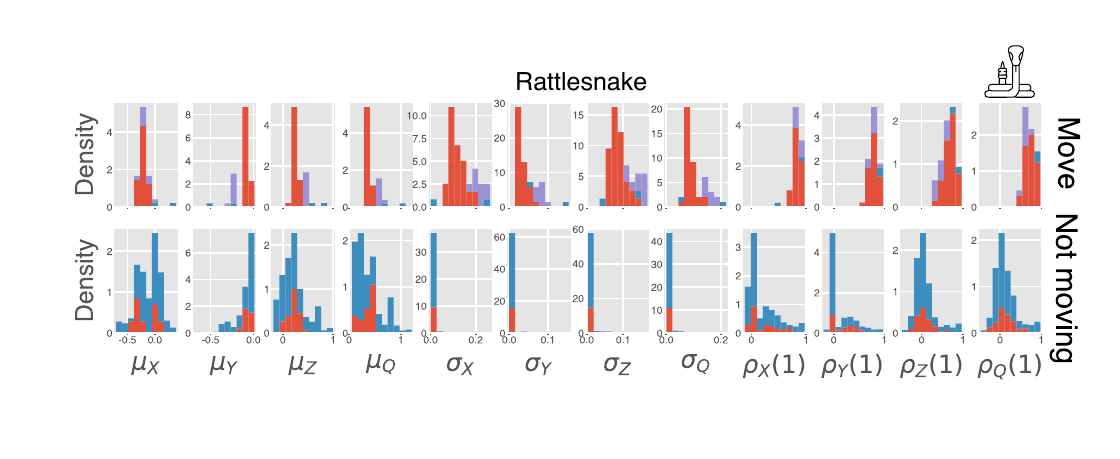}
\caption{Rattlesnake dataset summary statistics of the accelerometer $(x,y,z)$-axes and $q$-axis (root-mean-square amplitude), for each class. Each color represents a different individual (not all individuals are shown): the histograms are stacked. $\mu$: mean, $\sigma$: standard deviation, $\rho(1)$: one-sample auto-correlation. Summary statistics computed over 60 seconds. These statistics show within and between class variation, as well as between individual variation. For example, note the difference in the standard deviation statistics between the two classes.} 
\label{rattlesnake_histogram_supplement}
\end{figure}

\begin{figure}[H]
\centering
\includegraphics[width=\textwidth]{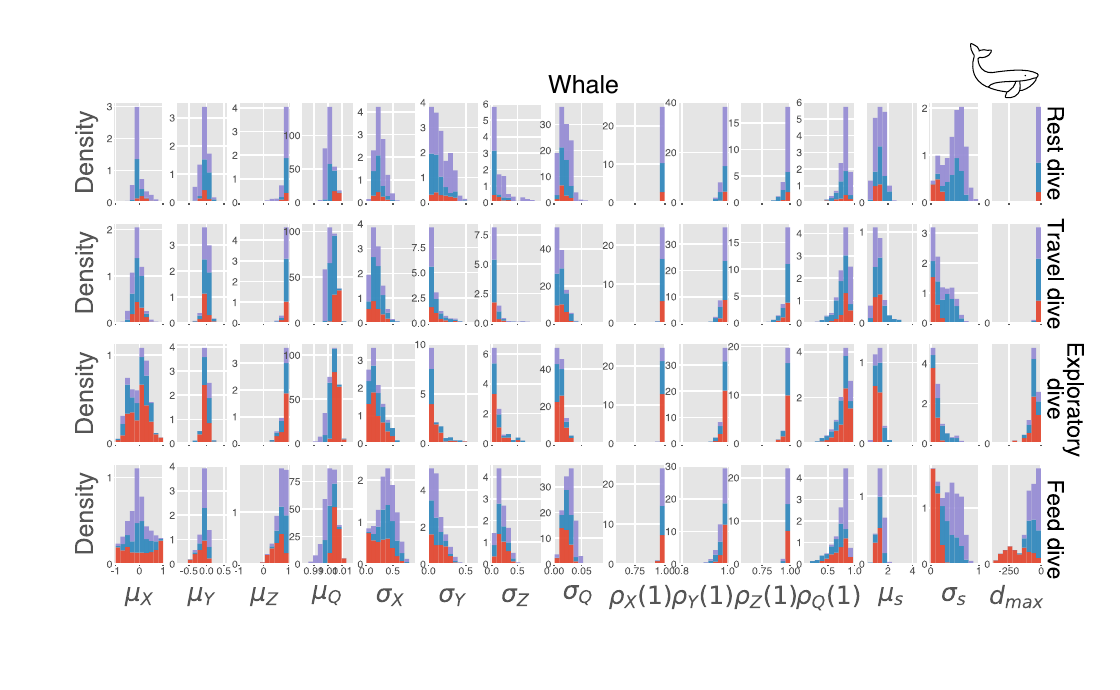}
\caption{Whale dataset summary statistics of the accelerometer $(x,y,z)$-axes, $q$-axis (root-mean-square amplitude), speed $s$, and depth $d$ for each class. Each color represents a different individual (not all individuals are shown): the histograms are stacked. $\mu$: mean, $\sigma$: standard deviation, $\rho(1)$: one-sample auto-correlation. Summary statistics computed over 60 seconds. These statistics show within and between class variation, as well as between individual variation. For example, note the individual variability in the \textit{feeding dives} for $\mu_X$, $\sigma_S$, and $d_{\mathrm{max}}$. None of the statistics obviously distinguish the classes.} 
\label{whale_histogram_supplement}
\end{figure}

\begin{figure}[H]
\centering
\includegraphics[width=\textwidth]{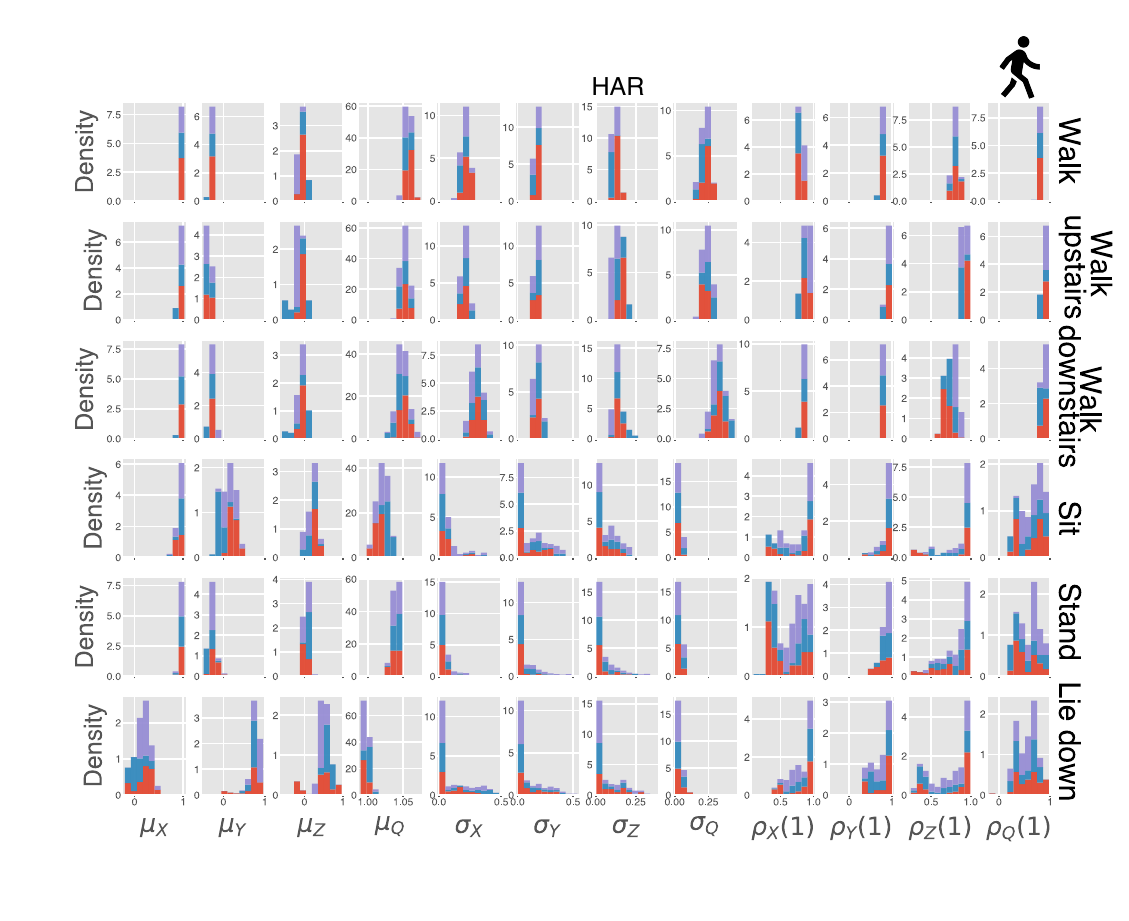}
\caption{Human dataset summary statistics of the accelerometer $(x,y,z)$-axes and $q$-axis (root-mean-square amplitude). Each color represents a different individual (not all individuals are shown): the histograms are stacked. $\mu$: mean, $\sigma$: standard deviation, $\rho(1)$: one-sample auto-correlation. Summary statistics computed over 10 seconds. These statistics show within and between class variation, as well as between individual variation. For example, note relatively minimal between individual variability. \textit{Lie down} is highly distinct from other classes based on $\mu$ statistics.} 
\label{human_histogram_supplement}
\end{figure}

\begin{figure}[H]
\centering
\includegraphics[width=\textwidth]{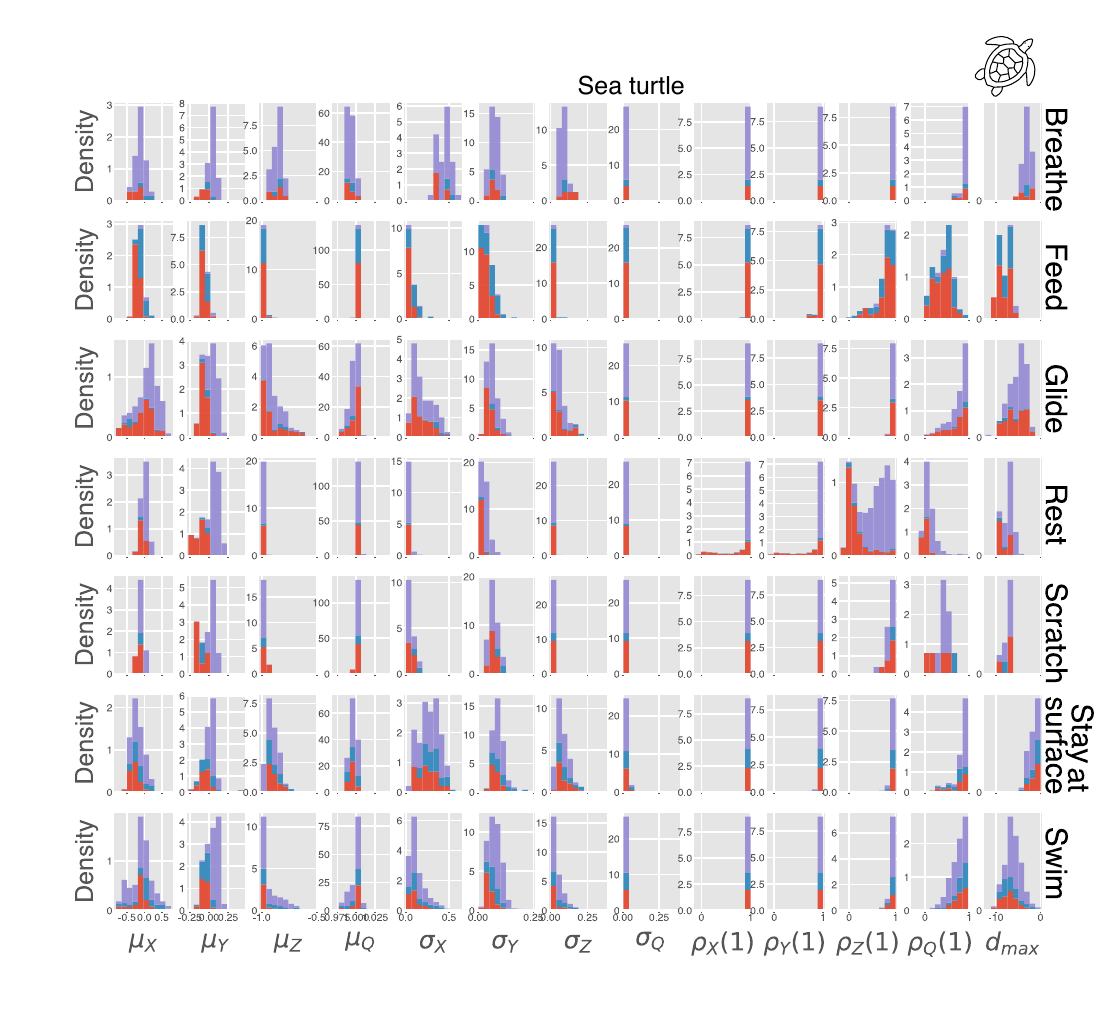}
\caption{Turtle dataset summary statistics of the accelerometer $(x,y,z)$-axes and $q$-axis (root-mean-square amplitude), as well as depth $d$. Each color represents a different individual (not all individuals are shown): the histograms are stacked. $\mu$: mean, $\sigma$: standard deviation, $\rho(1)$: one-sample auto-correlation. Summary statistics computed over 20 seconds. These statistics show within and between class variation, as well as between individual variation. For example, \textit{Stay at surface} and \textit{Breathe} are both characterized by $d_{\textrm{max}}$ near zero, whereas other behaviors tend to occur at deeper depths.} 
\label{turtle_histogram_supplement}
\end{figure}

\begin{figure}[H]
\centering
\includegraphics[width=\textwidth]{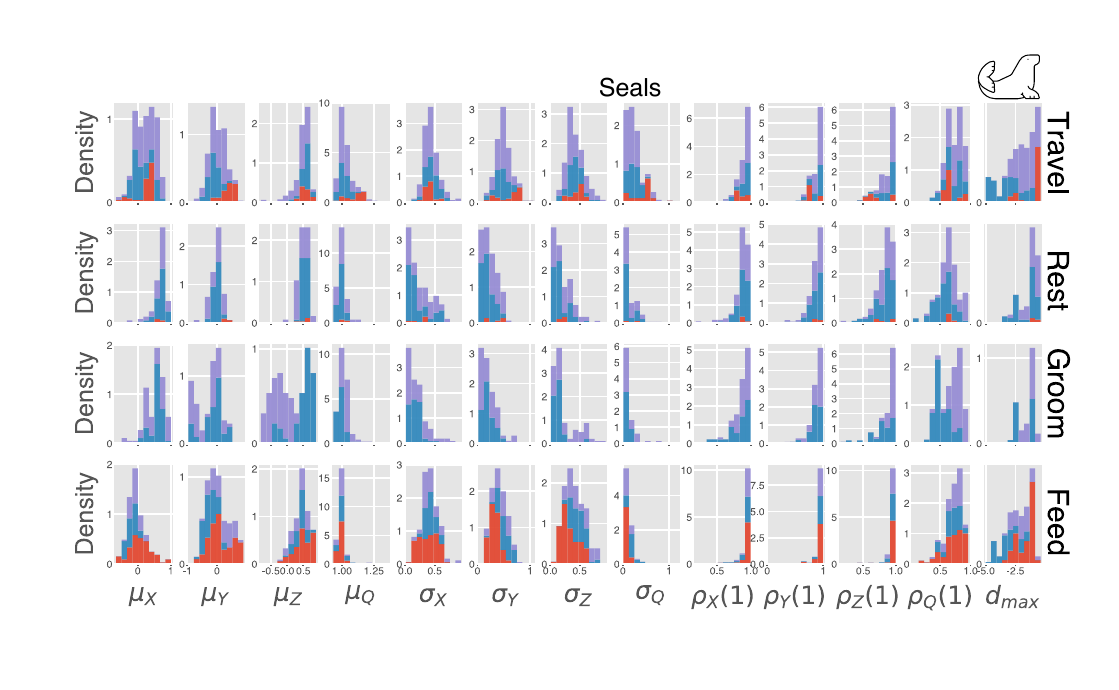}
\caption{Seals dataset summary statistics of the accelerometer $(x,y,z)$-axes and $q$-axis (root-mean-square amplitude), as well as depth $d$. Each color represents a different individual (not all individuals are shown): the histograms are stacked. $\mu$: mean, $\sigma$: standard deviation, $\rho(1)$: one-sample auto-correlation. Summary statistics computed over 10 seconds. These statistics show within and between class variation, as well as between individual variation. For example, \texttt{groom} appears to have relatively high between-individual variation, note the bimodality in $\mu_X$ and $\rho_Q(1)$.} 
\label{seals_histogram_supplement}
\end{figure}

\begin{figure}[H]
\centering
\includegraphics[width=\textwidth]{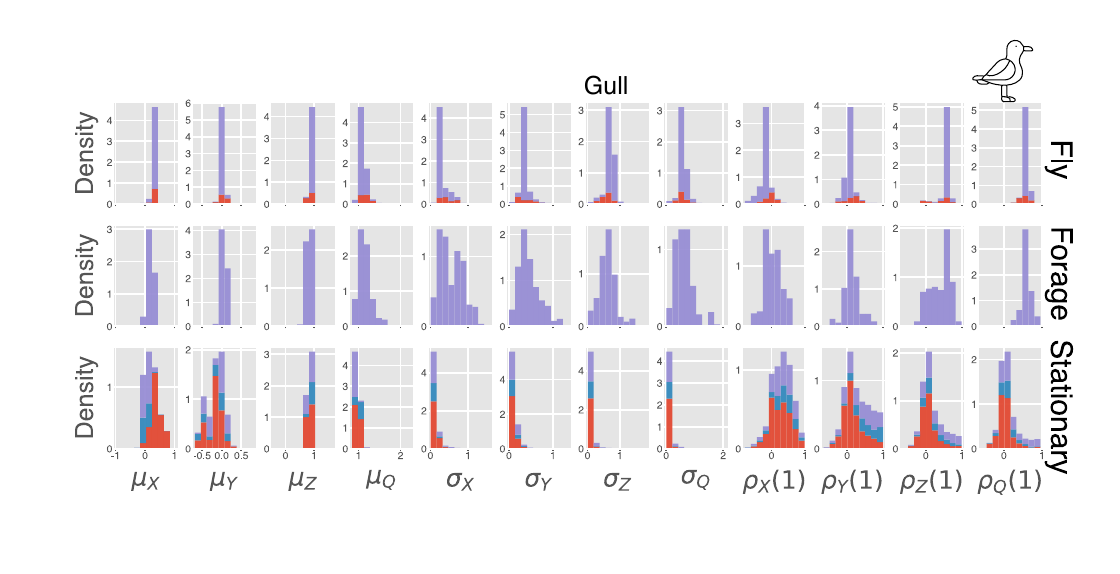}
\caption{Gull dataset summary statistics of the accelerometer $(x,y,z)$-axes and $q$-axis (root-mean-square amplitude). Each color represents a different individual (not all individuals are shown): the histograms are stacked. $\mu$: mean, $\sigma$: standard deviation, $\rho(1)$: one-sample auto-correlation. Summary statistics computed over 30 seconds. These statistics show within and between class variation, as well as between individual variation. For example, \textit{stationary} is distinguished from the other classes in the low values of the $\sigma$ statistics, while \textit{flying} and \textit{foraging} have similar modes. We cannot assess between-individual variability for \textit{foraging} in these statistics: it is a rare class and only one of the three sampled individuals performed this behavior.} 
\label{gull_histogram_supplement}
\end{figure}

\begin{figure}[H]
\centering
\includegraphics[width=\textwidth]{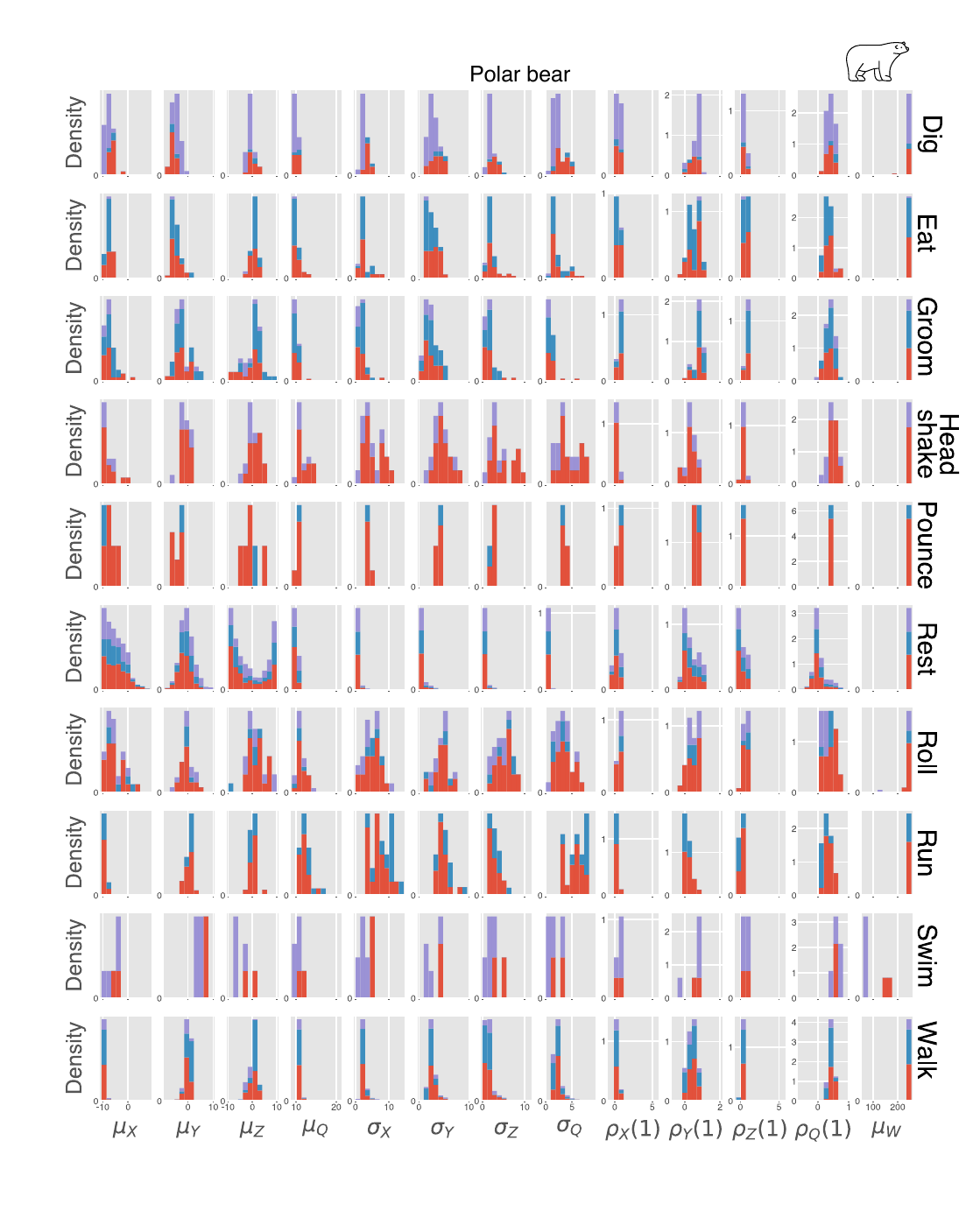}
\caption{Polar bear dataset summary statistics of the accelerometer $(x,y,z)$-axes and $q$-axis (root-mean-square amplitude), as well as the mean wet/dry value. Each color represents a different individual (not all individuals are shown): the histograms are stacked. $\mu$: mean, $\sigma$: standard deviation, $\rho(1)$: one-sample auto-correlation. Summary statistics computed over 30 seconds. These statistics show within and between class variation, as well as between individual variation. For example,  \textit{head shake} and \textit{roll} have relatively high variation in these summary statistics.} 
\label{bear_histogram_supplement}
\end{figure}

\begin{figure}[H]
\centering
\includegraphics[width=\textwidth]{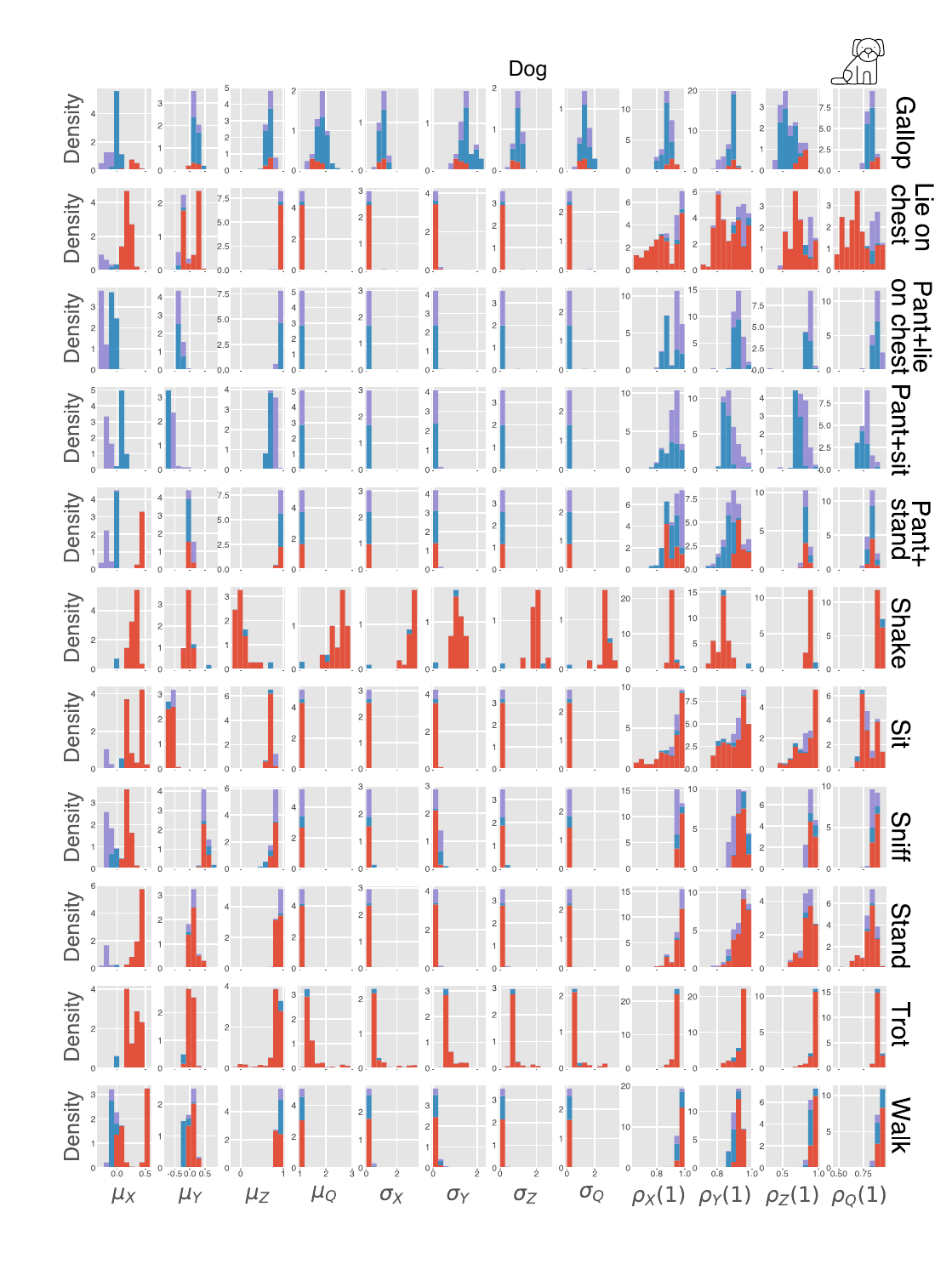}
\caption{Dog dataset summary statistics of the accelerometer $(x,y,z)$-axes and $q$-axis (root-mean-square amplitude), from the sensor placed on the back of the dogs. Each color represents a different individual (not all individuals are shown): the histograms are stacked. $\mu$: mean, $\sigma$: standard deviation, $\rho(1)$: one-sample auto-correlation. Summary statistics computed over 30 seconds. These statistics show within and between class variation, as well as between individual variation. For example,  one dog (red) panted relatively little. \textit{Trot} appears to be stereotyped behavior, given the summary statistics are highly peaked.} 
\label{dog_histogram_supplement}
\end{figure}

\begin{figure}[H]
\centering
\includegraphics[width=1.1\textwidth,angle=90]{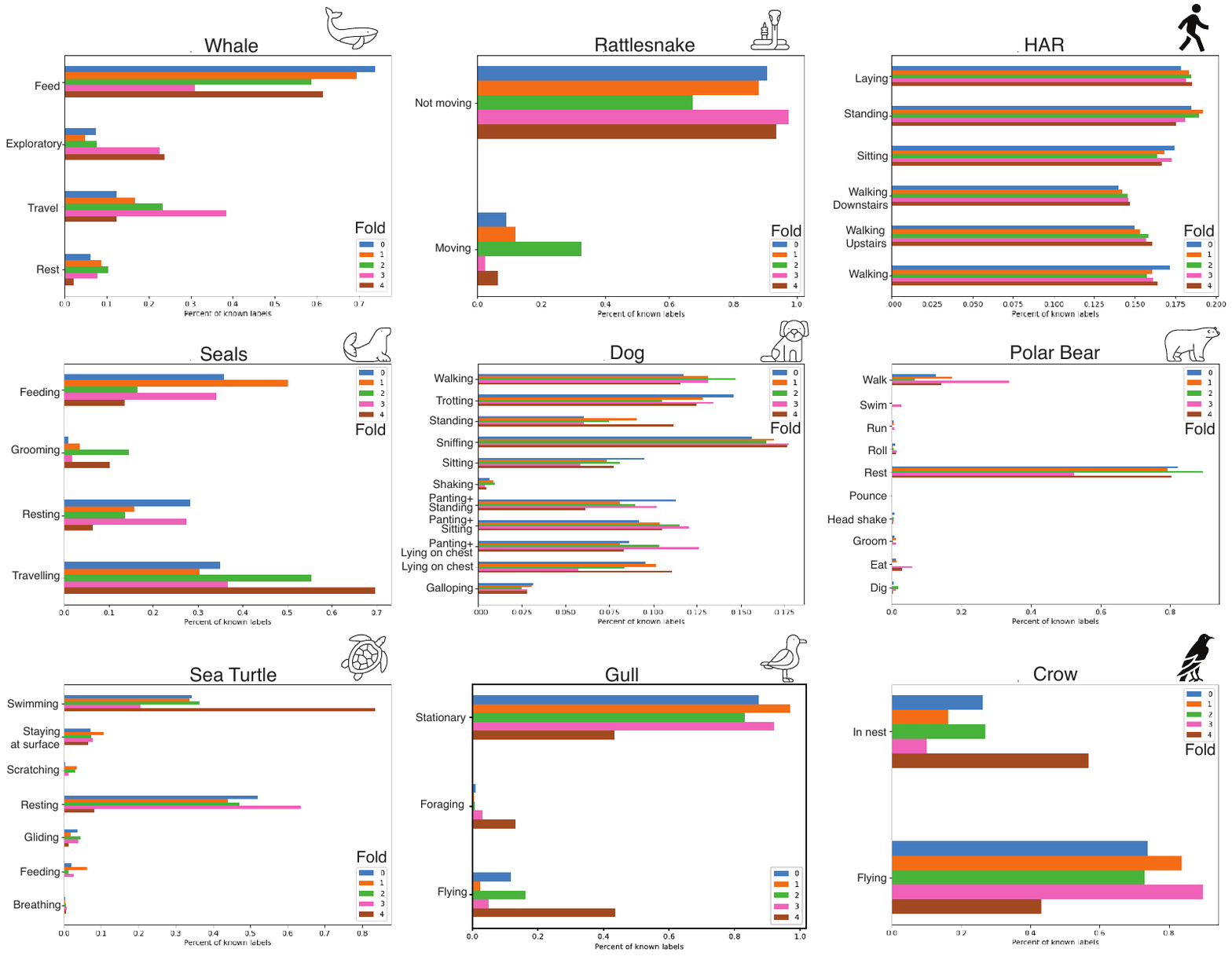}
\caption{Representation of each behavioral class in the BEBE datasets. The bars represent the proportion of sampled time steps with the given annotation, as a fraction of the total time steps with a known behavioral annotation in that fold. All behavioral classes for each dataset are listed.}
\label{supplement_class_representation}
\end{figure}


\begin{figure}[htbp]
\centering
\includegraphics[width=0.8\textwidth]{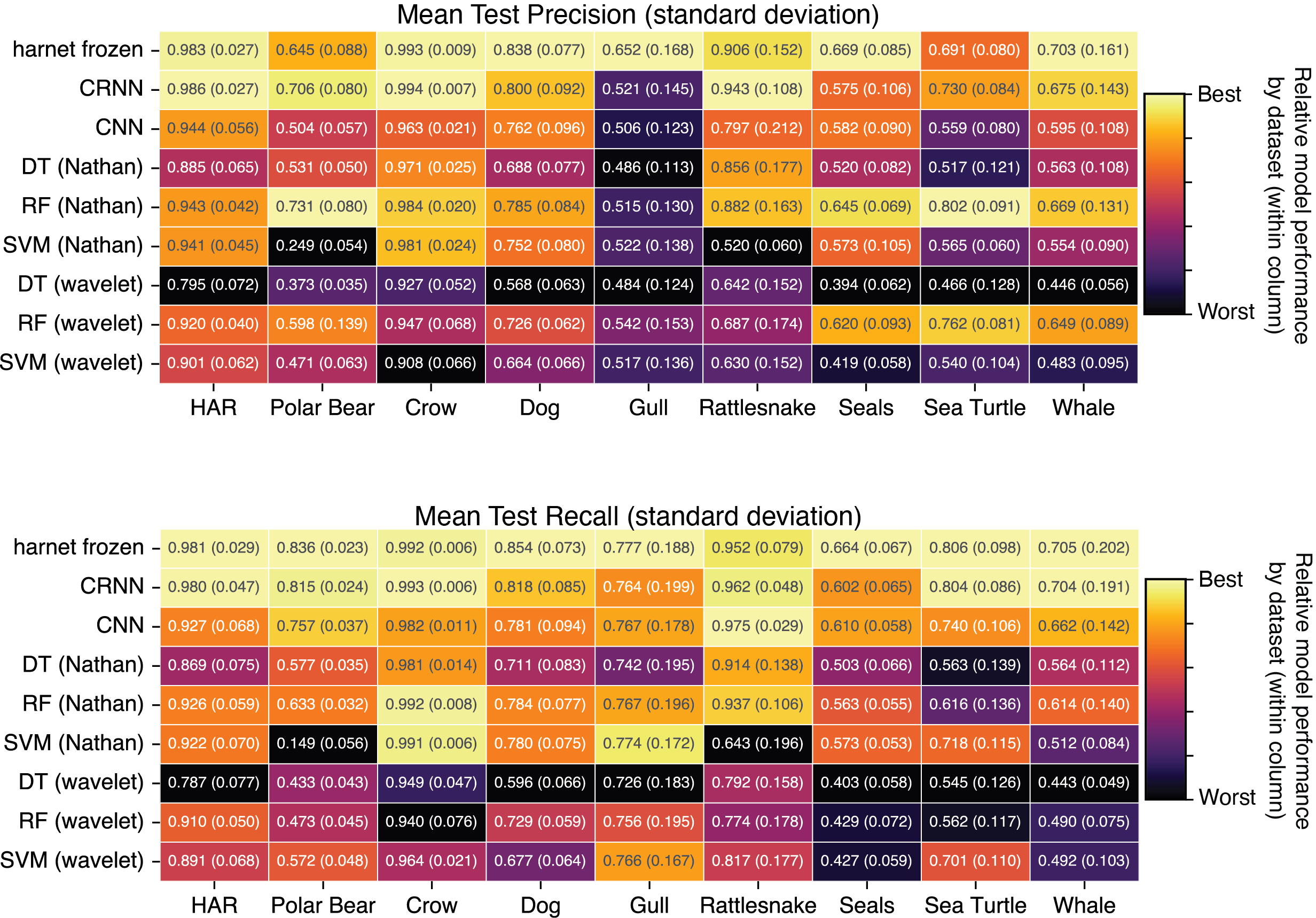}
\caption{Precision and recall results for deep neural networks and classical models. These scores correspond to Figure~\ref{basic_results}, using the full data setting and including gyroscope channels. See Figure~\ref{basic_results} for color coding. A) Precision. Neural networks (\texttt{harnet frozen} and \texttt{CRNN}) perform best on 7/9 datasets, while \texttt{RF (Nathan}  performs best on two datasets. B) Recall. One of the three neural networks shows the best performance across all nine datasets.}
\label{supplement_precision_recall_basic_results}
\end{figure}

\begin{figure}[htbp]
\centering
\includegraphics[width=0.8\textwidth]{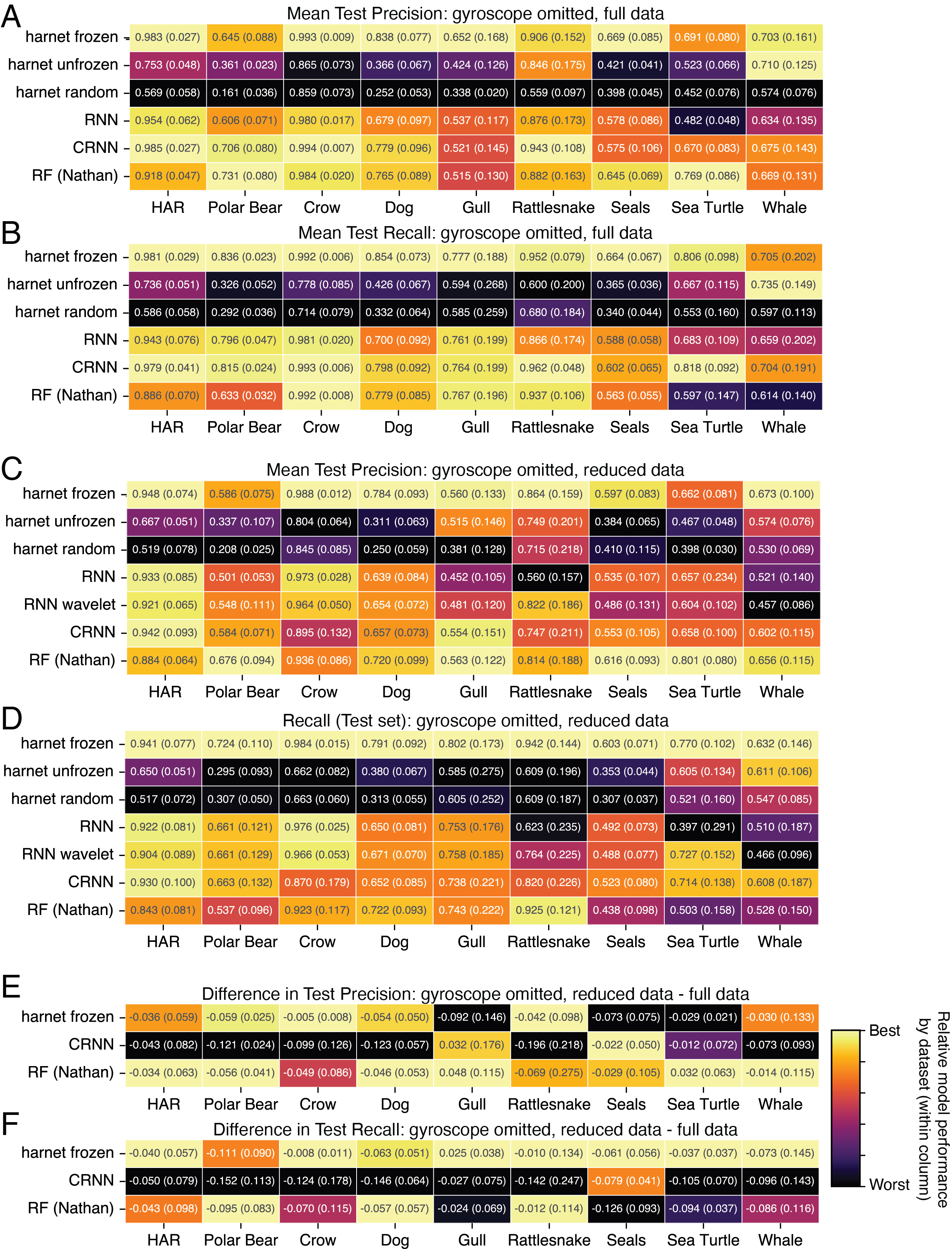}
\caption{Precision and recall results for experiments without gyroscope in both the full data and reduced data settings. Panels A/B correspond to  Figure~\ref{representation_learning_results}D, panels C/D correspond to Figure~\ref{representation_learning_results}E, and E/F correspond to Figure~\ref{representation_learning_results}F. See Figure~\ref{basic_results} for color coding. A) Precision for full data experiments with gyroscope omitted. \texttt{harnet frozen} performs best on 3/9 datasets, and \texttt{CRNN} performs best on 3/9 datsets. B) Recall for full data experiments with gyroscope omitted. \texttt{harnet frozen} performs best on 5/9 datasets. C) Precision for reduced data experiments with gyroscope omitted. \texttt{harnet frozen} performs best on 5/9 datasets. D) Recall for reduced data experiments with gyroscope omitted. \texttt{harnet frozen} performs best on all datasets. E) Difference in precision scores for reduced data and full data setting. \texttt{RF} shows the smallest decrease in performance in the reduced data setting, across 5/9 datasets. F) Difference in recall scores for reduced data and full data setting. \texttt{harnet frozen} shows the smallest decrease in performance in the reduced data setting on 6/9 datasets.}
\label{supplement_precision_recall_representation_results}
\end{figure}

\begin{figure}[htbp]
\centering
\includegraphics[width=0.5\textwidth]{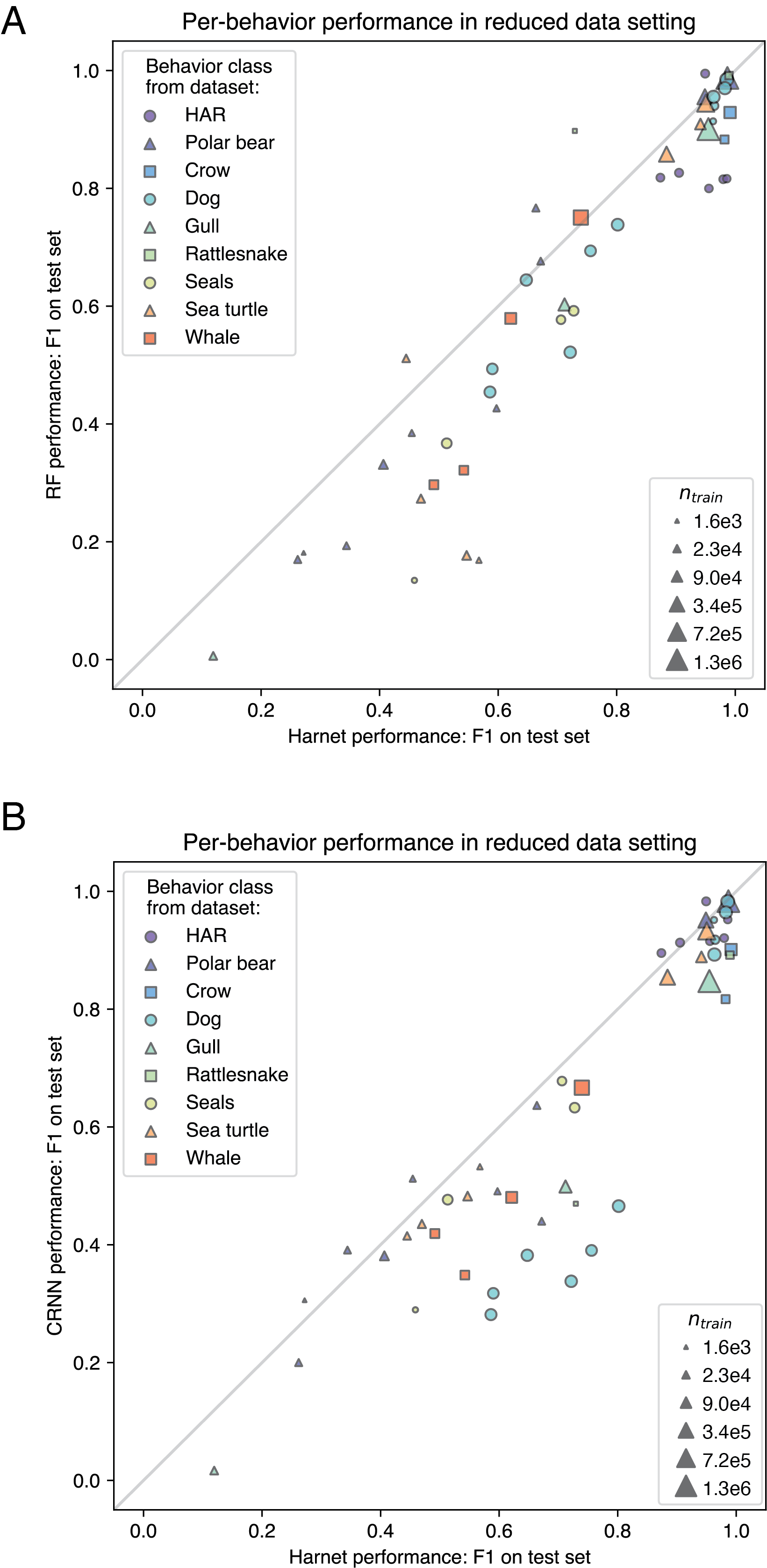}
\caption{Performance of \texttt{harnet} compared to A) \texttt{RF} and B) \texttt{CRNN} in the reduced data setting, by behavior class. \texttt{harnet} performance is better than the other models on most behavior classes, including  classes with both relatively small and large train set sizes.}
\label{harnet_vs_others_reduced_by_class}
\end{figure}

\begin{figure}[htbp]
\centering
\includegraphics[width=0.7\textwidth]{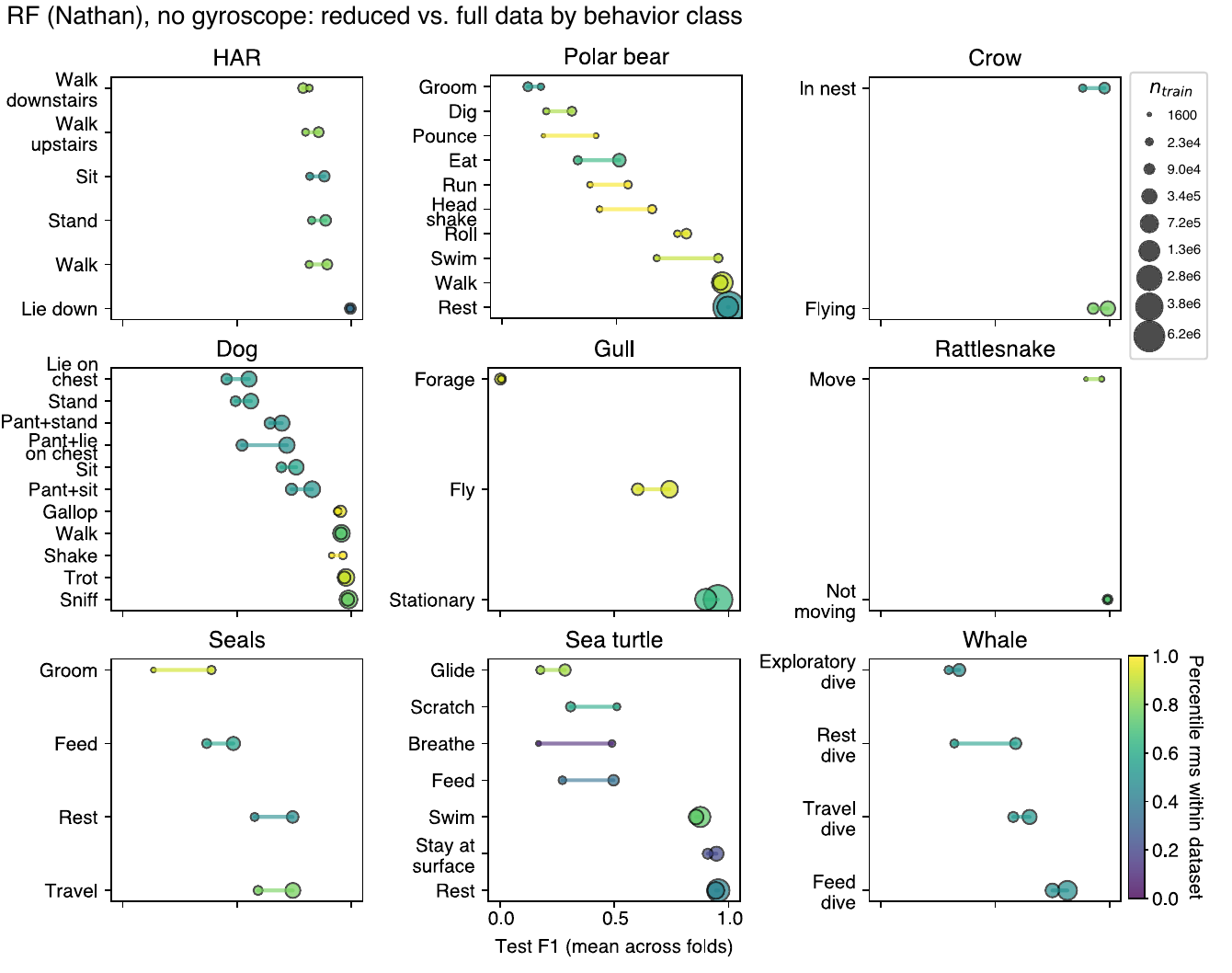}
\caption{Performance of \texttt{RF (Nathan)} in reduced and full data settings, by behavior class. Size of the marker indicates training dataset size (mean across folds). Color indicates the percentile of the average root-mean-square amplitude of datapoints in that class, as compared to the root-mean-square (rms) amplitude of all labeled datapoints.}
\label{supplement_rf_results_by_class}
\end{figure}

\begin{figure}[htbp]
\centering
\includegraphics[width=0.7\textwidth]{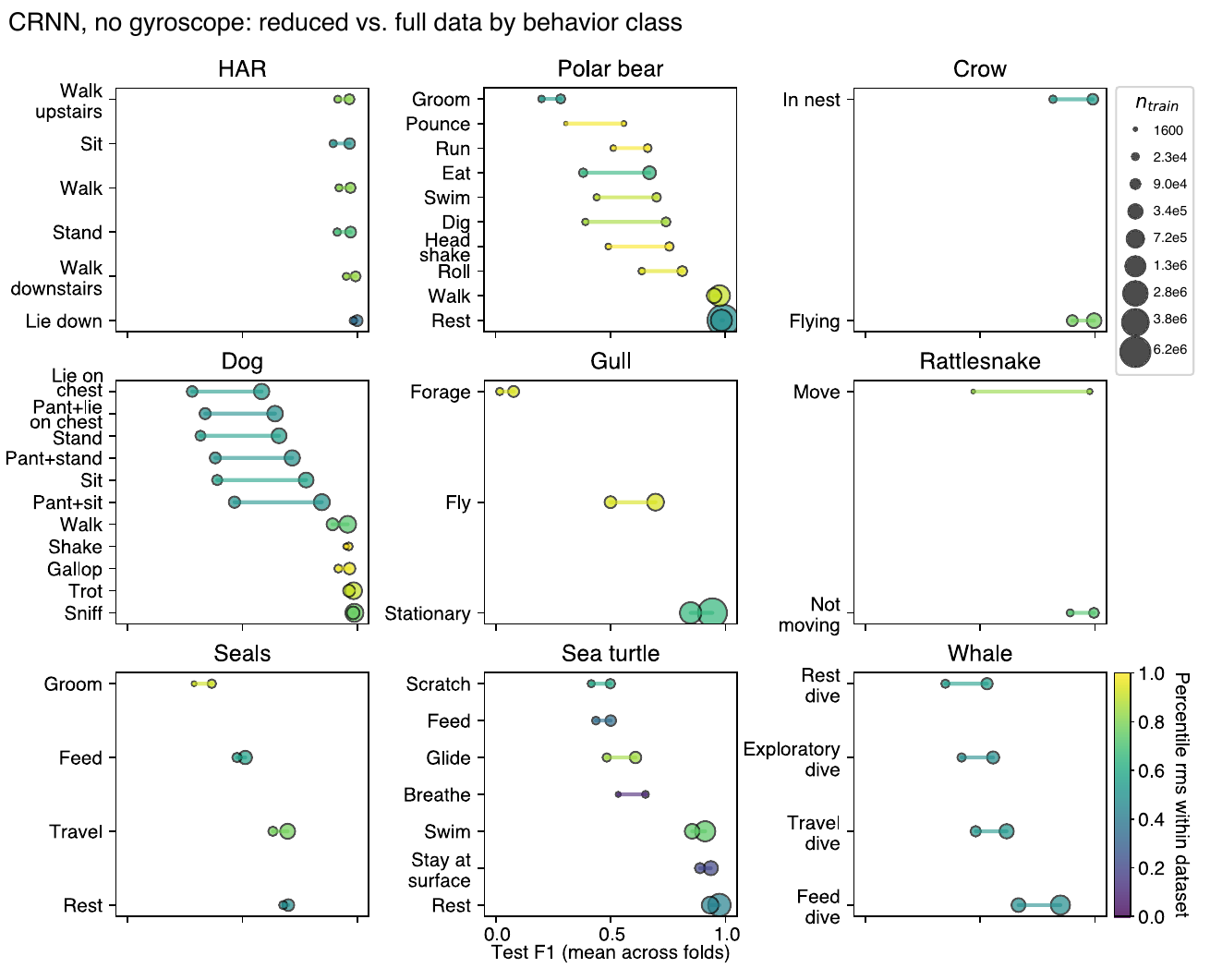}
\caption{Performance of \texttt{CRNN} in reduced and full data settings, by behavior class. See previous caption.}
\label{supplement_crnn_results_by_class}
\end{figure}

\clearpage
\begin{figure}[H]
\centering
\includegraphics[width=\textwidth]{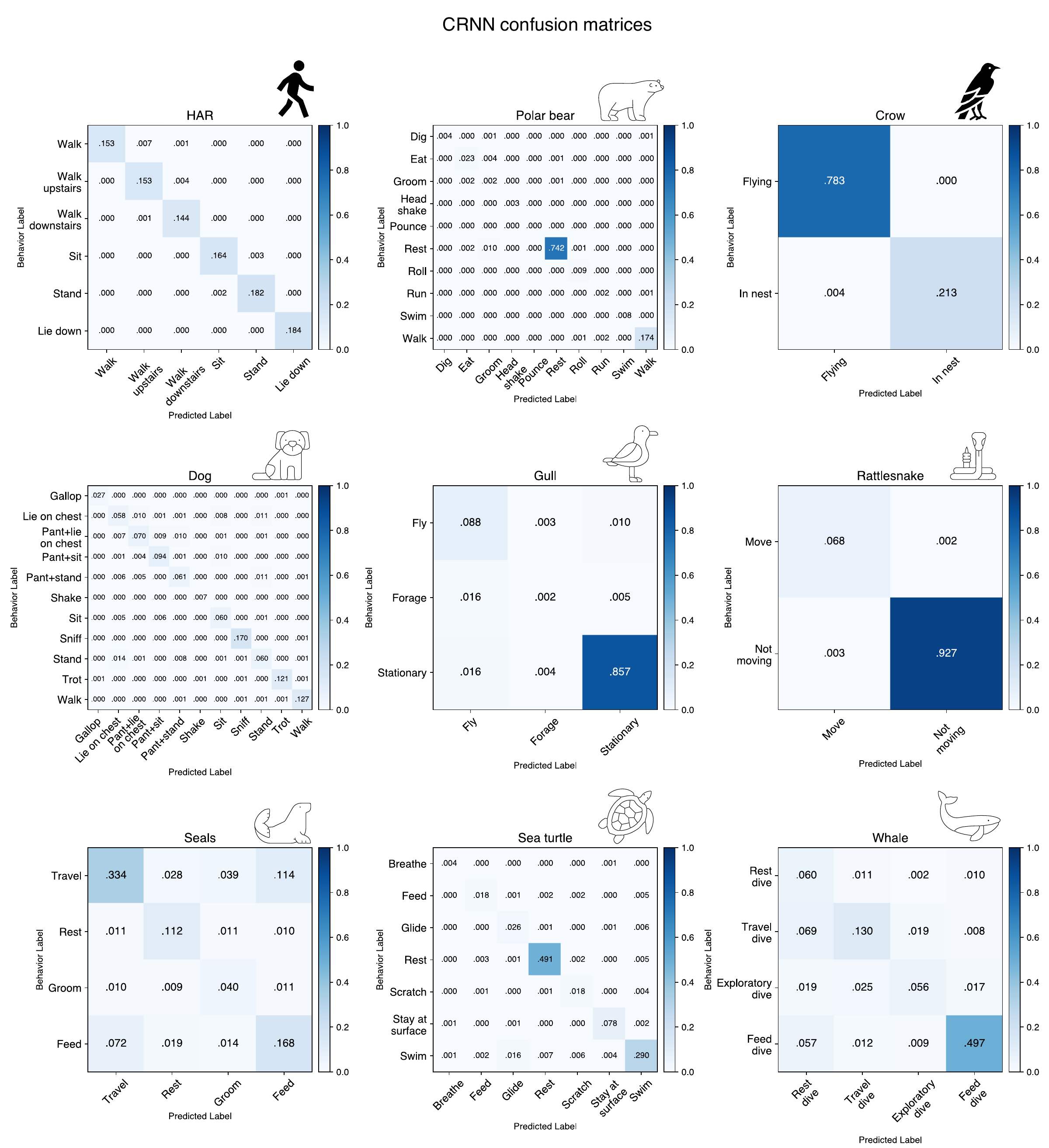} 
\caption{Confusion matrices for \texttt{CRNN} predictions versus behavioral labels, for all nine datasets in BEBE. Numbers represent the fraction of total labeled data. Computed for data taken from the test sets of the four cross validation steps that were not used for hyperparameter selection. Confusion matrices for the other experiments can be found on the Zenodo data repository.}
\label{supplement_confusion_matrices_crnn}
\end{figure}

\clearpage
\begin{figure}[H]
\centering
\includegraphics[width=\textwidth]{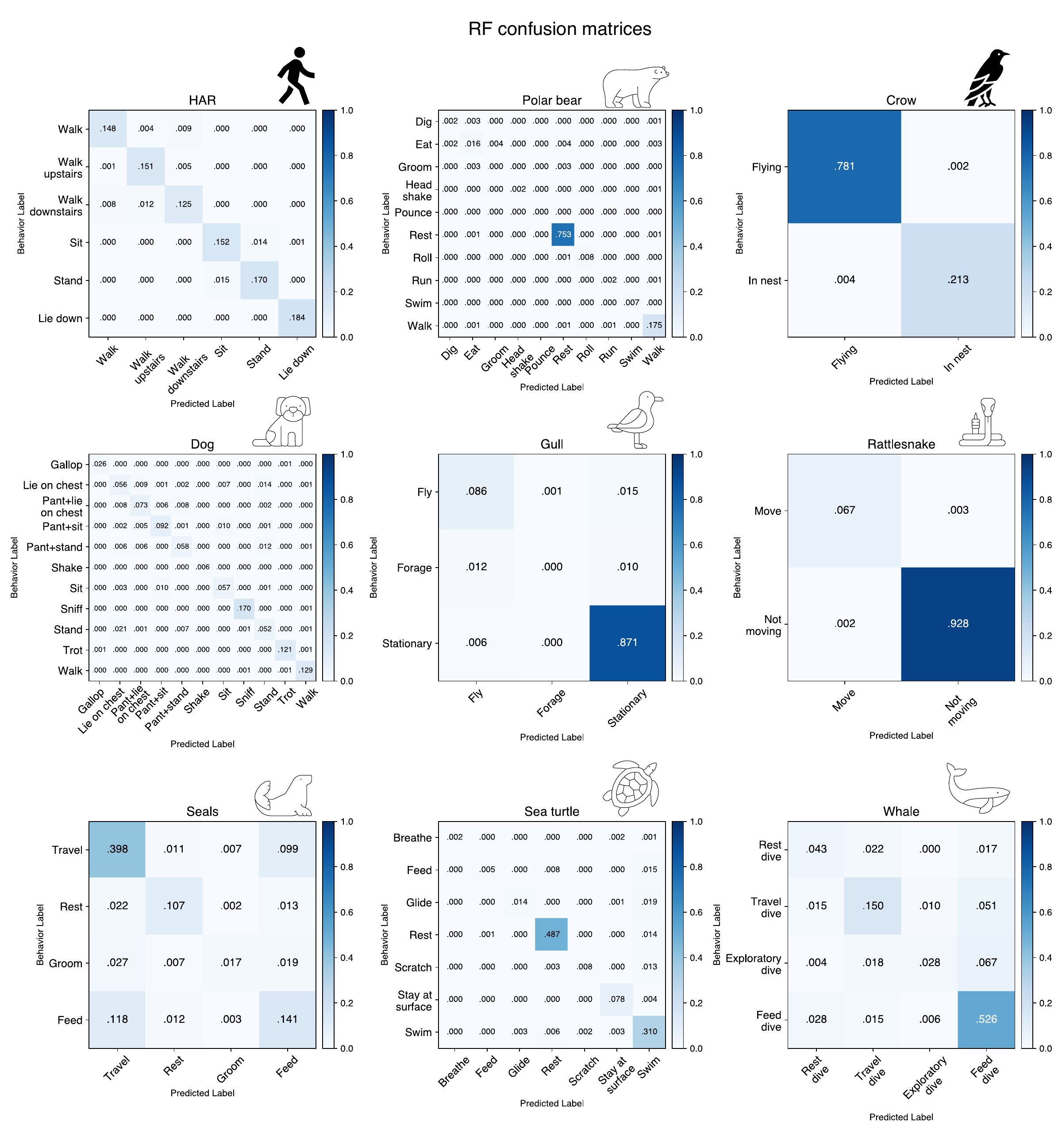} 
\caption{Confusion matrices for \texttt{RF} predictions versus behavioral labels, for all nine datasets in BEBE. Numbers represent the fraction of total labeled data. Computed for data taken from the test sets of the four cross validation steps that were not used for hyperparameter selection. Confusion matrices for the other experiments can be found on the Zenodo data repository.}
\label{supplement_confusion_matrices_rf}
\end{figure}

\begin{figure}[H]
\centering
\includegraphics[width=0.9\textwidth]{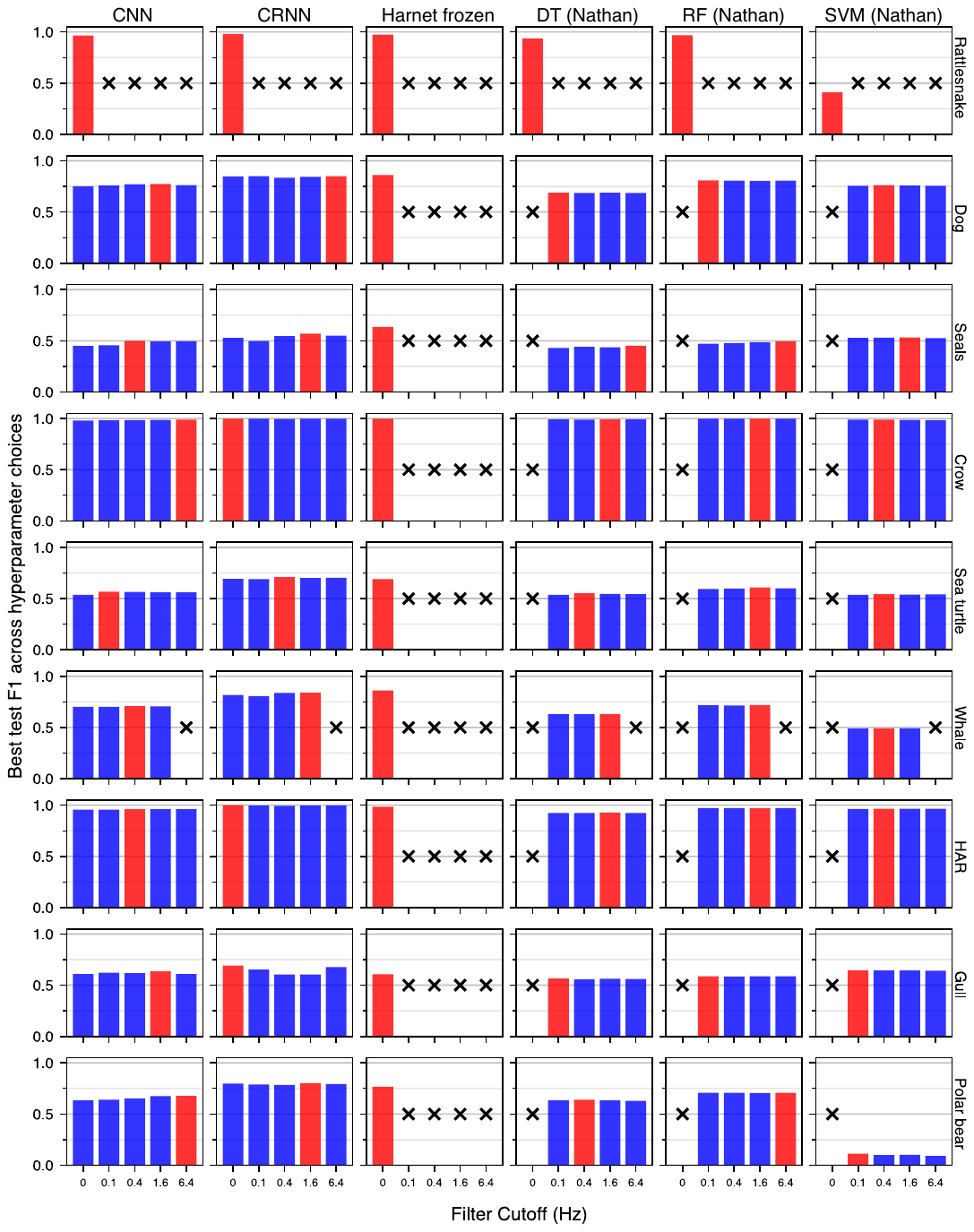}
\caption{Hyperparameter optimization of low frequency acceleration cutoff frequency. The $y$-axis indicates the best F1 score on the test set of the fold used for hyperparameter optimization, chosen from all hyperparameter with the same cutoff frequency. A cross marker indicates that this dataset/model pair did not test hyperparameters for the cutoff.  The hyperparameter chosen was not consistent within a dataset. Most models do not show large performance variation based on this hyperparameter.}
\label{supplement_static_cutoff_supervised}
\end{figure}

\begin{table}
    \centering
    \begin{tabular}{ c c c c c }
    \hline\hline
    \textbf{Model} & \textbf{Dataset} & \textbf{\#classes} & \textbf{Spearman correlation} & \textbf{p-value} \\
    \hline
    Harnet frozen&HAR&6&0.943&0.017\\
    Harnet frozen&Polar bear&10&0.952&<0.001\\
    Harnet frozen&Crow&2&1.000&1.000\\
    Harnet frozen&Dog&11&0.973&<0.001\\
    Harnet frozen&Gull&3&1.000&0.333\\
    Harnet frozen&Rattlesnake&2&1.000&1.000\\
    Harnet frozen&Seals&4&1.000&0.083\\
    Harnet frozen&Sea turtle&7&1.000&<0.001\\
    Harnet frozen&Whale&4&0.800&0.333\\
    CRNN&HAR&6&0.714&0.136\\
    CRNN&Polar bear&10&0.867&0.002\\
    CRNN&Crow&2&1.000&1.000\\
    CRNN&Dog&11&0.982&<0.001\\
    CRNN&Gull&3&1.000&0.333\\
    CRNN&Rattlesnake&2&1.000&1.000\\
    CRNN&Seals&4&1.000&0.083\\
    CRNN&Sea turtle&7&1.000&<0.001\\
    CRNN&Whale&4&1.000&0.083\\
    RF&HAR&6&0.543&0.297\\
    RF&Polar bear&10&0.976&<0.001\\
    RF&Crow&2&1.000&1.000\\
    RF&Dog&11&0.964&<0.001\\
    RF&Gull&3&1.000&0.333\\
    RF&Rattlesnake&2&1.000&1.000\\
    RF&Seals&4&1.000&0.083\\
    RF&Sea turtle&7&0.821&0.034\\
    RF&Whale&4&1.000&0.083\\
    \hline\hline
    \end{tabular}
    \caption{Spearman correlation between reduced and full data settings within a dataset, performance for the given model. p-values computed by two-sided permutation test. We note that the small number of classes limits meaningful hypothesis tests, e.g., Rattlesnake and Crow datasets have $p=1$ because there are only two classes in these datasets.}
    \label{spearmancorrelations}
\end{table}

\paragraph{Disclaimer:} Any use of trade, firm, or product names is for descriptive purposes only and does not imply endorsement by the United States Government.

\end{document}